\documentclass[11pt]{article}
\makeatletter\def\input@path{{Arxiv_version/}}\makeatother
\newcommand{\arxivonly}[1]{#1}
\newcommand{\arxivbibdir}{}
\IfFileExists{Arxiv_version/references.bib}{\renewcommand{\arxivbibdir}{Arxiv_version/}}{}
\let\oldbib\bibliography\renewcommand{\bibliography}[1]{\oldbib{\arxivbibdir#1}}
\let\oldbibstyle\bibliographystyle\renewcommand{\bibliographystyle}[1]{\oldbibstyle{\arxivbibdir#1}}
\usepackage{geometry}
\usepackage[utf8]{inputenc}
\usepackage[T1]{fontenc}
\usepackage{microtype}
\usepackage{natbib}
\setcitestyle{authoryear,round,citesep={;},aysep={,},yysep={;}}

\usepackage{amsmath,amsfonts,bm}

\def\eqref#1{equation~\ref{#1}}

\def\1{\bm{1}}

\DeclareMathAlphabet{\mathsfit}{\encodingdefault}{\sfdefault}{m}{sl}
\SetMathAlphabet{\mathsfit}{bold}{\encodingdefault}{\sfdefault}{bx}{n}

\usepackage{mathtools}
\usepackage{amssymb}
\usepackage{graphicx}
\usepackage{xcolor}
\usepackage{colortbl}
\usepackage{booktabs}
\usepackage{multirow}
\usepackage{float}
\usepackage{stfloats}
\fnbelowfloat
\usepackage{wrapfig}
\usepackage{capt-of}
\usepackage{xspace}
\usepackage{tikz}
\usetikzlibrary{arrows.meta, calc, positioning, matrix, fit, backgrounds, decorations.pathreplacing}
\usepackage{tcolorbox}
\tcbuselibrary{breakable}
\usepackage{algorithm}
\usepackage{algpseudocode}

\definecolor{promptframe}{HTML}{2980B9}
\definecolor{responseframe}{HTML}{E74C3C}

\newcommand{\Psig}[1]{\mathbf{P}^{\mathrm{B}}_{#1}}

\definecolor{starred}{HTML}{E6194B}
\newcommand{\starmark}{\textcolor{starred}{\ensuremath{\bigstar}}}

\newtcolorbox{researchquestion}{
  width=\textwidth,
  halign=justify,
  colback=gray!6,
  colframe=gray!40,
  boxrule=0pt,
  leftrule=1.2pt,
  left=9pt, right=6pt, top=4pt, bottom=4pt,
  before skip=4pt, after skip=6pt,
}

\newtcolorbox{annotex}[1]{
  breakable,
  width=\textwidth,
  colback=gray!4,
  colframe=gray!45,
  boxrule=0.4pt,
  left=6pt, right=6pt, top=4pt, bottom=4pt,
  before skip=6pt, after skip=6pt,
  fonttitle=\bfseries\small,
  coltitle=black,
  fontupper=\small,
  title={#1},
}

\usepackage{hyperref}
\usepackage{url}
\usepackage[nameinlink,capitalise]{cleveref}

\newcommand{\redbench}{\textsc{RedBench}\xspace}
\newcommand{\strongreject}{\textsc{StrongREJECT}\xspace}
\newcommand{\llmname}[1]{\textsc{#1}}
\newcommand{\llamaseven}{\llmname{Llama-2-7B}\xspace}
\newcommand{\mistralseven}{\llmname{Mistral-7B}\xspace}
\newcommand{\llamathirteen}{\llmname{Llama-2-13B}\xspace}
\newcommand{\nemo}{\llmname{Mistral-Nemo-12B}\xspace}
\newcommand{\qwenthree}{\llmname{Qwen2.5-3B}\xspace}
\newcommand{\qwenseven}{\llmname{Qwen2.5-7B}\xspace}
\newcommand{\qwenjudge}{\llmname{Qwen2.5-32B}\xspace}

\newcommand{\taskname}[1]{\textit{#1}}
\newcommand{\tasksumm}{\taskname{Summarization}\xspace}
\newcommand{\taskqa}{\taskname{Question Answering}\xspace}
\newcommand{\taskdtt}{\taskname{Data-to-Text}\xspace}

\newcommand{\baselinename}[1]{\textit{#1}}
\newcommand{\lookback}{\baselinename{LookbackLens}\xspace}
\newcommand{\lb}{\baselinename{LB}\xspace}
\newcommand{\hs}{\baselinename{HS}\xspace}
\newcommand{\lbalign}{\baselinename{LB{+}Align}\xspace}
\newcommand{\probbase}{\baselinename{Prob}\xspace}
\newcommand{\charm}{\baselinename{CHARM}\xspace}

\newcommand{\lemon}{\textbf{LEMON}\xspace}
\newcommand{\lemonW}{\textbf{LEMON-}$\boldsymbol{\mathcal{W}}$\xspace}
\newcommand{\lemonWpc}{\textbf{LEMON-}$\boldsymbol{\mathcal{W}}$\textbf{-per-cell}\xspace}

\title{LEMON: Equivariant Learning over Attention Maps for Zero-Shot Misbehavior Monitoring on New LLMs}

\author{}
\date{}

\newcommand{\hlt}[2]{%
  \begingroup\setlength{\fboxsep}{1pt}%
  \smash{\colorbox{#1}{\vphantom{Ag}#2}}%
  \endgroup}

\begin{document}

\maketitle

\arxivonly{
\vspace{-3em}
\begin{center}
\begin{tabular}{cc}
\textbf{Guy Bar-Shalom} & \textbf{Mor Geva} \\
Technion & Tel Aviv University \\[0.7em]
\textbf{Fabrizio Frasca}\textsuperscript{*} & \textbf{Haggai Maron}\textsuperscript{*} \\
Technion & Technion, NVIDIA \\
\end{tabular}
\end{center}
\vspace{-1.5em}

\begingroup
\renewcommand{\thefootnote}{\fnsymbol{footnote}}
\footnotetext[1]{Equal supervision.}
\endgroup
}

\begin{abstract}
\vspace{-0.1cm}
Large language models exhibit misbehaviors that hinder deployment in high-stakes settings. 
One approach to monitor such failures is to use an external LLM as a monitor, but this is prohibitively expensive. A well-known and efficient class of monitors can detect such failures from internal representations (such as activation probes), but are tied to the model they were trained on, requiring costly re-annotation as new LLMs appear. 
A natural way to avoid this repeated annotation cost is to develop internal-representation based monitors that transfer zero-shot across LLMs. However, such transfer is challenging because internal representations are not naturally aligned across models; in particular, attention maps, known to contain signals of misbehavior, may differ in head function, ordering, and model depth.
In this work, we characterize those cross-model attention mismatches as algebraic symmetries and use geometric deep learning principles to build \lemon (\textbf{\underline{L}}anguage-model \textbf{\underline{E}}quivariant \textbf{\underline{MO}}nitoring \textbf{\underline{N}}etwork) an \emph{online} graph neural network-based monitor, whose predictions are invariant under these symmetries.
\lemon processes attention maps token-by-token, as they are produced and outputs a score after every token (or every window of $W$ tokens) at negligible cost. 
Across 16 LLMs and six datasets, in the zero-shot cross-model setting, \lemon outperforms all baselines on both hallucination and jailbreak monitoring tasks, improving over the strongest by up to 14 ROC-AUC points.\arxivonly{\footnote{Code will be released soon.}}

\end{abstract}

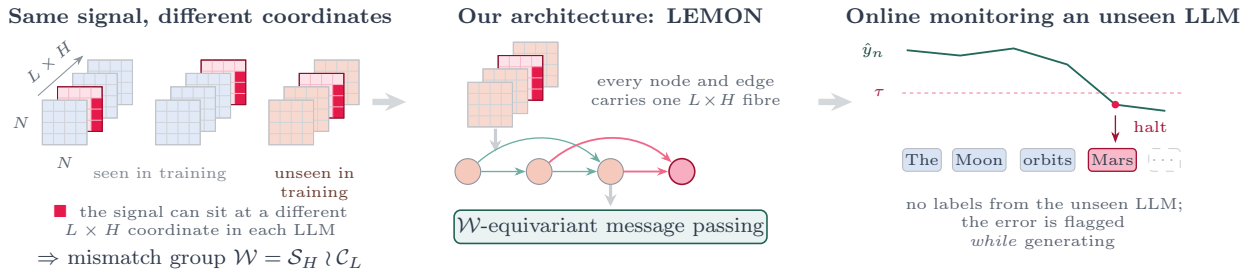
\begin{figure}[b]
  \centering
  \definecolor{trainC}{HTML}{A8C4E4}
  \definecolor{testC}{HTML}{E9896A}
  \definecolor{inkC}{HTML}{2F3B4C}
  \definecolor{hotC}{HTML}{E6194B}
  \definecolor{eagnC}{HTML}{2E8B7A}

  \begin{tikzpicture}[
    x=1.18cm, y=1.18cm, >=Stealth, font=\scriptsize, text=inkC,
    gcell/.style={draw=inkC!25, line width=0.25pt, minimum size=0.22cm,
                  inner sep=0pt},
    ttl/.style={font=\scriptsize\bfseries, text=inkC, anchor=north},
    lbl/.style={font=\tiny, text=inkC!80, align=center},
    tok/.style={circle, draw=inkC!55, fill=testC!45, minimum size=0.34cm,
                inner sep=0pt},
    flow/.style={-{Stealth[length=1.5mm,width=1.1mm]}, line width=0.5pt,
                 draw=eagnC!70},
    hop/.style={-{Stealth[length=1.6mm,width=1.6mm]}, line width=1.2pt,
                draw=inkC!25},
    axis/.style={-{Stealth[length=1.4mm,width=1.0mm]}, line width=0.4pt,
                 draw=inkC!55},
    big/.style={-{Stealth[length=2.8mm,width=2.4mm]}, line width=2.4pt,
                draw=inkC!20},
    chip/.style={rounded corners=1.5pt, draw=inkC!35, line width=0.3pt,
                 minimum height=0.30cm, inner sep=1pt, font=\tiny,
                 anchor=west},
  ]

  \useasboundingbox (0.20, 1.16) rectangle (13.80, 4.44);

  \def\teaserhotcells{2/4, 3/4, 4/4}%

  \newcommand{\teaserslice}[6]{%
    \draw[fill=#3, draw=#4, line width=0.35pt]
      (#1, #2) rectangle ++(0.50, 0.50);
    \ifnum#6=1\relax
      \foreach \r/\c in \teaserhotcells{%
        \fill[hotC] ({#1+0.125*(\c-1)}, {#2+0.50-0.125*\r})
          rectangle ++(0.125, 0.125);}
    \fi
    \foreach \i in {1,2,3}{%
      \draw[draw=#5, line width=0.12pt]
        ({#1+0.125*\i}, {#2}) -- ({#1+0.125*\i}, {#2+0.50});
      \draw[draw=#5, line width=0.12pt]
        ({#1}, {#2+0.125*\i}) -- ({#1+0.50}, {#2+0.125*\i});}
    \draw[draw=#4, line width=0.35pt] (#1, #2) rectangle ++(0.50, 0.50);}

  \newcommand{\teasercube}[4]{%
    \begin{scope}[shift={(#1,#2)}]
      \foreach \k in {3,2,1,0}{%
        \ifnum\k=#4\relax
          \teaserslice{0.17*\k}{0.15*\k}{hotC!12}{hotC!60!black}{hotC!30}{1}
        \else
          \teaserslice{0.17*\k}{0.15*\k}{#3}{inkC!30}{inkC!22}{0}
        \fi}
    \end{scope}}

  \newcommand{\teasergrid}[5]{%
    \begin{scope}[shift={(#1,#2)}]
      \foreach \r in {1,...,4}{\foreach \c in {1,...,4}{%
        \node[gcell, fill=#3] at (0.22*\c-0.11, 0.11-0.22*\r) {};}}
      \node[gcell, fill=hotC, draw=hotC!65!black, line width=0.5pt]
        at (0.22*#5-0.11, 0.11-0.22*#4) {};
    \end{scope}}

  \node[ttl] at (2.35, 4.36) {Same signal, different coordinates};

  \draw[axis] (0.53, 3.27) -- (1.04, 3.72);
  \node[lbl, rotate=41, anchor=south] at (0.785, 3.50) {$L\times H$};
  \node[lbl, anchor=east] at (0.55, 2.97) {$N$};
  \node[lbl, anchor=north] at (0.83, 2.70) {$N$};

  \teasercube{0.58}{2.72}{trainC!35}{1}
  \teasercube{1.85}{2.72}{trainC!35}{3}
  \teasercube{3.12}{2.72}{testC!32}{2}

  \node[lbl, anchor=north, text=inkC!60] at (1.90, 2.58) {seen in training};
  \node[lbl, anchor=north, text=testC!45!black] at (3.63, 2.58)
    {unseen in\\ training};

  \node[lbl, anchor=north, text width=4.0cm] at (2.35, 2.16)
    {\textcolor{hotC}{$\blacksquare$}\; the signal can sit at a different
     $L\times H$ coordinate in each LLM};
  \node[anchor=north, font=\scriptsize] at (2.35, 1.66)
    {$\Rightarrow$ mismatch group $\mathcal{W}=\mathcal{S}_H\wr\mathcal{C}_L$};

  \draw[big] (4.28, 3.20) -- (4.66, 3.20);

  \node[ttl] at (6.95, 4.36) {Our architecture: \lemon};

  \teasercube{5.35}{2.92}{testC!32}{2}

  \node[lbl, anchor=west, text width=2.5cm] at (6.62, 3.32)
    {every node and edge carries one $L\times H$ fibre};

  \draw[hop] (5.66, 2.90) -- (5.66, 2.64);

  \node[tok] (t1) at (5.35, 2.42) {};
  \node[tok] (t2) at (6.15, 2.42) {};
  \node[tok] (t3) at (6.95, 2.42) {};
  \node[tok, fill=hotC!35, draw=hotC!70!black, line width=0.6pt]
    (t4) at (7.75, 2.42) {};
  \draw[flow, draw=hotC!65, line width=0.7pt] (t3) -- (t4);
  \foreach \a/\b in {t1/t2, t2/t3} \draw[flow] (\a) -- (\b);
  \draw[flow] (t1) to[bend left=42] (t3);
  \draw[flow, draw=hotC!65, line width=0.7pt] (t2) to[bend left=42] (t4);

  \draw[hop] (t3.south) -- (6.95, 2.03);
  \node[rounded corners=2.5pt, draw=eagnC!70!black, fill=eagnC!10,
        line width=0.6pt, minimum width=4.0cm, minimum height=0.44cm,
        font=\scriptsize, inner sep=1pt]
    at (6.95, 1.80) {$\mathcal{W}$-equivariant message passing};

  \draw[big] (9.28, 3.20) -- (9.66, 3.20);

  \node[ttl] at (11.80, 4.36) {Online monitoring an unseen LLM};

  \draw[dash pattern=on 1.4pt off 1.4pt, draw=hotC!55, line width=0.4pt]
    (10.22, 3.30) -- (13.38, 3.30);
  \node[lbl, anchor=east, text=hotC!70!black] at (10.14, 3.30) {$\tau$};
  \draw[line width=0.7pt, draw=eagnC!75!black]
    (10.27, 3.78) -- (10.87, 3.72) -- (11.47, 3.80) -- (12.07, 3.62)
    -- (12.61, 3.17) -- (13.17, 3.10);
  \node[circle, fill=hotC, inner sep=1pt] at (12.61, 3.17) {};
  \node[lbl, anchor=east, text=eagnC!45!black] at (10.14, 3.78) {$\hat y_n$};

  \node[chip, fill=trainC!45, minimum width=0.50cm] at (10.22, 2.55) {The};
  \node[chip, fill=trainC!45, minimum width=0.70cm] at (10.78, 2.55) {Moon};
  \node[chip, fill=trainC!45, minimum width=0.70cm] at (11.54, 2.55) {orbits};
  \node[chip, fill=hotC!30, draw=hotC!70!black, line width=0.5pt,
        minimum width=0.62cm] at (12.30, 2.55) {Mars};
  \node[chip, draw=inkC!25, dashed, text=inkC!40,
        minimum width=0.40cm] at (12.98, 2.55) {\dots};

  \draw[flow, draw=hotC!70!black] (12.61, 3.05) -- (12.61, 2.74);
  \node[lbl, anchor=west, text=hotC!70!black] at (12.70, 2.92) {halt};

  \node[lbl, anchor=north, text width=4.0cm] at (11.80, 2.24)
    {no labels from the unseen LLM;\\
     the error is flagged \emph{while} generating};
  \end{tikzpicture}

  \caption{\textbf{Our framework for zero-shot monitoring of misbehavior across LLMs.}
  \textbf{(left)} The attention map $\mathbf{A}\in[0,1]^{N\times N\times
  L\times H}$ is a stack of attention maps indexed by
  (layer, head).
  \textbf{(middle)} \lemon operates on a structured, attributed graph constructed from the attention maps and applies $\mathcal{W}$-equivariant message passing.
  \textbf{(right)} \lemon can flag misbehavior \emph{during} generation.}
  \label{fig:teaser}
\end{figure}
\vspace{-0.5cm}
\section{Introduction}
\label{sec:Introduction}
Despite their impressive capabilities, LLMs are known to exhibit undesirable behaviors: they may memorize training data~\citep{carlini2021extracting,nasr2023scalable}, hallucinate responses~\citep{pagnoni-etal-2021-understanding,cao-etal-2022-hallucinated,qiu-etal-2023-detecting}, fall prey to flawed reasoning~\citep{song2026large}, and be jailbroken~\citep{wei2023jailbroken,zou2023universal}, i.e., persuaded to generate harmful or otherwise restricted content. As LLMs find growing use in high-stakes domains, developing \emph{lightweight}, \emph{online}\footnote{Throughout, \emph{online monitoring} means producing a score after each token (or each window of $W$ tokens) during generation.} monitors that flag potential misbehavior on-the-fly becomes increasingly important.

Misbehaviors can be monitored with separate, capable LLMs acting as \emph{judges}: they can process the provided prompt, along with the generated response to assess whether a specific misbehavior has occurred \citep{manakul2023selfcheckgpt,souly2024strongreject,mazeika2024harmbench}. This approach is, however, fundamentally incompatible with online applications due to its prohibitive computational cost incurred by the additionally required LLM executions.
A complementary and more agile class of monitors leverage instead the internal signals of the target LLM itself~\citep{azaria2023internal,kramar2026building}: data such as hidden representations, attention maps, output token log-probabilities. A pioneering example is that of probing classifiers~\citep{belinkov2022probing}, which train logistic regression models over activations to detect, e.g., the presence of the high-level concept of incorrectness. More recent approaches have generalized this idea by learning 
from more holistic views of the model’s internal computation and developing tailored architectures to process them~\cite{bar2025beyond,frasca2026neural,samuel2026uniprobe}. With their low latency, these approaches are fundamentally more suitable to monitor the generation process on the fly. The best performing methods in this class, however, typically require access to annotated training datasets to estimate their parameters. This represents a limiting factor preventing their immediate application to newly released LLMs, as each requires curating new misbehavior datasets and monitor retraining.

Encouragingly, evidence suggests that internal representations preserve a similar underlying structure across models, differing primarily in the coordinates used to encode it. For example, \citep{huh2024platonic,morcos2018insights,moschella2022relative} showed that the activation spaces of different models capture similar semantic information and can even be related through linear transformations. Likewise, \citep{ivanitskiy2025motifs,elhelo2025inferring} found recurring patterns in attention maps across models, although, potentially, at distinct layer-head locations. 
These observations motivate the main question we tackle in this paper: \textit{``Can we develop a learnable monitor capturing shared, model-agnostic, features from attention maps that are predictive for misbehaviors?''}

In this paper, we first demonstrate, through a concrete example of hallucination detection (see \Cref{fig:head-perm-motivation}), that a well-known attention-based heuristic score~\citep{chuang2024lookback} shows essentially no correlation across corresponding layers and heads across different models. 
We then show that these mismatches can be characterized by algebraic symmetries of multi-head attention. These correspond to transformations of the attention maps that change them while leaving the overall attention function unchanged. We argue that these should, therefore, be treated equivalently by the processing architecture, i.e., that we should design an \emph{invariant} monitor.
We formally characterize these symmetries, which intuitively correspond to \emph{independently} permuting the attention heads within each layer of the model, while also accounting for shifts (translations) of the layers themselves\footnote{The resulting symmetry group is the wreath product $S_H \wr C_L$, where $S_H$ captures independent permutations of the $H$ heads within each of the $L$ layers, and $C_L$ acts by cyclically permuting the layers.}. Motivated by prior work~\citep{frasca2026neural}, we formulate training a misbehavior monitor as a learning problem over an attention-induced directed graph, which we propose to endow with $(L \times H)$-structured attention scores as node and edge features, subject to the symmetries highlighed above. We hence develop \lemon (\textbf{\underline{L}}anguage-model \textbf{\underline{E}}quivariant \textbf{\underline{MO}}nitoring \textbf{\underline{N}}etwork), a graph neural network architecture designed to be invariant to these symmetries, thereby explicitly agnostic to cross-model mismatches and hence directly applicable to unseen LLMs in a zero-shot fashion (see \Cref{fig:teaser}).

\lemon is evaluated comprehensively on two monitoring tasks, hallucination and jailbreak detection, across 16 LLMs and six datasets. In this zero-shot setting, \lemon substantially outperforms baselines while remaining extremely lightweight: it introduces less than 8\% total overhead relative to standard generation. Moreover, \lemon operates online, 
allowing to halt a response mid-generation, so that a misbehavior can be detected before completion. We formalize this setting of \emph{selective generation} along with token-level coverage and risk metrics, and show that it offers a useful trade-off between abstaining from generation and reducing error rates.

\paragraph{Contributions.}
(1) We characterize a class of algebraic symmetries in multi-head attention that induce layer-head mismatches across models. (2) We introduce \lemon, a symmetry-aware monitor that transfers zero-shot to unseen LLMs. (3) \lemon operates online, enabling \emph{selective generation} with less than 8\% overhead. (4) Across 16 LLMs and six datasets on hallucination and jailbreak detection, \lemon consistently outperforms prior baselines.

\section{Related Work} \label{sec:related_work}

\paragraph{Misbehaviors in LLMs and their monitoring.} We broadly define misbehaviors as undesirable model outputs or actions deviating from intended operation. They can take many forms, including hallucinations~\citep{yang2025d,bang-etal-2025-hallulens}, jailbreaks and harmful assistance~\citep{bai2022training,yao2024survey}, reward hacking~\cite{gao2023scaling}, sycophancy~\citep{sharmatowards,wang2024fake}, and reasoning shortcuts~\citep{yuan2024llms,marconato2024bears,panas2025unreasonable}.
While our approach is applicable to a broad range of model misbehaviors, in this work, we focus on hallucinations and jailbreak-related behaviors, with an emphasis on hallucinations. \emph{Hallucinations} broadly refer to outputs unfaithful to the input context, prior reasoning, or external facts~\citep{liu2022token,huang2023survey,ji2023survey,rawte2023troubling}. 
Recent work has explored learning-based approaches that use model internals to monitor generation errors, drawing on signals such as hidden activations, attention maps, and output probabilities,~\citep{bar2025beyond,bar2026learning,chuang2024lookback,azaria2023internal}.
A complementary line of work avoids model internals: response-consistency
methods~\citep{manakul2023selfcheckgpt,kadavath2022language,Farquhar2024,kuhn2023semanticuncertaintylinguisticinvariances}
sample several generations per query, and LLM-as-a-judge monitors~\citep{zheng2023judging,liu2023g,korbak2025chain}
score the (query, response) pair with a separate model. Both therefore require
one or more additional LLM inference passes, which is prohibitive for online monitoring. 
A cheaper family uses the target model's output distribution, using token
probabilities as an uncertainty signal~\citep{guerreiro2022looking,varshney2023stitch,huang2023look}.

\paragraph{Equivariant Neural Networks.} Invariance and equivariance principles guide the design of neural architectures for structured data by encoding known symmetries, such as rotations, translations~\citep{kondor2018generalization,esteves2018learning,esteves2019equivariant,cohen2016group,thomas2018tensor}, and permutations~\citep{zaheer2017deep,maron2018invariant}, so that models do not need to relearn them from data. A canonical example is \emph{Graph Neural Networks} (GNNs)~\citep{gilmer2017neural}, with \emph{Message Passing Neural Networks} (MPNNs) as a particular instantiation that aggregates neighboring information in a permutation-equivariant manner.

\paragraph{Cross-model representation universality.} A growing body of evidence suggests that independently trained neural networks converge on similar internal representations. 
Prior work has focused mostly on \emph{activations}: model stitching~\citep{bansal2021revisiting},
representation-similarity measures~\citep{kornblith2019similarity}, the platonic representation
hypothesis~\citep{huh2024platonic}, and neurons that recur across LLM seeds and
families~\citep{gurnee2024universal}. For \emph{attention}, studies
have found head \emph{roles} that recur across models, e.g., induction and retrieval
heads~\citep{olsson2022context,wu2025retrieval,elhelo2025inferring}
In both cases, the model appear to be converging toward similar computation, despite expressed in different coordinates.
To the best of our knowledge, we are the first to move beyond hand-crafted procedures for recovering those coordinates, and instead learn a permutation-equivariant function that extracts the relevant signal for monitoring any unseen LLM zero-shot.

\vspace{-0.2cm}
\section{Extracting Universal Attention Representations} \label{sec:Extracting Universal Attention Representations}
While attention computation~\citep{vaswani2017attention} is widely used across LLMs, layer--head indices are not comparable across models: functionally similar heads may appear at different coordinates.

\begin{figure}[t]
  \centering
  \includegraphics[width=0.8\linewidth]{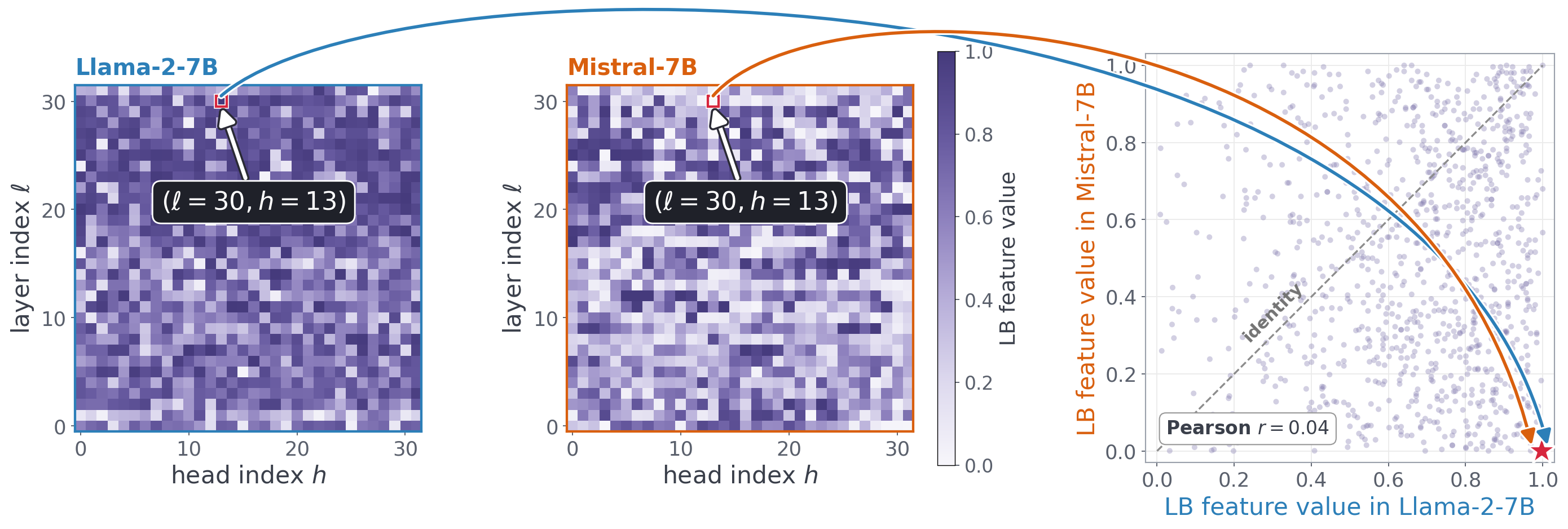}
    \vspace{-0.3cm}
\caption{\textbf{Motivating example: Head and layer roles vary across LLMs.} The same hallucinated token produces distinct $32\times32$ head patterns, with matched $(\ell,h)$ values uncorrelated ($r=0.04$).}
  \label{fig:head-perm-motivation}
  \vspace{-0.7cm}
\end{figure}

\paragraph{Motivating example.}
To make this coordinate mismatch concrete, we consider \lookback (\lb) \citep{chuang2024lookback}, a per-token, per-head feature derived from attention scores for detecting hallucinations in LLMs. This setting allows us to test whether functionally similar signals occupy consistent coordinates across models. The basic idea behind the example is simple: \lb takes on low values at specific layers and heads for tokens that are likely to hallucinate. 
However, we show that the particular layers and heads exhibiting this hallucination-associated pattern differ substantially across models.
Here, we highlight only the key insights, deferring the full details of the example to \Cref{app:motivating-example} and further background on \lb to \Cref{app:lookbacklens}. We use the RagTruth data-to-text benchmark \citep{niu2024ragtruth}, and the two LLMs, \llamaseven \citep{touvron2023llama2} and \mistralseven \citep{jiang2023mistral}, both of which have the same number of layers and heads -- $32$. The two heatmaps in \Cref{fig:head-perm-motivation} (left) show this feature
over each model's $32\times32$ (layer $\times$ head) grid: 
both exhibit a pronounced hallucination-associated signal, but at \emph{different} coordinates.
For example, at layer $30$, head $13$
(the highlighted cell), the feature is near-maximal in Llama-2-7B
yet vanishes in Mistral-7B. The scatter plot in
\Cref{fig:head-perm-motivation} (right) makes this precise: each point pairs
the two models' feature values at one \emph{matched} $(\ell,h)$ coordinate. If
heads and layers played the same roles in both models, the points would
concentrate along the diagonal; instead they are essentially uncorrelated
(Pearson $r=0.04$ across all $32\times32$ cells), with the highlighted cell
landing in the bottom-right corner (\starmark). Although we present this insight using a single token for clarity, \Cref{app:motivating-example} shows that the same phenomenon holds across all hallucinated tokens.

\paragraph{Aligning heads is not enough.} A natural fix to this mismatch is to \emph{undo} it by relabeling each LLM's layer--head coordinates into a shared canonical indexing, thereby matching heads across models, and then training a standard classifier on the aligned features. We develop such an alignment procedure (\Cref{app:alignment-methods}), but show experimentally (\Cref{sec:experiments}) that it is suboptimal compared to an architecture that is inherently invariant to these mismatches, which is the main focus of this work. We conclude this section with the central research question of this work:

\begin{researchquestion}
\centering
\emph{
Are the attention signatures of misbehavior shared across LLMs up to a relabeling of heads and layers? And can a monitor learn to read it in an unseen LLM?
}
\end{researchquestion}
\vspace{-0.3cm}
\subsection{Preliminaries and Setup}\label{subsec:preliminaries-and-setup}

Given a misbehavior identification task of interest, e.g., hallucination, we assume access to labeled data collected from $\mathcal{K}$ LLMs, indexed by $k \in \{1,\ldots,\mathcal{K}\}$. Each example consists of a query $p_q$, optionally accompanied by contextual information $p_c$, and a response $r$ generated by the model. Each example is associated with a binary label $y$, which may be defined either at the token level or at the whole sentence level. The complete dataset is given by, $\bigcup_{k=1}^{\mathcal{K}}
  \left\{
    \left(
      p_q^{(i,k)},
      p_c^{(i,k)},
      r^{(i,k)},
      y^{(i,k)}
    \right)
  \right\}_{i=1}^{M_k}$,
where the superscript $(i,k)$ denotes the $i$-th example associated with the $k$-th LLM, and $M_k$ denotes the number of labeled examples available for model $\mathcal{L}_k$. 
We use the LLM's \emph{attention} as its internal signal, representing each example by the attention map $\mathbf{A} \in [0,1]^{N \times N \times L_{\mathcal{L}} \times H_{\mathcal{L}}}$, where $N$ is the number of the considered tokens in the sentence, and $L_{\mathcal{L}}$ and $H_{\mathcal{L}}$ are the numbers of layers and heads in LLM~$\mathcal{L}$. Thus, the training dataset $\mathcal{D}$ is given by:
  $\mathcal{D} = \bigcup_{k=1}^{\mathcal{K}} \left\{ \bigl(\mathbf{A}^{(i,k)},\, y^{(i,k)}\bigr) \right\}_{i=1}^{M_k},$
where $\mathbf{A}^{(i,k)}$ is the attention map for the $i$-th example from LLM~$\mathcal{L}_k$. Our goal is to develop an architecture that learns to extract model-invariant attentional signals from $\mathcal{D}$, enabling \emph{zero-shot} transfer to a previously unseen LLM for which no task-specific labeled data are available.
\subsection{Cross-Model Attention Symmetries and Equivariant Layers}\label{subsec:attention-symmetries-and-equivariant-layers}
In what follows, we formalize cross-model mismatches in attention maps as \emph{symmetries} and characterize them as group actions. We then introduce \lemon a network invariant to these symmetries.

\begin{wrapfigure}{r}{0.3\textwidth}
  \vspace{-16pt}
  \centering
  \resizebox{\linewidth}{!}{%
  \begin{tikzpicture}[>=Stealth, font=\small,
    swaparr/.style={<->, line width=1.1pt, draw=red!68!black,
                    shorten >=2pt, shorten <=2pt},
    blk/.style={align=center, font=\footnotesize},
    dim/.style={font=\tiny, text=black!55},
    divider/.style={line width=0.6pt, dash pattern=on 2pt off 1.8pt},
  ]
    \begin{scope}
      \clip[rounded corners=4pt] (0,-0.62) rectangle (3,-1.58);
      \fill[blue!9]  (0,-0.62)   rectangle (1.5,-1.58);
      \fill[blue!24] (1.5,-0.62) rectangle (3,-1.58);
    \end{scope}
    \draw[rounded corners=4pt, line width=1pt, blue!55!black]
      (0,-0.62) rectangle (3,-1.58);
    \draw[divider, blue!40] (1.5,-0.62) -- (1.5,-1.58);

    \node[blk] at (0.75,-1.00) {$\mathbf{A}^{\ell,1}\mathbf{V}^{\ell,1}$};
    \node[dim] at (0.75,-1.36) {$N\times\tfrac{d}{2}$};
    \node[blk] at (2.25,-1.00) {$\mathbf{A}^{\ell,2}\mathbf{V}^{\ell,2}$};
    \node[dim] at (2.25,-1.36) {$N\times\tfrac{d}{2}$};

    \draw[swaparr] (0.75,-0.50) to[bend left=45] (2.25,-0.50);
    \node[red!68!black, font=\footnotesize\bfseries, anchor=south]
      at (1.5,-0.16) {$\Psig{\ell}$};

    \node at (3.62,-1.10) {$\times$};

    \begin{scope}
      \clip[rounded corners=4pt] (4.25,0) rectangle (6.45,-2.2);
      \fill[orange!13] (4.25,0)    rectangle (6.45,-1.1);
      \fill[orange!34] (4.25,-1.1) rectangle (6.45,-2.2);
    \end{scope}
    \draw[rounded corners=4pt, line width=1pt, orange!65!black]
      (4.25,0) rectangle (6.45,-2.2);
    \draw[divider, orange!50] (4.25,-1.1) -- (6.45,-1.1);

    \node[blk] at (5.35,-0.45) {$\mathbf{W}^{O,\ell,1}$};
    \node[dim] at (5.35,-0.82) {$\tfrac{d}{2}\times d$};
    \node[blk] at (5.35,-1.55) {$\mathbf{W}^{O,\ell,2}$};
    \node[dim] at (5.35,-1.92) {$\tfrac{d}{2}\times d$};
    \node[orange!65!black, font=\footnotesize\bfseries, anchor=south]
      at (5.35,0.02) {$\mathbf{W}^{O,\ell}$};

    \draw[swaparr] (6.58,-0.45) to[bend left=45] (6.58,-1.55);
    \node[red!68!black, font=\footnotesize\bfseries, anchor=west]
      at (6.98,-1.00) {$({\Psig{\ell}})^{\!\top}$};

  \end{tikzpicture}%
  }
  \vspace{-6pt}
  \caption{MHA symmetries.}
  \label{fig:head-perm-symmetry}
\end{wrapfigure}
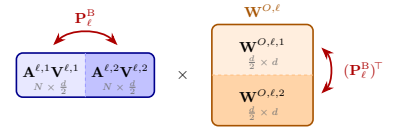
\paragraph{Attention head-permutation symmetries.} In a standard \emph{Multi-Head Attention} (MHA) block at layer~$\ell$, each head $h \in \{1,\ldots,H\}$ produces an attention map $\mathbf{A}^{\ell,h} \in \mathbb{R}^{N \times N}$ and value projections $\mathbf{V}^{\ell,h} \in \mathbb{R}^{N \times d/H}$, and the concatenated head outputs are mixed by an output projection $\mathbf{W}^{O,\ell} \in \mathbb{R}^{d \times d}$.
For any head permutation $\sigma_\ell \in \mathcal{S}_H$ with the corresponding block-permutation matrix $\Psig{\ell} \coloneqq \mathbf{P}_{\sigma_\ell} \otimes \mathbf{I}_{d/H}$, inserting $\Psig{\ell}{\Psig{\ell}}^{\!\top} = \mathbf{I}$ into the MHA output\footnote{Grouped (multi-query) attention shares query/key/value projections among subsets of heads; permutations must then act within each sharing group. The extension is straightforward.} gives,
\begin{align}
  \operatorname{MHA}^{\ell}(\mathbf{X}^{\ell})
  &= \bigl[\mathbf{A}^{\ell,1}\mathbf{V}^{\ell,1} \,\|\, \cdots \,\|\, \mathbf{A}^{\ell,H}\mathbf{V}^{\ell,H}\bigr]\,
     \Psig{\ell}\,
     {\Psig{\ell}}^{\!\top}\mathbf{W}^{O,\ell}
  \nonumber\\
  &= \bigl[\mathbf{A}^{\ell,\sigma_\ell(1)}\mathbf{V}^{\ell,\sigma_\ell(1)} \,\|\, \cdots \,\|\, \mathbf{A}^{\ell,\sigma_\ell(H)}\mathbf{V}^{\ell,\sigma_\ell(H)}\bigr]\,
     {\Psig{\ell}}^{\!\top}\mathbf{W}^{O,\ell},
  \label{eq:mha}
\end{align}
where $\mathbf{X}^{\ell} \in \mathbb{R}^{N \times d}$ are the layer-$\ell$ token representations.
Hence the relabeling $\mathbf{A}^{\ell,h} \mapsto \mathbf{A}^{\ell,\sigma_\ell(h)}$, $\mathbf{V}^{\ell,h} \mapsto \mathbf{V}^{\ell,\sigma_\ell(h)}$, $\mathbf{W}^{O,\ell} \mapsto {\Psig{\ell}}^{\!\top}\mathbf{W}^{O,\ell}$ leaves the function computed by the block unchanged -- with a possibly \emph{different} permutation at every layer, as indicated by the subscript~$\ell$.
\Cref{fig:head-perm-symmetry} illustrates the case $H=2$, where $\mathbf{W}^{O,\ell,h}$ denotes the $h$-th head block of $\mathbf{W}^{O,\ell}$.

\begin{wrapfigure}{r}{0.4\textwidth}
  \vspace{-24pt}
  \centering
  \resizebox{\linewidth}{!}{%
    \begin{tikzpicture}[>=Stealth, font=\small, thick, x=1cm, y=1cm]

\fill[blue!18]   (0, 0) rectangle (1,-1);
\fill[blue!36]   (1, 0) rectangle (2,-1);
\fill[blue!54]   (2, 0) rectangle (3,-1);
\fill[orange!20] (0,-1) rectangle (1,-2);
\fill[orange!38] (1,-1) rectangle (2,-2);
\fill[orange!56] (2,-1) rectangle (3,-2);
\fill[teal!18]   (0,-2) rectangle (1,-3);
\fill[teal!36]   (1,-2) rectangle (2,-3);
\fill[teal!54]   (2,-2) rectangle (3,-3);

\draw[gray!50, thin] (0,0) grid (3,-3);
\draw[thick]         (0,0) rectangle (3,-3);

\node[above] at (0.5, 0) {\footnotesize $h\!=\!1$};
\node[above] at (1.5, 0) {\footnotesize $h\!=\!2$};
\node[above] at (2.5, 0) {\footnotesize $h\!=\!3$};
\node[left, blue!65!black]   at (0,-0.5) {\footnotesize $\ell\!=\!1$};
\node[left, orange!85!black] at (0,-1.5) {\footnotesize $\ell\!=\!2$};
\node[left, teal!75!black]   at (0,-2.5) {\footnotesize $\ell\!=\!3$};

\node[font=\tiny] at (0.5,-0.5) {$\mathbf{A}_{n_1, n_2}^{1,1}$}; \node[font=\tiny] at (1.5,-0.5) {$\mathbf{A}_{n_1, n_2}^{1,2}$}; \node[font=\tiny] at (2.5,-0.5) {$\mathbf{A}_{n_1, n_2}^{1,3}$};
\node[font=\tiny] at (0.5,-1.5) {$\mathbf{A}_{n_1, n_2}^{2,1}$}; \node[font=\tiny] at (1.5,-1.5) {$\mathbf{A}_{n_1, n_2}^{2,2}$}; \node[font=\tiny] at (2.5,-1.5) {$\mathbf{A}_{n_1, n_2}^{2,3}$};
\node[font=\tiny] at (0.5,-2.5) {$\mathbf{A}_{n_1, n_2}^{3,1}$}; \node[font=\tiny] at (1.5,-2.5) {$\mathbf{A}_{n_1, n_2}^{3,2}$}; \node[font=\tiny] at (2.5,-2.5) {$\mathbf{A}_{n_1, n_2}^{3,3}$};
\node[font=\Large] at (1.5,-3.75) {$\mathbf{A}$};

\draw[->, double, thick] (3.25,-1.5) -- (4.75,-1.5);
\node[above=2pt, font=\scriptsize] at (4.0,-1.5) {$g=(\pi,\vec{\sigma})$};
\node[draw=gray!55, rounded corners=2pt, fill=gray!8, inner sep=2pt,
      font=\scriptsize] at (4.0,-2.3) {$\pi\!=\!(1\;2\;3)$};

\fill[teal!36]   (5, 0) rectangle (6,-1);
\fill[teal!18]   (6, 0) rectangle (7,-1);
\fill[teal!54]   (7, 0) rectangle (8,-1);
\fill[blue!54]   (5,-1) rectangle (6,-2);
\fill[blue!36]   (6,-1) rectangle (7,-2);
\fill[blue!18]   (7,-1) rectangle (8,-2);
\fill[orange!20] (5,-2) rectangle (6,-3);
\fill[orange!38] (6,-2) rectangle (7,-3);
\fill[orange!56] (7,-2) rectangle (8,-3);

\draw[gray!50, thin] (5,0) grid (8,-3);
\draw[thick]         (5,0) rectangle (8,-3);

\node[above] at (5.5, 0) {\footnotesize $h\!=\!1$};
\node[above] at (6.5, 0) {\footnotesize $h\!=\!2$};
\node[above] at (7.5, 0) {\footnotesize $h\!=\!3$};
\node[right, teal!75!black]   at (8,-0.5) {\footnotesize $\ell\!=\!3$};
\node[right, blue!65!black]   at (8,-1.5) {\footnotesize $\ell\!=\!1$};
\node[right, orange!85!black] at (8,-2.5) {\footnotesize $\ell\!=\!2$};

\node[font=\tiny] at (5.5,-0.5) {$\mathbf{A}_{n_1, n_2}^{3,2}$}; \node[font=\tiny] at (6.5,-0.5) {$\mathbf{A}_{n_1, n_2}^{3,1}$}; \node[font=\tiny] at (7.5,-0.5) {$\mathbf{A}_{n_1, n_2}^{3,3}$};
\node[font=\tiny] at (5.5,-1.5) {$\mathbf{A}_{n_1, n_2}^{1,3}$}; \node[font=\tiny] at (6.5,-1.5) {$\mathbf{A}_{n_1, n_2}^{1,2}$}; \node[font=\tiny] at (7.5,-1.5) {$\mathbf{A}_{n_1, n_2}^{1,1}$};
\node[font=\tiny] at (5.5,-2.5) {$\mathbf{A}_{n_1, n_2}^{2,1}$}; \node[font=\tiny] at (6.5,-2.5) {$\mathbf{A}_{n_1, n_2}^{2,2}$}; \node[font=\tiny] at (7.5,-2.5) {$\mathbf{A}_{n_1, n_2}^{2,3}$};
\node[font=\Large] at (6.5,-3.75) {$w \cdot \mathbf{A}_{n_1, n_2}$};

\draw[<->, purple!80, thick, shorten >=2pt, shorten <=2pt]
  (5.5, 0.5) .. controls (6.5, 1.5) .. (7.5, 0.5);
\node[above, purple!80, font=\scriptsize] at (6.5, 1.2) {$\sigma_1\!=\!(1\;3)$};
\draw[<->, red!65, thick, shorten >=2pt, shorten <=2pt]
  (5.5, 0.5) .. controls (6.0, 0.88) .. (6.5, 0.5);
\node[right, red!65, font=\scriptsize] at (6.0, 0.88) {$\sigma_3\!=\!(1\;2)$};
\node[gray!70, font=\scriptsize] at (6.5,-3.3) {$\sigma_2\!=\!\mathrm{id}$};

\end{tikzpicture}%
  }
  \caption{Action of $\mathcal{W}=\mathcal{S}_H \wr \mathcal{C}_L$ on $\mathbf{A}$ for $L{=}H{=}3$: rows are reordered by~$\pi$; heads within each row by~$\sigma_\ell$.}
  \label{fig:wreath-action}
  \vspace{-8pt}
\end{wrapfigure}
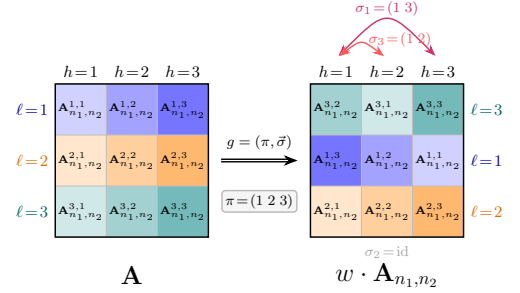
\paragraph{Joint layer--head symmetries.} Stacking MHA blocks over depths
$\ell \in \{1,\ldots,L\}$ yields attention maps $\{\mathbf{A}^{\ell,h}\}$.
Unlike heads, layers admit no exact interchange identity --- they are
sequentially composed, so their order matters --- yet their \emph{absolute}
indices carry no fixed meaning across models of different depths. Following \citep{bar2025beyond,kasten2024fast}, we model layer relabelings by $\mathcal{C}_L$: a modeling prior, not an exact symmetry, that removes dependence on absolute layer indices across depths. Since the head permutations might be different across the different layers, we encode the joint symmetry as the wreath product
$\mathcal{W} = \mathcal{S}_H \wr \mathcal{C}_L$ -- exactly the group whose elements $w = (\sigma_\ell, \pi_\ell)$ pair a cyclic layer relabeling
$\pi_\ell$ with an \emph{independent} head permutation
$\sigma_\ell \in \mathcal{S}_H$ at every depth. The action, illustrated in \Cref{fig:wreath-action}, is
\begin{align}
  \forall w \in \mathcal{W}, \quad (w \cdot \mathbf{A})^{l,h}_{n_1,n_2}
  = \mathbf{A}^{\pi_\ell(l),\, \sigma_{\pi_\ell(l)}^{-1}(h)}_{n_1,\, n_2}.
\end{align}

\begin{wrapfigure}{r}{0.27\textwidth}
  \vspace{-26pt}
  \centering
  \resizebox{\linewidth}{!}{%
\begin{tikzpicture}[>=Stealth, font=\small, thick, x=1cm, y=1cm]

\fill[gray!10] (0,0) rectangle (2.6,-2.6);
\fill[yellow!60!orange!30] (1.3,-0.65) rectangle (1.95,-1.3);
\draw[gray!45, thin] (0,0) grid[step=0.65] (2.6,-2.6);
\draw[thick] (0,0) rectangle (2.6,-2.6);

\node[above, font=\scriptsize] at (0.325, 0) {$1$};
\node[above, font=\scriptsize] at (0.975, 0) {$2$};
\node[above, font=\scriptsize\bfseries, text=orange!80!black] at (1.625, 0) {$3$};
\node[above, font=\scriptsize] at (2.275, 0) {$4$};
\node[left, font=\scriptsize] at (0, -0.325) {$1$};
\node[left, font=\scriptsize\bfseries, text=orange!80!black] at (0, -0.975) {$2$};
\node[left, font=\scriptsize] at (0, -1.625) {$3$};
\node[left, font=\scriptsize] at (0, -2.275) {$4$};
\node[above, font=\footnotesize] at (1.3, 0.32) {$n_2$};
\node[left,  font=\footnotesize] at (-0.28, -1.3) {$n_1$};
\node[below, font=\normalsize] at (1.3, -2.7) {$\mathbf{A}$};

\draw[<->, purple!70!black, thin, dashed, shorten >=2pt, shorten <=2pt]
  (-0.55, -0.325) .. controls (-0.95,-1.3) .. (-0.55,-2.275);
\node[left, purple!70!black, font=\scriptsize, align=center] at (-0.95,-1.3) {$\tau\in\mathcal{C}_N$\\(approx.)};
\draw[<->, purple!70!black, thin, dashed, shorten >=2pt, shorten <=2pt]
  (0.325, 0.55) .. controls (1.3, 0.95) .. (2.275, 0.55);
\node[above, purple!70!black, font=\scriptsize] at (1.3, 0.95) {$\tau\in\mathcal{C}_N$};

\draw[dashed, gray!55, thin] (1.95,-0.65) -- (4.0, 1.0);
\draw[dashed, gray!55, thin] (1.95,-1.30) -- (4.0,-2.0);

\begin{scope}[shift={(4.0, 1.0)}]

\fill[blue!18]   (0, 0) rectangle (1,-1);   \fill[blue!36]   (1, 0) rectangle (2,-1);   \fill[blue!54]   (2, 0) rectangle (3,-1);
\fill[orange!20] (0,-1) rectangle (1,-2);   \fill[orange!38] (1,-1) rectangle (2,-2);   \fill[orange!56] (2,-1) rectangle (3,-2);
\fill[teal!18]   (0,-2) rectangle (1,-3);   \fill[teal!36]   (1,-2) rectangle (2,-3);   \fill[teal!54]   (2,-2) rectangle (3,-3);

\draw[gray!50, thin] (0,0) grid (3,-3);
\draw[thick, orange!80!black] (0,0) rectangle (3,-3);

\node[above, font=\scriptsize] at (0.5, 0) {$h\!=\!1$};
\node[above, font=\scriptsize] at (1.5, 0) {$h\!=\!2$};
\node[above, font=\scriptsize] at (2.5, 0) {$h\!=\!3$};
\node[left, blue!65!black,   font=\scriptsize] at (0,-0.5) {$\ell\!=\!1$};
\node[left, orange!85!black, font=\scriptsize] at (0,-1.5) {$\ell\!=\!2$};
\node[left, teal!75!black,   font=\scriptsize] at (0,-2.5) {$\ell\!=\!3$};

\node[font=\scriptsize] at (0.5,-0.5) {$\mathbf{A}^{1,1}_{2,3}$}; \node[font=\scriptsize] at (1.5,-0.5) {$\mathbf{A}^{1,2}_{2,3}$}; \node[font=\scriptsize] at (2.5,-0.5) {$\mathbf{A}^{1,3}_{2,3}$};
\node[font=\scriptsize] at (0.5,-1.5) {$\mathbf{A}^{2,1}_{2,3}$}; \node[font=\scriptsize] at (1.5,-1.5) {$\mathbf{A}^{2,2}_{2,3}$}; \node[font=\scriptsize] at (2.5,-1.5) {$\mathbf{A}^{2,3}_{2,3}$};
\node[font=\scriptsize] at (0.5,-2.5) {$\mathbf{A}^{3,1}_{2,3}$}; \node[font=\scriptsize] at (1.5,-2.5) {$\mathbf{A}^{3,2}_{2,3}$}; \node[font=\scriptsize] at (2.5,-2.5) {$\mathbf{A}^{3,3}_{2,3}$};

\node[below, font=\normalsize] at (1.5,-3.15) {$\mathbf{A}_{2,3}$};

\draw[<->, purple!80, thick, shorten >=2pt, shorten <=2pt]
  (0.5, 0.55) .. controls (1.5, 1.3) .. (2.5, 0.55);
\node[above, purple!80, font=\scriptsize] at (1.5, 1.3) {$(\pi,\vec{\sigma})\in\mathcal{S}_H \wr \mathcal{C}_L$};

\end{scope}
\end{tikzpicture}}
  \vspace{-22pt}
  \caption{
  $\mathbf{A}$ symmetries.
  }
  \label{fig:attention-tensor-symmetry}

\end{wrapfigure}
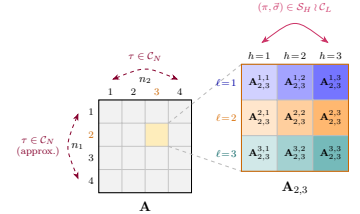
\paragraph{Tokens.} Token order is meaningful, so the token axes admit no exact symmetry. Still, the absolute indices $(n_1,n_2)$ carry no shared meaning across tokenizers and sequence lengths, so we treat translation along the token
axes as an \emph{approximate} symmetry, modeled by the cyclic group
$\mathcal{C}_N$ acting on both axes by the same shift. \Cref{fig:attention-tensor-symmetry} visualizes the full symmetries of the attention map $\mathbf{A}$. 

In the next section, we use the symmetry analysis above to guide the design of \lemon, combining components suited to the data and its symmetries without necessarily enforcing exact invariance throughout. This is analogous to Vision Transformers with positional encodings outperforming CNNs despite breaking translation symmetry. 

\subsection{From attention maps to Graphs with Structured Features}\label{subsec:attention-graphs-structured}

\paragraph{Overview.}
The attention map $\mathbf{A}$ admits a natural graph representation, with tokens as ordered nodes and attention scores defining node and edge attributes. 
Motivated by \charm~\citep{frasca2026neural}, we cast the task as learning on this attributed directed graph. Our formulation differs from~\citep{frasca2026neural} in two key ways: (i) node and edge attributes retain their structured $L\times H$ attention slices, which transform under the layer--head symmetry $\mathcal{W}$ developed in \Cref{subsec:attention-symmetries-and-equivariant-layers}, and (ii) the graph is sequential, preserving token order. We then introduce \lemon, a message-passing-based architecture whose updates respect $\mathcal{W}$ and that operates online, producing scores token by token as the response is generated.

\paragraph{Attention graphs.} Concretely, we view the attention $\mathbf{A} \in \mathbb{R}^{N \times N \times L \times H}$ as an attributed directed graph $G=(V,E)$: each token position $n$ is a node $v_n \in V$ with self-attention attribute $\mathbf{A}_{n,n} \in [0,1]^{L\times H}$ (scores across all layers and heads); each directed edge $(n_1,n_2)\in E$ carries cross-attention attribute $\mathbf{A}_{n_1,n_2} \in [0,1]^{L\times H}$ (attention from token~$n_1$ to token~$n_2$). \emph{Importantly, these node and edge attributes are structured $L\times H$ tensors rather than vectors.}

\paragraph{Our \textbf{\underline{L}}anguage-model \textbf{\underline{E}}quivariant \textbf{\underline{MO}}nitoring \textbf{\underline{N}}etwork (\lemon) architecture.}
\lemon follows the Message Passing Neural Network (MPNN) framework~\citep{gilmer2017neural}, with the key distinction that the \emph{update function} is $\mathcal{W}$-equivariant. We focus here on the core construction and defer preprocessing and further implementation details to \Cref{app:From attention maps to Graphs with Structured Features - Extended Section}. Let $\mathbf{h}^{(l)} \in \mathbb{R}^{|V| \times L \times H \times d}$ denote node features at layer $l$, $\mathbf{e} \in \mathbb{R}^{|V| \times |V| \times L \times H \times d}$ edge features, and $\mathcal{N}(n)$ the preceding-token neighbors of node $n$. Each edge feature $\mathbf{e}_{n_1,n_2}$ is initialized by applying a shared linear map to the corresponding attention score $\mathbf{A}_{n_1,n_2}$ independently across layers and heads. For node features, we combine self-attention scores with token probabilities, a model-agnostic signal that is also comparable across models. More specifically, we concatenate self-attention scores with token probabilities\footnote{Response tokens use their model probability; prompt tokens use $0$.} and map the resulting features to $\mathbb{R}^d$ with a shared linear layer.

\paragraph{\lemon's message passing.} Our \emph{Message-Passing} (MP) component builds on GINE~\citep{hu2019strategies}, with the key modification that its update function respects the group symmetry $\mathcal{W}$ derived in \Cref{subsec:attention-symmetries-and-equivariant-layers}. More specifically, at each layer $l$, an MPNN updates the node features according to,
\begin{equation}
    \mathbf{h}_n^{(l)}
    =
    \operatorname{up}_{\mathcal{W}}^{(l)}
    \bigg(
        (1+\epsilon^{(l)})\,\mathbf{h}_n^{(l-1)}
        +
        \sum_{n' \in \mathcal{N}(n)}
        \big(
            \mathbf{h}_{n'}^{(l-1)}
            +
            \mathbf{e}_{n',n}
        \big)
    \bigg),
    \qquad n \in V.
    \label{eq:eagn-gine-short}
\end{equation}
\vspace{-0.3cm}
where, \emph{crucially}, $\operatorname{up}_{\mathcal{W}}^{(l)} : \mathbb{R}^{L \times H \times d} \rightarrow \mathbb{R}^{L \times H \times d}$ is an update function that \emph{is $\mathcal{W}$-equivariant}.

\paragraph{Constructing the update function $\operatorname{up}_{\mathcal{W}}$.}
Building on the general characterization of \emph{linear} $\mathcal{S}_H\wr\mathcal{C}_L$-equivariant layers~\citep{wang2020equivariant}, we extend their construction to nonlinear neural networks, yielding a more general class of $\mathcal{W}$-equivariant maps. We define $f_{\mathcal{W}}:\mathbb{R}^{L\times H}\to\mathbb{R}^{L\times H}$ as follows:
\begin{align}
  \label{eq:w-equiv}
  f_\mathcal{W}(\mathbf{A}_{n_1, n_2}) &= f_{\mathcal{C}_L}(\mathbf{A}_{n_1, n_2}\vec{1}_H)\vec{1}_H^\top + f_{\mathcal{S}_H}^{\text{siamese}}(\mathbf{A}_{n_1, n_2}), \\   \label{eq:w-equiv2}
  \text{where} \quad f_{\mathcal{S}_H}^{\text{siamese}}(\mathbf{A}_{n_1, n_2})^l &\coloneqq f_{\mathcal{S}_H}(\mathbf{A}_{n_1, n_2}^l) \;\; \forall\, l \in [L],
\end{align}
where $\vec{1}_H \in \mathbb{R}^H$ is the all-ones vector and $f_{\mathcal{S}_H} : \mathbb{R}^{H} \rightarrow \mathbb{R}^{H}$ and $f_{\mathcal{C}_L} : \mathbb{R}^{L} \rightarrow \mathbb{R}^{L}$ are any functions equivariant to $\mathcal{S}_H$ and $\mathcal{C}_L$, respectively\footnote{Note that, for clarity, \Cref{eq:w-equiv,eq:w-equiv2} omits features, although they are included in the implementation}; See \Cref{app:W-vis} for an illustration.

\paragraph{Two instantiations of $\operatorname{up}_{\mathcal{W}}$.}
We consider two $\mathcal{W}$-equivariant variants of \lemon: \lemonW, which mixes information across layers and heads, and \lemonWpc, which applies the same update independently to each layer-head cell (see \Cref{app:From attention maps to Graphs with Structured Features - Extended Section}). This mirrors the difference between a full DeepSets~\citep{zaheer2017deep} equivariant layer and its element-wise term alone. 

After stacking several equivariant MPNN layers, we apply a \emph{causal cross-token transformer} across nodes independently for each layer-head coordinate (see \Cref{app:From attention maps to Graphs with Structured Features - Extended Section}). Finally, we apply a $\mathcal{W}$-invariant prediction head, operating at either the node or graph level depending on the granularity of the task.

\begin{wrapfigure}{r}{0.29\textwidth}
  \vspace{-10pt}
  \centering
  \resizebox{\linewidth}{!}{%
  \begin{tikzpicture}[>=Stealth, font=\small,
    node/.style={circle, draw=blue!55!black, fill=blue!10, minimum size=8.5mm, inner sep=0pt},
    cur/.style={circle, draw=blue!60!black, line width=1.3pt, fill=blue!25, minimum size=8.5mm, inner sep=0pt},
    ghost/.style={circle, draw=gray!60, dashed, fill=gray!4, minimum size=8.5mm, inner sep=0pt, text=gray!55},
    msg/.style={->, semithick, draw=cyan!60!blue},
    ghostmsg/.style={->, semithick, draw=gray!45, dashed},
    blocked/.style={->, semithick, draw=red!65!black, dashed},
    scorearr/.style={->, thick, draw=green!45!black},
  ]
    \fill[blue!6, rounded corners=9pt] (-0.62,-0.55) rectangle (4.22,0.55);
    \node[font=\scriptsize, text=blue!55!black, anchor=south west] at (-0.60,0.60)
      {visible to $\hat{y}_n$};

    \node[node]  (a)  at (0.00,0) {$1$};
    \node[font=\small, text=blue!45!black] at (1.15,0) {$\cdots$};
    \node[node]  (b)  at (2.30,0) {$n{-}1$};
    \node[cur]   (c)  at (3.60,0) {$n$};
    \node[ghost] (g1) at (4.90,0) {$n{+}1$};
    \node[ghost] (g2) at (6.20,0) {$n{+}2$};
    \node[font=\scriptsize, text=gray!60, anchor=north] at (5.55,-0.95)
      {not yet generated};

    \draw[msg] (a.south) to[bend right=34] (c.south west);
    \draw[msg] (a.south) to[bend right=30] (b.south west);
    \draw[msg] (b.south) to[bend right=30] (c.south);

    \draw[ghostmsg] (c.south)  to[bend right=28] (g1.south);
    \draw[ghostmsg] (c.south)  to[bend right=40] (g2.south);
    \draw[ghostmsg] (g1.south) to[bend right=28] (g2.south);

    \draw[blocked] (g1.north) to[bend right=40] (c.north east);
    \node[red!65!black, font=\footnotesize] at (4.28,0.80) {$\times$};

    \node[rounded corners=2.5pt, draw=green!45!black, fill=green!8,
          font=\small, inner sep=2.5pt] (y) at (3.60,1.45) {$\hat{y}_n$};
    \draw[scorearr] (c.north) -- (y.south);
    \node[font=\scriptsize, text=green!40!black, anchor=east] at (3.30,1.45)
      {scored online};

    \draw[->, gray!60, semithick] (-0.30,-1.70) --
      node[below, font=\scriptsize, text=gray!70] {generation order} (3.90,-1.70);
  \end{tikzpicture}%
  }
  \vspace{-20pt}
  \caption{\lemon is online.
  }
  \label{fig:eagn-online-main}
\end{wrapfigure}
\paragraph{\lemon is \emph{online}.} MP follows \emph{forward-only} graph edges, the cross-token transformer attends only to past positions, and the prediction head acts independently across tokens. Hence, a node prediction $\hat{y}_n$ depends only on tokens $m\leq n$ (see \Cref{fig:eagn-online-main}). This has two consequences: (i) \lemon can score tokens (or window of tokens) as they are generated, incrementally adding each new node at \emph{linear} cost in the prefix length rather than recomputing the full quadratic graph. 
(ii) A prefix receives the same score online as within the completed response, so offline training on full responses and online deployment are provably the same computation.
We note that \charm~\citep{frasca2026neural} also shares this online property.

\section{Experiments: Monitoring for LLM misbehaviors with \lemon}
\label{sec:experiments}

In this section, we present a comprehensive empirical evaluation of \lemon for monitoring LLM misbehaviors, focusing on two: hallucinations and jailbreaks. Our evaluation consists of 16 LLMs, spanning six architecture families, sizes from 135M to 13B, and release dates from 2023 to 2025, and a diverse set of 6 datasets. We use the ROC-AUC ($\times 100$) as our evaluation metric, consistent with prior work \citep{frasca2026neural,azaria2023internal,chuang2024lookback}. We report the main results here; additional results and experimental details are provided in \Cref{app:experiments-extended}.

\paragraph{Training setup.} Unless otherwise specified, we use a \emph{leave-one-out} (LOO) setup. Given a pool of $\mathcal{K}$ LLMs, we train on $\mathcal{K}-1$ models and evaluate on the remaining held-out model, rotating the split so that each LLM serves as the test target exactly once. Thus, \emph{our \lemon monitor is always evaluated on unseen LLMs}. 
Our LLM pool is the following $\mathcal{K}=6$ LLMs: \llamaseven, \llamathirteen \citep{touvron2023llama2}, \mistralseven\citep{jiang2023mistral}, \nemo \citep{sreenivas2024llm}, \qwenthree, and \qwenseven \citep{qwen2}. 

\paragraph{Methods in comparison.} We compare \lemon against the following task-agnostic, learnable baselines\footnote{All trained baselines follow the same cross-LLM transfer protocol as \lemon.}: (i)~\textbf{CHARM} \citep{frasca2026neural}, a GNN-based architecture that operates on attention graphs but \emph{does not} explicitly account for the equivariance properties considered in our approach; and (ii)~\textbf{Hidden-state probing} (HS), which trains a probe on the LLM's hidden representations.\footnote{We use hidden states from layer 24 of each LLM for probing. We pad the sequences to a fixed length to account for the different feature dimensions across LLMs.} We additionally compare against task-specific baselines, which we introduce separately in the corresponding experimental subsections.

\subsection{Hallucination Monitoring}\label{subsec:hallucination-monitoring}
For our hallucination experiments, we use the RAGTruth benchmark \citep{niu2024ragtruth}, which consists of 3 token-level tasks: (i)~\textbf{Data2Text} (D2T); (ii)~\textbf{Natural Questions} (NQ); and (iii)~\textbf{Summarization} (SUM). While RAGTruth provides labels for only a subset of the LLMs in our pool, we annotate the outputs of the remaining LLMs ourselves using \qwenjudge as the labeling tool; See \Cref{app:data-annotation} for further details. Finally, \lemon, as well as all baseline methods described below, are trained on the union of the three tasks, while evaluation results are reported separately for each.

\paragraph{Task-specific baselines.} We compare \lemon against the following task-specific baselines: (i)~\textbf{\lookback} (\lb) \citep{chuang2024lookback}, a heuristic that uses attention ratios to detect hallucinations, see \Cref{app:lookbacklens} for details; (ii)~a \textbf{probability-based baseline} (\probbase), which uses the LLM's output token probabilities to detect hallucinations; and (iii)~\textbf{\lookback{+}\emph{Alignment}} (\lbalign), which applies the \lookback baseline to attention features \emph{after} aligning them using our alignment method described in \Cref{app:alignment-methods}.

\begin{table}[t]
  \begin{minipage}[t]{0.52\textwidth}
    \centering
\captionof{table}{%
  \textbf{LOO results.}
  Each row is one held-out LLM; every value is the AUC averaged over the three tasks.
  Per-task breakdown: \Cref{tab:loo-zero-shot-full}.
  \textbf{Bold} = best per row; \underline{underline} = second best.
}
\label{tab:loo-zero-shot}
\scriptsize
\setlength{\tabcolsep}{3pt}
\resizebox{\linewidth}{!}{%
  \begin{tabular}{@{}l ccccc | cc@{}}
    \toprule
    \multirow{2}{*}{Held-out LLM}
      & \multirow{2}{*}{\probbase}
      & \multirow{2}{*}{\lb}
      & \multirow{2}{*}{\hs}
      & \multirow{2}{*}{\charm}
      & \multirow{2}{*}{\lbalign}
      & \multicolumn{2}{c}{\textbf{\lemon (Ours)}} \\
    \cmidrule(lr){7-8}
    & & & & & & $\mathcal{W}$ & $\mathcal{W}$-per-cell \\
    \midrule
    Llama-2-7B & 57.8 & 48.8 & 52.9 & 45.5 & 61.3 & \underline{71.4} & \textbf{74.9} \\
    Llama-2-13B & 52.3 & 53.9 & 52.5 & 57.3 & 54.2 & \underline{73.8} & \textbf{75.5} \\
    Mistral-7B & 67.6 & 49.5 & 51.0 & 42.2 & 57.5 & \underline{68.2} & \textbf{68.5} \\
    Mistral-Nemo & 58.3 & 52.4 & 50.3 & 44.2 & 60.6 & \underline{63.9} & \textbf{64.8} \\
    Qwen2.5-3B & 61.6 & 54.3 & 49.3 & 41.9 & 50.1 & \underline{66.0} & \textbf{66.2} \\
    Qwen2.5-7B & 61.1 & 50.0 & 52.2 & 57.8 & 58.7 & \underline{63.6} & \textbf{66.3} \\
    \midrule
    \multicolumn{1}{@{}l}{\textbf{Mean (18 shards)}}
      & 59.8 & 51.5 & 51.4 & 48.1 & 57.1 & \underline{67.8} & \textbf{69.4} \\
    \bottomrule
  \end{tabular}%
}%

  \end{minipage}\hfill
  \begin{minipage}[t]{0.46\textwidth}
    \centering
\captionof{table}{%
  \textbf{Monitoring for errors in the wild.} \textbf{Bold} = best
  per row; \underline{underline} = second best.
}
\label{tab:hallucination-monitoring-in-the-wild}
\scriptsize
\setlength{\tabcolsep}{4pt}
\resizebox{\linewidth}{!}{%
  \begin{tabular}{@{}ll c | c | cc@{}}
    \toprule
    \multirow{2}{*}{\textbf{Dataset}} & \multirow{2}{*}{\textbf{Target LLM}}
      & \multirow{2}{*}{Test size} & \multirow{2}{*}{\probbase}
      & \multicolumn{2}{c}{\textbf{\lemon (Ours)}} \\
    \cmidrule(lr){5-6}
    & & & & $\mathcal{W}$ & $\mathcal{W}$-per-cell \\
    \midrule
    \multirow{2}{*}{Mu-SHROOM} & Mistral-7B-v0.2 & 53 & 58.7 & \textbf{63.5{\scriptsize$\pm$1.7}} & \underline{62.7{\scriptsize$\pm$1.6}} \\
     & Pythia-Chat-7B & 54 & 61.8 & \underline{72.9{\scriptsize$\pm$4.7}} & \textbf{74.0{\scriptsize$\pm$6.9}} \\
    \cmidrule(lr){2-6}
    & \textbf{Mean (Mu-SHROOM)} & & 60.3 & \underline{68.2} & \textbf{68.4} \\
    \midrule
    \multirow{8}{*}{PsiloQA} & TinyLlama-1.1B & 100 & 50.1 & \textbf{75.8{\scriptsize$\pm$1.7}} & \underline{75.7{\scriptsize$\pm$1.8}} \\
     & Zephyr-7B-$\beta$ & 100 & 48.3 & \underline{64.2{\scriptsize$\pm$1.0}} & \textbf{65.2{\scriptsize$\pm$1.5}} \\
     & Phi-4-mini & 99 & \textbf{59.7} & \underline{56.5{\scriptsize$\pm$1.6}} & 55.1{\scriptsize$\pm$0.6} \\
     & Hermes-2-Mistral-7B & 100 & 54.2 & \textbf{64.1{\scriptsize$\pm$2.2}} & \underline{63.5{\scriptsize$\pm$1.8}} \\
     & Pythia-Chat-7B (v0.16) & 100 & 36.3 & \underline{75.8{\scriptsize$\pm$1.6}} & \textbf{77.0{\scriptsize$\pm$0.4}} \\
     & SmolLM2-1.7B & 99 & 57.7 & \textbf{79.8{\scriptsize$\pm$1.1}} & \underline{79.4{\scriptsize$\pm$2.3}} \\
     & SmolLM2-360M & 100 & 56.1 & \underline{63.2{\scriptsize$\pm$2.4}} & \textbf{63.7{\scriptsize$\pm$1.5}} \\
     & SmolLM2-135M & 100 & 54.8 & \underline{57.7{\scriptsize$\pm$2.9}} & \textbf{60.3{\scriptsize$\pm$2.9}} \\
    \cmidrule(lr){2-6}
    & \textbf{Mean (PsiloQA)} & & 52.1 & \underline{67.1} & \textbf{67.5} \\
    \midrule
    \multicolumn{3}{@{}l}{\textbf{Mean (all 10 shards)}} & 53.8 & \underline{67.4} & \textbf{67.7} \\
    \bottomrule
  \end{tabular}%
}%
\vspace{-0.6cm}

  \end{minipage}
\end{table}
\paragraph{LOO results.} The results are presented in \Cref{tab:loo-zero-shot}. Both \lemon variants substantially outperform all baselines. \lemon achieves the best performance in 16 of 18 model-benchmark combinations, and on average, \lemonWpc improves over the strongest baseline, \probbase, by approximately 10 ROC-AUC points. We also note that \lemonWpc slightly outperforms \lemonW, suggesting that layer-head mixing, even when performed in a manner equivariant to $\mathcal{W}$, is not always beneficial. The learnable baselines perform around chance in the LOO setting, highlighting the importance of accounting for cross-model symmetries when generalizing to unseen models. We complement these results with two additional analyses. First, in \Cref{app:few-shot-adaptation}, we show that \lemon can be adapted to the held-out LLM using only a small number of labeled examples, yielding further improvements over its zero-shot performance. Second, in \Cref{app:LOI-Results}, we consider a \emph{substantially more challenging setting} in which \lemon is trained on \emph{a single LLM} (i.e., $\mathcal{K}=1$) and evaluated on the remaining five. Despite the substantially reduced training diversity relative to the LOO setting, \lemon continues to outperform all baselines by a large margin.

\paragraph{Estimating errors in the wild.}
To further simulate a real-world deployment of \lemon, we evaluate its monitoring performance \emph{in the wild} --- on both new datasets and new LLMs that \emph{were unseen during training}. Specifically, we use the datasets \textbf{Mu-SHROOM}\citep{vazquez2025semeval} and \textbf{PsiloQA}\citep{rykov2025models}.\footnote{Both datasets provide character-level span annotations rather than token-level labels; we mark a token as positive if it overlaps with an annotated span.} Both datasets consist of open-domain QA, where the target LLM answers solely from its parametric knowledge. This setting differs substantially from RAGTruth's context-grounded generation task. We evaluate on the official held-out \emph{test} split of each dataset using 10 target LLMs spanning five different architecture families. 
To evaluate \lemon's ability to estimate errors in the wild, we assess the six trained \lemon models for each variant used in \Cref{tab:loo-zero-shot} (see the last two columns). Specifically, we evaluate each model on the 10 shards described above and report the mean$\pm$std across the six models for each variant. We compare against the strongest baseline from \Cref{tab:loo-zero-shot}, namely \probbase. Both \lemon variants transfer essentially without degradation: averaged over the 10 shards, \lemonW reaches 67.4 and \lemonWpc 67.7. which is roughly 14 AUC points above the strongest baseline, \probbase, which scores 53.8. Finally, we note that the spread across the six checkpoints is small (median std $1.7$): the choice of training pool barely matters, so any single monitor could be deployed as-is and perform well.

\paragraph{Selective generation.} Due to space limitations, we defer this experiment to \Cref{subsec:selective-generation} and summarize it here. Because \lemon is causal, it can score a response while it is being generated. We use it as an online confidence signal that halts generation once a hallucination becomes likely, and serve only the preceding prefix. Sweeping the halting threshold traces a \emph{risk--coverage curve}~\cite{el2010foundations}, which trades the fraction of content served against the hallucination rate of what is served (formalized in \Cref{app:selective-generation}). \lemon attains the lowest area under this curve (AURC) in 17 of 18 shards, averaging $8.9$ (\lemonWpc) and $9.1$ (\lemonW) against $12.3$ for \probbase.

\newsavebox{\jbtab}
\savebox{\jbtab}{%
  \scriptsize
  \setlength{\tabcolsep}{2.5pt}%
  \begin{tabular}{@{}l c c c | c c@{}}
    \toprule
    \multirow{2}{*}{\textbf{Left-out}}
      & \multirow{2}{*}{Attn${\to}$Sys}
      & \multirow{2}{*}{\charm}
      & \multirow{2}{*}{\hs}
      & \multicolumn{2}{c@{}}{\textbf{\lemon (Ours)}} \\
    \cmidrule(l){5-6}
      & & & & $\mathcal{W}$ & $\mathcal{W}$-pc \\
    \midrule
    Llama-2-7B   & 59.1             & 30.4          & \underline{73.1} & \textbf{75.1}    & 62.6             \\
    Llama-2-13B  & 47.8             & \textbf{73.7} & 26.4             & \underline{69.5} & 66.5             \\
    Mistral-7B   & \underline{61.5} & 53.9          & 50.9             & \textbf{67.5}    & 59.2             \\
    Mistral-Nemo & 47.3             & \textbf{58.3} & 48.7             & 55.0             & \underline{57.3} \\
    Qwen2.5-3B   & \textbf{67.6}    & 38.5          & 45.7             & 63.9             & \underline{64.8} \\
    Qwen2.5-7B   & \underline{67.8} & 46.1          & 38.6             & 59.3             & \textbf{69.1}    \\
    \midrule
    \textbf{Mean} & 58.5 & 50.1 & 47.2 & \textbf{65.0} & \underline{63.3} \\
    \bottomrule
  \end{tabular}}
\begin{wraptable}{r}{\wd\jbtab}
  \vspace{-\intextsep}
  \centering
  \captionof{table}{%
    \textbf{Jailbreak monitoring results.}
    \textbf{Bold} = best; \underline{underline} = second best.
  }
  \label{tab:jailbreak-redbench}
  \usebox{\jbtab}
  \vspace{-\intextsep}
\end{wraptable}

\subsection{Jailbreak Monitoring}\label{subsec:jailbreak-monitoring}
\paragraph{Setup.} We use the \redbench~\citep{dang2026redbench} dataset, appending a \emph{single universal} GCG suffix~\citep{zou2023universal} to every harmful request, taken from HarmBench~\citep{mazeika2024harmbench}. The suffix is fixed across every request and every monitored LLM; see \Cref{app:datasets} for further details.
Each of the same $\mathcal{K}=6$ LLMs generates a response greedily, whose attack success is scored by \strongreject~\citep{souly2024strongreject} on a scale from $0$ to $1$; we binarize its label at $0.5$. Unlike hallucination, which is labeled at the token level, this task has a single binary label for the entire graph, so \lemon is read out at the final prompt token.
Here, the monitor observes only the \emph{prompt}, predicting attack success \emph{before} any response tokens are generated. We use $3000$ prompts per LLM, split $1800/450/750$ into train/validation/test at the prompt level and shared across all six LLMs.

\paragraph{A training-free baseline.} Beyond the learning-based baselines introduced in \Cref{sec:experiments}, motivated by previous works~\citep{hung2025attention,ben2026universal} we use the following baseline, which measures how much the final prompt token attends to instruction tokens $S$ (which may include the system-prompt) averaging across tokens, layers, and heads: $\mathrm{AttnSys}(\mathbf{x})=\frac{1}{LH}\sum_{l,h}\frac{1}{|S|}\sum_{j\in S}\mathbf{A}^{l,h}_{N,j}$. Lower values indicate weaker attention to the instruction/system prompt, so we use \(-\mathrm{AttnSys}\) as the jailbreak predictor. The results are summarized in \Cref{tab:jailbreak-redbench}. On average, both \lemon variants outperform all baselines: \lemonW improves over the strongest baseline by 6.5 AUC points, while \lemonWpc improves by 4.8 points. We note that in this case \lemonW performs better than \lemonWpc by almost 2 AUC units -- which suggests that sharing information across layers and heads is beneficial for this task. Moreover, across all six LLMs, \lemon consistently ranks either first or second.

\subsection{Ablation Studies}
\label{subsec:ablation-studies}
To ablate \lemon, we compare the \hlt{blue!8}{main variants} with four ablations:
(i) \lemon with \hlt{green!12}{no cross-token mixing} (no-cross), measuring the importance of sequence modeling; (ii) \hlt{violet!10}{non-equivariant} \lemon (Dense), which mixes information across layers and heads in a non-equivariant manner\footnote{\lemon (Dense) replaces the $\mathcal{W}$-equivariant update $    \operatorname{up}_{\mathcal{W}}^{(l)}$ of \Cref{eq:eagn-gine-short} by plain MLPs.}; (iii) \lemon with \hlt{red!10}{no graph structure} (without graph), using only probabilities as node features; and (iv) \lemon with \hlt{orange!15}{no probability} (no prob) node features. The results are in \Cref{tab:ablation-xattn-w8-andpool}. We observe that the \hlt{blue!8}{main variants}, \lemonW\ and \lemonWpc, outperform all ablations by a large margin, except \lemon (no cross) variant, which, importantly, is still invariant to $\mathcal{W}$, underscoring the importance of our derivation in \Cref{subsec:attention-symmetries-and-equivariant-layers}.
The \hlt{violet!10}{non-equivariant} variant performs worst, highlighting the importance of equivariance for cross-LLM transfer.

\subsection{Run-Time Analysis}
We report \lemon's online monitoring cost in \Cref{fig:eagn-latency}.
We compare an unmonitored decode step against \emph{all the three
terms} monitoring adds: (i) extracting attention (which forces the eager kernel in place
of 
\textsc{sdpa},
and is paid at every token), (ii) updating the graph, and (iii) running \lemon as the critic. We report this on the \emph{largest LLM} in our pool, \llamathirteen, on the first sample of the RAGTruth NQ shard. \emph{Monitoring is extremely cheap}: on average it adds $+8\%$ with \lemonW and $+6\%$ with \lemonWpc on \llamathirteen. Notably, the critic itself is the smallest of the three terms --- $0.3$ of the $2.6$\,ms added by \lemonWpc --- so the execution of the equivariant architecture is negligible, and what remains is dominated by attention extraction and graph maintenance, both of which could be further optimized with tailored hardware kernels. 

\begin{figure}[t]
  \begin{minipage}[t]{0.58\textwidth}
    \vspace{0pt}
    \providecommand{\lemonWpcshort}{\textbf{LEMON-}$\boldsymbol{\mathcal{W}}$\textbf{-pc}\xspace}
\centering
\captionof{table}{%
  \textbf{\lemon ablations in the LOO setup.} Mean OOD ROC-AUC over the three tasks per
  held-out LLM. \textbf{Bold} = best in column, \underline{underline} = second best.}
\label{tab:ablation-xattn-w8-andpool}
\small
\setlength{\tabcolsep}{3pt}
\resizebox{\linewidth}{!}{%
  \begin{tabular}{l cccccc | c}
    \toprule
    & \multicolumn{6}{c}{Held-out LLM (mean over 3 tasks)} & \\
    \cmidrule(lr){2-7}
    Variant
      & L2-7B & L2-13B & M-7B & Nemo & Q-3B & Q-7B
      & Mean \\
    \midrule
    \rowcolor{blue!8}
    \lemonW
      & 71.4 & 73.8 & 68.2 & 63.9 & \underline{66.0} & 63.6 & 67.8 \\
    \rowcolor{blue!8}
    \lemonWpcshort
      & \textbf{74.9} & \textbf{75.5} & 68.5 & \underline{64.8} & \textbf{66.2} & \textbf{66.3} & \textbf{69.4} \\
    \midrule
    \rowcolor{orange!15}
    \lemonW (no prob)
      & 69.6 & 66.5 & 68.2 & 62.2 & 61.4 & 59.6 & 64.6 \\
    \rowcolor{orange!15}
    \lemonWpcshort (no prob)
      & 70.2 & 67.2 & 66.6 & 63.4 & 65.6 & 60.5 & 65.6 \\
    \midrule
    \rowcolor{green!12}
    \lemonW (no cross)
      & 72.0 & \underline{75.1} & \textbf{68.7} & 63.1 & 63.9 & 64.9 & 68.0 \\
    \rowcolor{green!12}
    \lemonWpcshort (no cross)
      & \underline{72.9} & 74.8 & \underline{68.6} & \textbf{65.1} & \underline{66.0} & \underline{66.2} & \underline{68.9} \\
    \midrule
    \rowcolor{violet!10}
    \lemon (Dense)
      & 56.5 & 67.4 & 58.3 & 59.7 & 51.1 & 59.1 & 58.7 \\
    \midrule
    \rowcolor{red!10}
    \lemon (without graph)
      & 64.7 & 67.1 & 67.1 & 61.9 & 63.0 & 63.0 & 64.5 \\
    \bottomrule
  \end{tabular}%
}
\vspace{-0.6cm}
  \end{minipage}\hfill
  \begin{minipage}[t]{0.40\textwidth}
    \vspace{0pt}
    \input{figures/fig_eagn_latency}
  \end{minipage}
\end{figure}

\section{Conclusion}

We introduce \lemon, a learnable attention-based monitor that transfers to unseen decoder-only LLMs without target-model supervision. Our monitor, \lemon, is based on GNNs and is specifically designed to account for symmetries that may arise across attention heads and layers of different LLMs, which is precisely what enables it to generalize zero-shot in the wild. Importantly, \lemon is lightweight, adding less than 10\% monitoring overhead, and operates online: it can flag misbehavior during the generation process itself, without requiring the full response to be completed. 

Many directions remain open. \lemon can be evaluated on additional tasks, such as machine-generated text detection, memorization, and more. 
The application of selective generation with \lemon can be studied further, as can its use with other, model-specific monitors.
Extending \lemon to multimodal models, such as vision-language models, is an interesting avenue for future research. Finally, using \lemon as an interpretability tool -- to automatically find important layer-head coordinates, is interesting.

\arxivonly{
\subsection*{Acknowledgments}

F.F.\ conducted this work supported by an Aly Kaufman Post-Doctoral Fellowship. G.B.\ is supported by the Jacobs Qualcomm PhD Fellowship. H.M.\ is supported by the Israel Science Foundation through a personal grant (ISF 264/23) and an equipment grant (ISF 532/23), and by the Career Advancement Chairs in Artificial Intelligence -- Schmidt Futures.
}

\subsection*{Reproducibility statement}

The \lemon architecture is described in \Cref{sec:Extracting Universal Attention Representations} and detailed in \Cref{app:eagn-detailed,app:From attention maps to Graphs with Structured Features - Extended Section}. Training details, the LLMs, and the datasets and splits are given in \Cref{app:experiments-extended,app:llms-used,app:datasets}; our annotation procedure is in \Cref{app:data-annotation}, and the selective-generation metrics are defined in \Cref{app:selective-generation}. We will release our code, trained checkpoints, and annotations upon publication.

\bibliography{references}
\bibliographystyle{authoryear}

\appendix

\section*{Appendix roadmap}
The appendix is organized as follows.
\begin{itemize}
  \item \Cref{app:motivating-example}, \textit{Motivating Example: Full Query and Responses.} This section provides further details on the motivating example introduced in \Cref{sec:Extracting Universal Attention Representations}.
  \item \Cref{app:lookbacklens}, \textit{LookbackLens Features.} This section describes the \lookback features used in our attention-ratio baselines.
  \item \Cref{app:architecture-visualization}, \textit{Architecture Visualization.} This section provides visualizations of architectural components that we describe in the main text.
  \begin{itemize}
    \item \Cref{app:W-vis}, \textit{$\mathcal{W}$-equivariant layer.}
    \item \Cref{app:EAGM-vis}, \textit{\lemon architecture.}
  \end{itemize}
  \item \Cref{app:From attention maps to Graphs with Structured Features - Extended Section}, \textit{From attention maps to Graphs with Structured Features - Extended Section.} This section describes how we cast the attention map as an attributed directed graph, the preprocessing pipeline, and the LANGUAGE-MODEL EQUIVARIANT MONITORING NETWORK  (\lemon) architecture.
  \begin{itemize}
    \item \Cref{app:eagn-detailed}, \textit{\textbf{\underline{L}}anguage-model \textbf{\underline{E}}quivariant \textbf{\underline{MO}}nitoring \textbf{\underline{N}}etwork (\lemon) -- A Detailed Description.}
  \end{itemize}
  \item \Cref{app:alignment-methods}, \textit{Alignment Methods.} This section provides the full details of the head alignment methods we use in our experiments.
  \item \Cref{app:data-annotation}, \textit{Dataset Annotation.} This section documents how the RAGTruth-native annotations are used and how the three additional LLMs were annotated, together with a manual audit of the resulting labels.
  \item \Cref{app:experiments-extended}, \textit{Experiments: Monitoring for LLM misbehaviors with \lemon --- Extended Section.} 
  \begin{itemize}
    \item \Cref{app:baselines}, \textit{Baselines and Ablations: Implementation Details.}
    \item \Cref{app:llms-used}, \textit{LLMs Used In Our Experiments.}
    \item \Cref{app:datasets}, \textit{Datasets Used In Our Experiments.}
    \item \Cref{app:loo-per-task}, \textit{LOO results --- Extended results.}
    \item \Cref{app:selective-generation-full}, \textit{Selective Generation.}
    \item \Cref{app:few-shot-adaptation}, \textit{Few-shot adaptation.}
    \item \Cref{app:LOI-Results}, \textit{LOI Results.}
  \end{itemize}
  \item \Cref{app:selective-generation}, \textit{Selective Generation: Formal Framework.} This section formalizes the selective-generation setting of \Cref{subsec:selective-generation}: the selective generator, the served output, and the coverage, token-coverage, selective-risk, and token-savings estimators, together with the construction of a selective generator from a confidence function.
  \begin{itemize}
    \item \Cref{app:selective-generation-construct}, \textit{Constructing Selective Generators and the Risk--Coverage Curve.}
  \end{itemize}
\end{itemize}

\clearpage
\section{Motivating Example: Full Query and Responses}
\label{app:motivating-example}

\newcommand{\hspan}[1]{{\setlength{\fboxsep}{2.5pt}\setlength{\fboxrule}{0.6pt}%
  \fcolorbox{red!65!black}{red!6}{#1}}}
  
  Here we provide the full details regarding the motivating example in
\Cref{sec:Extracting Universal Attention Representations}. The example is a single data-to-text
instance from RAGTruth (sample~147), answered by both Llama-2-7B and Mistral-7B. The
task is to write an overview of a business (``The Lark'', a Santa Barbara restaurant)
from a structured JSON record, using \emph{only} the provided data. In the responses
below, \hspan{all spans annotated as hallucinated by RAGTruth are boxed}, and the
specific token analyzed in \Cref{fig:head-perm-motivation} --- \textbf{Sunday} --- is shown in \textbf{bold} within its
box. Both models hallucinate this token (\textbf{Sunday}), in different sentences and at
different output positions.

\paragraph{Why \texttt{"sunday"} is a hallucination.}
The structured data lists per-day opening hours that vary across the week
(Monday closed), yet both models collapse them into a fabricated
open-every-day claim: ``open \emph{Monday through \textbf{Sunday}} from 17:00
to 22:00'' (Llama-2-7B) and ``open \emph{from Monday to \textbf{Sunday}} from
5 pm to midnight'' (Mistral-7B). RAGTruth annotates this claim as a
hallucinated span in each response, and \texttt{"sunday"} is hallucinated as
part of the invented range. This provides a clean choice for comparison: the error is localized, appears in both models within the same fabricated claim, and the token \texttt{"sunday"} is identical across models, allowing a direct comparison of their attention maps. See \Cref{tab:tokenizer-days} for the tokenization of all day names in both models.

\begin{tcolorbox}[breakable,colback=gray!5,colframe=gray!50,title=Query]
\textbf{Instruction.} Write an objective overview about the following local business
based only on the provided structured data in the JSON format. You should include
details and cover the information mentioned in the customers' reviews. The overview
should be 100--200 words. Don't make up information.

\medskip
\textbf{Context (structured data, reviews abbreviated).}
\texttt{\{'name': 'The Lark', 'address': '131 Anacapa St, Ste A', 'city': 'Santa
Barbara', 'state': 'CA', 'categories': 'Nightlife, Bars, Cocktail Bars, Food, American
(New), Desserts, Restaurants', 'hours': \{'Monday': '0:0-0:0', 'Tuesday': '17:0-21:0',
'Wednesday': '17:0-21:0', 'Thursday': '17:0-21:0', 'Friday': '17:0-22:0', 'Saturday':
'17:0-22:0', 'Sunday': '17:0-21:0'\}, 'attributes': \{'BusinessParking': \{'street':
True, \ldots\}, 'RestaurantsReservations': True, 'OutdoorSeating': True, 'WiFi':
'free', 'RestaurantsTakeOut': False, 'RestaurantsGoodForGroups': True, 'Music': False,
'Ambience': \{'trendy': True, 'classy': True, \ldots\}\}, 'business\_stars': 4.0,
'review\_info': [\{'review\_stars': 5.0, 'review\_text': 'Truly a delicious "Shared
Dining Experience" \ldots Crispy Brussel Sprouts \ldots Highly recommend making
reservations \ldots'\}, \{'review\_stars': 3.0, 'review\_text': 'The food was good to
great, where the experience fell apart was in the service \ldots'\}, \{'review\_stars':
5.0, 'review\_text': 'My partner and I celebrated our anniversary \ldots our waiter
Oscar was friendly \ldots'\}]\}}
\end{tcolorbox}

\begin{tcolorbox}[breakable,colback=blue!3,colframe=blue!40,title=Llama-2-7B response]
Based on the provided structured data, here is an objective overview of The Lark:

The Lark is a popular nightlife spot located in the heart of Santa Barbara's Funk
Zone. With categories such as Nightlife, Bars, Cocktail Bars, Food, American (New),
Desserts, and Restaurants, The Lark offers a unique shared dining experience for
groups of all sizes. Their hours of operation are \hspan{Monday through \textbf{Sunday}
from 17:00 to 22:00}, \hspan{with validated parking available} off-street and street
parking nearby.

According to customer reviews, the quality of the food at The Lark is consistently
high, with standout dishes including crispy Brussels sprouts, winter beets marinated
in balsamic vinegar, caramelized cauliflower, cast iron Morro Bay black cod, creekstone
prime hanger steak, and buttermilk fried chicken. Wine enthusiasts will appreciate the
extensive selection of wine-by-the-glass options, including Grimm's Bluff, Happy
Canyon, and Rose of Cabernet Sauvignon.

However, some customers have reported issues with the service at The Lark, including a
lack of attentiveness from servers and confusion with regard to reservations. Despite
these minor drawbacks, the majority of customers have praised The Lark for its cozy
atmosphere, excellent food, and exceptional service. One reviewer even noted that their
server, Laurel, went above and beyond to ensure a special night, earning a 5-star
rating for service. Overall, The Lark appears to be a top choice for dinner and drinks
in Santa Barbara, offering a memorable dining experience for visitors and locals alike.
\end{tcolorbox}

\begin{tcolorbox}[breakable,colback=red!3,colframe=red!40,title=Mistral-7B response]
The Lark is a restaurant located in the heart of the Santa Barbara Funk Zone at 131
Anacapa St, Ste A. The restaurant offers a variety of food options including American
cuisine, desserts, and cocktails. The restaurant has outdoor seating and serves WiFi.
It is classified under the categories of Nightlife, Bars, Cocktail Bars, Food, American
(New), Desserts, Restaurants, and is \hspan{open from Monday to \textbf{Sunday} from 5 pm
to midnight}. Based on customer reviews, the restaurant appears to have a good
reputation for its food quality and the friendliness of its staff. However, some
customers have reported issues with the service and the attentiveness of the staff.
Overall, The Lark seems like a good place to visit \hspan{for a romantic dinner} or
\hspan{a casual evening} out with friends.
\end{tcolorbox}

\begin{table}[H]
  \centering
  \begin{tabular}{lll}
    \toprule
    Day & \llamaseven & \mistralseven \\
    \midrule
    Monday    & \texttt{Monday} & \texttt{Monday} \\
    Tuesday   & \texttt{T|ues|day} & \texttt{Tuesday} \\
    Wednesday & \texttt{Wed|nes|day} & \texttt{Wednesday} \\
    Thursday  & \texttt{Th|urs|day} & \texttt{Thursday} \\
    Friday    & \texttt{Friday} & \texttt{Friday} \\
    Saturday  & \texttt{Saturday} & \texttt{Saturday} \\
    \textbf{Sunday} & \texttt{Sunday} & \texttt{Sunday} \\
    \bottomrule
  \end{tabular}
  \caption{Subword tokenization of the seven day names for \llamaseven and \mistralseven.}
  \label{tab:tokenizer-days}
\end{table}

\paragraph{Generality of \Cref{fig:head-perm-motivation}.}
The comparison of \Cref{fig:head-perm-motivation} is not specific to the selected
token. We repeat it for \emph{every} hallucinated content word the two models share as follows:
for each RAGTruth test sample answered by both LLMs, we take every word (alphabetic,
longer than three characters) that both tokenizers represent as a single token and
that both models' annotations flag as hallucinated, and compute the Pearson
correlation between the two models' $32\times32$ \lookback maps at that token. Over all, we have $104$ such tokens ($44$ question answering, $15$ summarization, $45$
data-to-text), the correlation \emph{never} exceeds $r=0.19$ (mean $0.07$, median $0.07$,
minimum $-0.03$). The misalignment shown in \Cref{fig:head-perm-motivation} is thus
the rule rather than an artifact of the chosen example.

\section{LookbackLens Features}
\label{app:lookbacklens}

Here we provide further details on \lookback (\lb) features~\citep{chuang2024lookback}. \citet{chuang2024lookback} showed that how much the model ``looks back'' at the prompt and context when generating each token is predictive of contextual hallucinations: if a response token is produced while attending mainly to earlier response tokens rather than to the input, it is more likely to be a hallucination.
They define a simple heuristic per (layer, head) and per response token.
Consider the $t$-th response token. It is produced by the query at the preceding position, whose attention distribution at layer $\ell$ and head $h$ we denote $\mathbf{a} \coloneqq \mathbf{A}^{\ell,h}_{n_p+t-1,\,\cdot}$, where $n_p$ is the number of prompt/context tokens; it is supported on the $n_p$ prompt tokens and the $t-1$ response tokens generated so far.
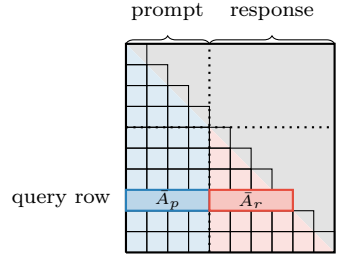
\begin{figure}[H]
  \centering
  \begin{tikzpicture}[every node/.style={font=\small\sffamily}, scale=0.92]
    \def\s{0.3}
    \def\np{4}
    \fill[gray!22] (0, 10*\s) -- (10*\s, 10*\s) -- (10*\s, 0) -- (0, 10*\s);
    \begin{scope}
      \clip (0, 10*\s) -- (0, 0) -- (10*\s, 0) -- (0, 10*\s);
      \fill[promptframe!16] (0, 0) rectangle (\np*\s, 10*\s);
      \fill[responseframe!16] (\np*\s, 0) rectangle (10*\s, 10*\s);
    \end{scope}
    \draw[thin] (0, 9*\s) rectangle (\s, 10*\s);
    \draw[thin] (0, 8*\s) rectangle (\s, 9*\s);
    \draw[thin] (\s, 8*\s) rectangle (2*\s, 9*\s);
    \draw[thin] (0, 7*\s) rectangle (\s, 8*\s);
    \draw[thin] (\s, 7*\s) rectangle (2*\s, 8*\s);
    \draw[thin] (2*\s, 7*\s) rectangle (3*\s, 8*\s);
    \draw[thin] (0, 6*\s) rectangle (\s, 7*\s);
    \draw[thin] (\s, 6*\s) rectangle (2*\s, 7*\s);
    \draw[thin] (2*\s, 6*\s) rectangle (3*\s, 7*\s);
    \draw[thin] (3*\s, 6*\s) rectangle (4*\s, 7*\s);
    \draw[thin] (0, 5*\s) rectangle (\s, 6*\s);
    \draw[thin] (\s, 5*\s) rectangle (2*\s, 6*\s);
    \draw[thin] (2*\s, 5*\s) rectangle (3*\s, 6*\s);
    \draw[thin] (3*\s, 5*\s) rectangle (4*\s, 6*\s);
    \draw[thin] (4*\s, 5*\s) rectangle (5*\s, 6*\s);
    \draw[thin] (0, 4*\s) rectangle (\s, 5*\s);
    \draw[thin] (\s, 4*\s) rectangle (2*\s, 5*\s);
    \draw[thin] (2*\s, 4*\s) rectangle (3*\s, 5*\s);
    \draw[thin] (3*\s, 4*\s) rectangle (4*\s, 5*\s);
    \draw[thin] (4*\s, 4*\s) rectangle (5*\s, 5*\s);
    \draw[thin] (5*\s, 4*\s) rectangle (6*\s, 5*\s);
    \draw[thin] (0, 3*\s) rectangle (\s, 4*\s);
    \draw[thin] (\s, 3*\s) rectangle (2*\s, 4*\s);
    \draw[thin] (2*\s, 3*\s) rectangle (3*\s, 4*\s);
    \draw[thin] (3*\s, 3*\s) rectangle (4*\s, 4*\s);
    \draw[thin] (4*\s, 3*\s) rectangle (5*\s, 4*\s);
    \draw[thin] (5*\s, 3*\s) rectangle (6*\s, 4*\s);
    \draw[thin] (6*\s, 3*\s) rectangle (7*\s, 4*\s);
    \draw[thin] (0, 2*\s) rectangle (\s, 3*\s);
    \draw[thin] (\s, 2*\s) rectangle (2*\s, 3*\s);
    \draw[thin] (2*\s, 2*\s) rectangle (3*\s, 3*\s);
    \draw[thin] (3*\s, 2*\s) rectangle (4*\s, 3*\s);
    \draw[thin] (4*\s, 2*\s) rectangle (5*\s, 3*\s);
    \draw[thin] (5*\s, 2*\s) rectangle (6*\s, 3*\s);
    \draw[thin] (6*\s, 2*\s) rectangle (7*\s, 3*\s);
    \draw[thin] (7*\s, 2*\s) rectangle (8*\s, 3*\s);
    \draw[thin] (0, \s) rectangle (\s, 2*\s);
    \draw[thin] (\s, \s) rectangle (2*\s, 2*\s);
    \draw[thin] (2*\s, \s) rectangle (3*\s, 2*\s);
    \draw[thin] (3*\s, \s) rectangle (4*\s, 2*\s);
    \draw[thin] (4*\s, \s) rectangle (5*\s, 2*\s);
    \draw[thin] (5*\s, \s) rectangle (6*\s, 2*\s);
    \draw[thin] (6*\s, \s) rectangle (7*\s, 2*\s);
    \draw[thin] (7*\s, \s) rectangle (8*\s, 2*\s);
    \draw[thin] (8*\s, \s) rectangle (9*\s, 2*\s);
    \draw[thin] (0, 0) rectangle (\s, \s);
    \draw[thin] (\s, 0) rectangle (2*\s, \s);
    \draw[thin] (2*\s, 0) rectangle (3*\s, \s);
    \draw[thin] (3*\s, 0) rectangle (4*\s, \s);
    \draw[thin] (4*\s, 0) rectangle (5*\s, \s);
    \draw[thin] (5*\s, 0) rectangle (6*\s, \s);
    \draw[thin] (6*\s, 0) rectangle (7*\s, \s);
    \draw[thin] (7*\s, 0) rectangle (8*\s, \s);
    \draw[thin] (8*\s, 0) rectangle (9*\s, \s);
    \draw[thin] (9*\s, 0) rectangle (10*\s, \s);
    \draw[thick] (0, 0) -- (10*\s, 0) -- (10*\s, 10*\s) -- (0, 10*\s) -- (0, 0);
    \draw[decorate, decoration={brace, amplitude=3pt}] (0, 10*\s) -- (\np*\s, 10*\s);
    \node[above, font=\scriptsize] at (2*\s, 10*\s+0.2) {prompt};
    \draw[decorate, decoration={brace, amplitude=3pt}] (\np*\s, 10*\s) -- (10*\s, 10*\s);
    \node[above, font=\scriptsize] at (7*\s, 10*\s+0.2) {response};
    \draw[dotted, thick] (\np*\s, 0) -- (\np*\s, 10*\s);
    \draw[dotted, thick] (0, 6*\s) -- (10*\s, 6*\s);
    \draw[decorate, decoration={brace, amplitude=2pt, mirror}] (0, 2*\s) -- (0, 3*\s);
    \node[left=2pt, font=\scriptsize] at (0, 2.5*\s) {query row};
    \fill[promptframe!32] (0, 2*\s) rectangle (\np*\s, 3*\s);
    \draw[promptframe, thick] (0, 2*\s) rectangle (\np*\s, 3*\s);
    \fill[responseframe!32] (\np*\s, 2*\s) rectangle (8*\s, 3*\s);
    \draw[responseframe, thick] (\np*\s, 2*\s) rectangle (8*\s, 3*\s);
    \node[font=\tiny] at (2*\s, 2.5*\s) {$\bar{A}_p$};
    \node[font=\tiny] at (6*\s, 2.5*\s) {$\bar{A}_r$};
  \end{tikzpicture}
  \caption{\lookback: average attention from the query generating the $t$-th response token to the prompt ($\bar{A}_p$) vs.\ the earlier response ($\bar{A}_r$). This example assumes a single layer and a single attention head.}
  \label{fig:lookback-heuristic}
\end{figure}
\lb compares the average attention mass placed on the two segments,
\begin{align*}
  \bar{A}_p^{\ell,h}(t) &= \frac{1}{n_p} \sum_{i=1}^{n_p} a_i, &
  \bar{A}_r^{\ell,h}(t) &= \frac{1}{t-1} \sum_{i=n_p+1}^{n_p+t-1} a_i,
\end{align*}
with $\bar{A}_r^{\ell,h}(1) \coloneqq 0$. Their heuristic is the \emph{prompt-attention ratio}
\begin{align*}
  \rho^{\ell,h}(t) = \frac{\bar{A}_p^{\ell,h}(t)}{\bar{A}_p^{\ell,h}(t) + \bar{A}_r^{\ell,h}(t)} \in [0,1];
\end{align*}
low $\rho$ means the model attended mostly to the response so far, and is taken as a signal of possible hallucination. See \Cref{fig:lookback-heuristic} for an example of a single layer and head.
In our baselines, we flatten $\{\rho^{\ell,h}(t)\}_{\ell,h}$, zero-pad to $L_{\max} H_{\max}$ where $L_{\max}$ and $H_{\max}$ are the maximum number of layers and heads in our LLM pool, and train logistic regression.

\clearpage
\section{Architectures Visualization}
\label{app:architecture-visualization}

\subsection{$\mathcal{W}$-equivariant layer.}
\label{app:W-vis}
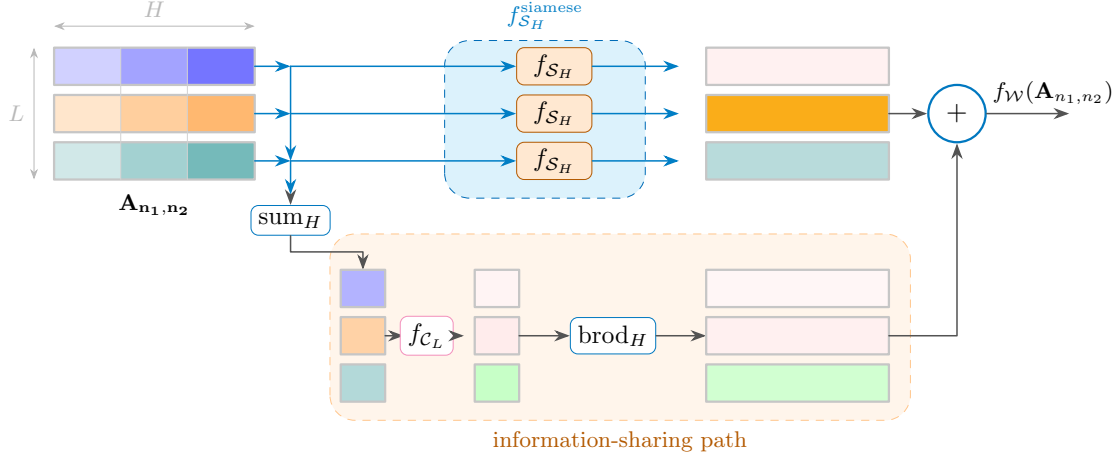
\begin{figure}[htbp]
  \centering
  \resizebox{0.9\linewidth}{!}{%
  \begin{tikzpicture}[
    >=Stealth,
    line cap=round,
    line join=round,
    arr/.style={-{Stealth[length=2.2mm,width=1.6mm]}, semithick, draw=gray!65!black},
    flow/.style={-{Stealth[length=2.2mm,width=1.6mm]}, semithick, draw=cyan!60!blue},
    blk/.style={draw=gray!45, line width=0.9pt},
    opbox/.style={
      fill=white,
      draw=cyan!55!blue,
      rounded corners=3pt,
      inner sep=3pt,
      font=\small
    },
    fnbox/.style={
      fill=orange!18,
      draw=orange!70!black,
      rounded corners=3pt,
      minimum width=1.05cm,
      minimum height=0.52cm,
      inner sep=2pt,
      font=\small
    },
    font=\small
  ]
\fill[cyan!12, rounded corners=10pt] (5.45,-0.45) rectangle (8.25,1.75);
\draw[cyan!45!blue, dashed, rounded corners=10pt] (5.45,-0.45) rectangle (8.25,1.75);
\node[anchor=south, font=\footnotesize, text=cyan!60!blue] at (6.85,1.77)
  {$f^{\mathrm{siamese}}_{\mathcal{S}_H}$};

\fill[orange!8, rounded corners=10pt] (3.85,-3.55) rectangle (11.95,-0.95);
\draw[orange!45, dashed, rounded corners=10pt] (3.85,-3.55) rectangle (11.95,-0.95);
\node[anchor=north, font=\footnotesize, text=orange!70!black] at (7.90,-3.57)
  {information-sharing path};

\coordinate (A) at (0,1.65);
\def\IW{2.8}
\def\IH{0.52}

\fill[blue!18]   (A)              rectangle ++(0.93,-\IH);
\fill[blue!36]   ($(A)+(0.93,0)$) rectangle ++(0.93,-\IH);
\fill[blue!54]   ($(A)+(1.86,0)$) rectangle ++(0.94,-\IH);
\fill[orange!20] ($(A)+(0,-0.66)$)              rectangle ++(0.93,-\IH);
\fill[orange!38] ($(A)+(0.93,-0.66)$) rectangle ++(0.93,-\IH);
\fill[orange!56] ($(A)+(1.86,-0.66)$) rectangle ++(0.94,-\IH);
\fill[teal!18]   ($(A)+(0,-1.32)$)              rectangle ++(0.93,-\IH);
\fill[teal!36]   ($(A)+(0.93,-1.32)$) rectangle ++(0.93,-\IH);
\fill[teal!54]   ($(A)+(1.86,-1.32)$) rectangle ++(0.94,-\IH);
\draw[blk] (A) rectangle ++(\IW,-\IH);
\draw[blk] ($(A)+(0,-0.66)$) rectangle ++(\IW,-\IH);
\draw[blk] ($(A)+(0,-1.32)$) rectangle ++(\IW,-\IH);
\draw[gray!35, thin] ($(A)+(0.93,0)$) -- ++(0,-1.84);
\draw[gray!35, thin] ($(A)+(1.86,0)$) -- ++(0,-1.84);

\coordinate (r1) at (2.80,1.39);
\coordinate (r2) at (2.80,0.73);
\coordinate (r3) at (2.80,0.07);

\draw[{Stealth[length=1.5mm]}-{Stealth[length=1.5mm]}, thin, gray!55]
  ($(A)+(0,0.30)$) -- node[above, font=\footnotesize] {$H$} ($(A)+(\IW,0.30)$);
\draw[{Stealth[length=1.5mm]}-{Stealth[length=1.5mm]}, thin, gray!55]
  ($(A)+(-0.28,0)$) -- node[left, font=\footnotesize] {$L$} ($(A)+(-0.28,-1.84)$);
\node[below=2pt, font=\footnotesize] at ($(A)+(\IW/2,-1.84)$) {$\mathbf{A_{n_1, n_2}}$};

\def\yone{1.39}
\def\ytwo{0.73}
\def\ythree{0.07}
\def\busX{3.30}

\draw[flow] (r1) -- (\busX,\yone);
\draw[flow] (r2) -- (\busX,\ytwo);
\draw[flow] (r3) -- (\busX,\ythree);
\draw[flow] (\busX,\yone) -- (\busX,\ythree);

\draw[flow] (\busX,\ythree) -- (\busX,-0.40);
\node[opbox] (sumH) at (\busX,-0.76) {$\mathrm{sum}_H$};
\draw[arr] (\busX,-0.40) -- (sumH.north);
\draw[arr] (sumH.south) -- (\busX,-1.20) -- (4.31,-1.20) -- (4.31,-1.45);

\node[fnbox] (s1) at (6.98,\yone) {$f_{\mathcal{S}_H}$};
\node[fnbox] (s2) at (6.98,\ytwo) {$f_{\mathcal{S}_H}$};
\node[fnbox] (s3) at (6.98,\ythree) {$f_{\mathcal{S}_H}$};

\draw[flow] (\busX,\yone)   -- (s1.west);
\draw[flow] (\busX,\ytwo)   -- (s2.west);
\draw[flow] (\busX,\ythree) -- (s3.west);
\draw[flow] (s1.east) -- (8.72,\yone);
\draw[flow] (s2.east) -- (8.72,\ytwo);
\draw[flow] (s3.east) -- (8.72,\ythree);

\coordinate (T) at (9.10,1.65);
\def\TW{2.55}
\def\TH{0.52}

\fill[pink!22]          (T)              rectangle ++(\TW,-\TH);
\draw[blk]              (T)              rectangle ++(\TW,-\TH);
\fill[yellow!28!orange] ($(T)+(0,-0.66)$) rectangle ++(\TW,-\TH);
\draw[blk]              ($(T)+(0,-0.66)$) rectangle ++(\TW,-\TH);
\fill[teal!28]          ($(T)+(0,-1.32)$) rectangle ++(\TW,-\TH);
\draw[blk]              ($(T)+(0,-1.32)$) rectangle ++(\TW,-\TH);

\node[circle, draw=cyan!55!blue, line width=0.9pt, fill=white,
      minimum size=8mm, inner sep=0pt, font=\normalsize] (plus) at (12.60,\ytwo) {$+$};
\draw[arr] ($(T)+(\TW,0)+(0,-0.66)+(0,-\TH/2)$) -- (plus.west);
\draw[arr] (plus.east) -- node[above, font=\footnotesize, xshift=10pt] {$f_\mathcal{W}(\mathbf{A}_{n_1, n_2})$} ++(1.15,0);

\coordinate (B1) at (4.00,-1.45);
\def\NW{0.62}
\def\NH{0.52}

\fill[blue!30]    (B1)              rectangle ++(\NW,-\NH);
\draw[blk]        (B1)              rectangle ++(\NW,-\NH);
\fill[orange!35]  ($(B1)+(0,-0.66)$) rectangle ++(\NW,-\NH);
\draw[blk]        ($(B1)+(0,-0.66)$) rectangle ++(\NW,-\NH);
\fill[teal!30]    ($(B1)+(0,-1.32)$) rectangle ++(\NW,-\NH);
\draw[blk]        ($(B1)+(0,-1.32)$) rectangle ++(\NW,-\NH);

\draw[arr] (4.62,-2.37) -- (4.88,-2.37);
\node[opbox, draw=magenta!55] (fcl) at (5.20,-2.37) {$f_{\mathcal{C}_L}$};
\draw[arr] (fcl.east) -- (5.72,-2.37);

\coordinate (B2) at (5.87,-1.45);
\fill[pink!18]        (B2)              rectangle ++(\NW,-\NH);
\draw[blk]            (B2)              rectangle ++(\NW,-\NH);
\fill[pink!30]        ($(B2)+(0,-0.66)$) rectangle ++(\NW,-\NH);
\draw[blk]            ($(B2)+(0,-0.66)$) rectangle ++(\NW,-\NH);
\fill[green!22]       ($(B2)+(0,-1.32)$) rectangle ++(\NW,-\NH);
\draw[blk]            ($(B2)+(0,-1.32)$) rectangle ++(\NW,-\NH);

\node[opbox] (brod) at (7.80,-2.37) {$\mathrm{brod}_H$};
\draw[arr] (6.49,-2.37) -- (brod.west);
\draw[arr] (brod.east) -- (9.10,-2.37);

\coordinate (B3) at (9.10,-1.45);
\def\BW{2.55}
\fill[pink!14]        (B3)              rectangle ++(\BW,-\NH);
\draw[blk]            (B3)              rectangle ++(\BW,-\NH);
\fill[pink!24]        ($(B3)+(0,-0.66)$) rectangle ++(\BW,-\NH);
\draw[blk]            ($(B3)+(0,-0.66)$) rectangle ++(\BW,-\NH);
\fill[green!18]       ($(B3)+(0,-1.32)$) rectangle ++(\BW,-\NH);
\draw[blk]            ($(B3)+(0,-1.32)$) rectangle ++(\BW,-\NH);

\draw[arr] (11.65,-2.37) -- (12.60,-2.37) -- (plus.south);
  \end{tikzpicture}}
  \caption{The $\mathcal{W}$-equivariant layer $f_\mathcal{W}(\mathbf{A}_{n_1, n_2})$. The Siamese path (dashed, top) applies $f_{\mathcal{S}_H}$ to each layer slice independently; the information-sharing path (bottom) sums over heads, applies $f_{\mathcal{C}_L}$ across layers, and broadcasts the result back over heads. The two paths are summed.}
  \label{fig:encoder}
\end{figure}

In \Cref{fig:encoder} we provide a visual depiction of the $\mathcal{W}$-equivariant layer $f_\mathcal{W}$ described in \Cref{subsec:attention-symmetries-and-equivariant-layers}. Each layer slice $\mathbf{A}^{l}_{n_1,n_2}$ is processed independently by a shared (Siamese) $f_{\mathcal{S}_H}$, while the information-sharing path sums the input over heads, applies $f_{\mathcal{C}_L}$ across layers, and broadcasts the result back over heads; the outputs of the two paths are summed to produce $f_\mathcal{W}(\mathbf{A}_{n_1,n_2})$.

\subsection{\lemon architecture}
\label{app:EAGM-vis}
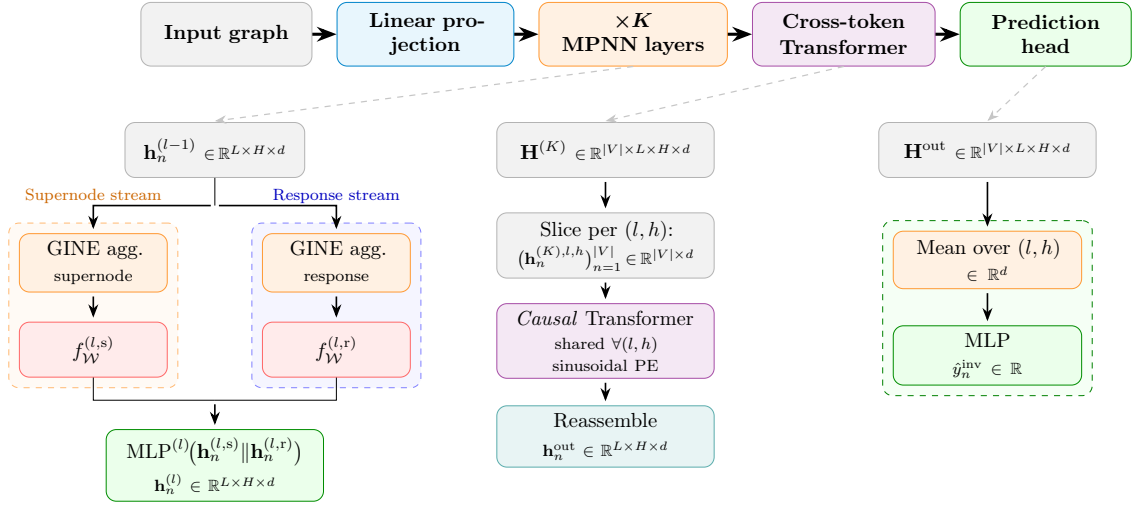
\begin{figure}[htb]
  \centering
  \resizebox{0.9\textwidth}{!}{%
  \begin{tikzpicture}[
      >=Stealth,
      every node/.style={font=\small},
      BI/.style={draw=gray!60,        fill=gray!10,   rounded corners=5pt,
                 align=center, minimum height=0.95cm, text width=#1},  BI/.default=2.8cm,
      BP/.style={draw=cyan!65!blue,   fill=cyan!8,    rounded corners=5pt,
                 align=center, minimum height=0.95cm, text width=#1},  BP/.default=2.8cm,
      BM/.style={draw=orange!80,      fill=orange!11, rounded corners=5pt,
                 align=center, minimum height=0.95cm, text width=#1},  BM/.default=2.3cm,
      BF/.style={draw=red!60,         fill=red!8,     rounded corners=5pt,
                 align=center, minimum height=0.95cm, text width=#1},  BF/.default=2.5cm,
      BG/.style={draw=green!55!black, fill=green!8,   rounded corners=5pt,
                 align=center, minimum height=0.95cm, text width=#1},  BG/.default=4.0cm,
      BT/.style={draw=violet!70,      fill=violet!9,  rounded corners=5pt,
                 align=center, minimum height=0.95cm, text width=#1},  BT/.default=2.7cm,
      BO/.style={draw=teal!65,        fill=teal!9,    rounded corners=5pt,
                 align=center, minimum height=0.95cm, text width=#1},  BO/.default=2.6cm,
      arr/.style ={->, thick,          shorten >=2pt, shorten <=2pt},
      arrh/.style={->, line width=1.5pt, shorten >=2pt, shorten <=2pt},
      pipearr/.style={->, line width=1.5pt},
      zoom/.style={->, dashed, gray!45, semithick, rounded corners=4pt},
      badge/.style={circle, draw=gray!55, fill=white,
                    font=\small\bfseries, inner sep=1pt, minimum size=16pt},
  ]

  \node[BI=2.7cm, minimum height=1.1cm] (A1) at (2.65, 0)
      {\textbf{Input graph}};
  \node[BP=2.8cm, minimum height=1.1cm, right=0.42cm of A1] (A2)
      {\textbf{Linear projection}};
  \node[BM=3.0cm, minimum height=1.1cm, right=0.42cm of A2] (A3)
      {$\boldsymbol{\times K}$\\\textbf{MPNN layers}};
  \node[BT=2.9cm, minimum height=1.1cm, right=0.42cm of A3] (A4)
      {\textbf{Cross-token}\\[1pt]\textbf{Transformer}};
  \node[BG=2.7cm, minimum height=1.1cm, right=0.42cm of A4] (A6)
      {\textbf{Prediction head}};

  \coordinate (Apipe) at (0,0);
  \draw[pipearr] (A1.east |- Apipe) -- (A2.west |- Apipe);
  \draw[pipearr] (A2.east |- Apipe) -- (A3.west |- Apipe);
  \draw[pipearr] (A3.east |- Apipe) -- (A4.west |- Apipe);
  \draw[pipearr] (A4.east |- Apipe) -- (A6.west |- Apipe);

  \begin{scope}[yshift=-2.00cm, xshift=2.45cm]

  \coordinate (ZoomBtop) at (0, 0.50);

  \node[BI=2.85cm] (Bin) at (0, 0)
      {$\mathbf{h}_n^{(l-1)}\;${\scriptsize$\in\!\mathbb{R}^{L\times H\times d}$}};

  \node[BM=2.30cm] (Bsg) at (-2.10,-1.95)
      {GINE agg.\\[1pt]{\scriptsize supernode}};
  \node[BF=2.30cm] (Bsf) at (-2.10,-3.45)
      {$f_{\mathcal{W}}^{(l,\mathrm{s})}$};

  \node[BM=2.30cm] (Brg) at ( 2.10,-1.95)
      {GINE agg.\\[1pt]{\scriptsize response}};
  \node[BF=2.30cm] (Brf) at ( 2.10,-3.45)
      {$f_{\mathcal{W}}^{(l,\mathrm{r})}$};

  \node[BG=3.50cm] (Bfu) at (0,-5.45)
      {$\mathrm{MLP}^{(l)}\!\bigl(\mathbf{h}_n^{(l,\mathrm{s})}\|\mathbf{h}_n^{(l,\mathrm{r})}\bigr)$\\[3pt]
       {\scriptsize$\mathbf{h}_n^{(l)}\!\in\!\mathbb{R}^{L\times H\times d}$}};

  \coordinate (Bsplit) at ($(Bin.south)!0.5!(Bin.south |- Bsg.north)$);
  \draw (Bin.south) -- (Bsplit);
  \draw[arr] (Bsplit) -| (Bsg.north);
  \draw[arr] (Bsplit) -| (Brg.north);
  \draw[arr] (Bsg.south) -- (Bsf.north);
  \draw[arr] (Brg.south) -- (Brf.north);
  \coordinate (Bmerge) at ($(Bfu.north)+(0,0.50)$);
  \draw (Bsf.south) |- (Bmerge);
  \draw (Brf.south) |- (Bmerge);
  \draw[arr] (Bmerge) -- (Bfu.north);

  \begin{pgfonlayer}{background}
    \node[draw=orange!55, dashed, rounded corners=5pt, fill=orange!4, inner sep=5pt,
          fit=(Bsg)(Bsf),
          label={[font=\scriptsize,orange!80!black,yshift=6pt]above:Supernode stream}] {};
    \node[draw=blue!50,   dashed, rounded corners=5pt, fill=blue!4,   inner sep=5pt,
          fit=(Brg)(Brf),
          label={[font=\scriptsize,blue!75!black,yshift=6pt]above:Response stream}] {};
  \end{pgfonlayer}

  \end{scope}

  \begin{scope}[yshift=-2.00cm, xshift=9.20cm]

  \coordinate (ZoomCtop) at (0, 0.50);

  \node[BI=3.50cm] (Ci) at (0, 0)
      {$\mathbf{H}^{(K)}\;${\scriptsize$\in\!\mathbb{R}^{|V|\times L\times H\times d}$}};

  \node[BI=3.50cm] (Csl) at (0,-1.65)
      {Slice per $(l,h)$:\\[2pt]
       {\scriptsize$\bigl(\mathbf{h}_n^{(K),l,h}\bigr)_{n=1}^{|V|}\!\in\!\mathbb{R}^{|V|\times d}$}};

  \node[BT=3.50cm] (Ctr) at (0,-3.30)
      {\emph{Causal} $\mathrm{Transformer}$\\[2pt]
       {\scriptsize shared $\forall(l,h)$\\[-1pt] sinusoidal PE}};

  \node[BO=3.50cm] (Cor) at (0,-4.95)
      {Reassemble\\[2pt]
       {\scriptsize$\mathbf{h}_n^{\mathrm{out}}\!\in\!\mathbb{R}^{L\times H\times d}$}};

  \draw[arr] (Ci.south)  -- (Csl.north);
  \draw[arr] (Csl.south) -- (Ctr.north);
  \draw[arr] (Ctr.south) -- (Cor.north);

  \end{scope}

  \begin{scope}[yshift=-2.00cm, xshift=15.80cm]

  \coordinate (ZoomDtop) at (0, 0.50);

  \node[BI=3.50cm] (Din) at (0, 0)
      {$\mathbf{H}^{\mathrm{out}}\;${\scriptsize$\in\!\mathbb{R}^{|V|\times L\times H\times d}$}};

  \node[BM=3.00cm] (Dm1) at (0,-1.90)
      {Mean over $(l,h)$\\[1pt]{\scriptsize$\in\!\mathbb{R}^{d}$}};
  \node[BG=3.00cm] (Dm2) at (0,-3.55)
      {MLP\\[1pt]{\scriptsize$\hat{y}_n^{\mathrm{inv}}\!\in\!\mathbb{R}$}};

  \draw[arr] (Din.south) -- (Dm1.north);
  \draw[arr] (Dm1.south) -- (Dm2.north);

  \begin{pgfonlayer}{background}
    \node[draw=green!50!black, dashed, rounded corners=5pt, fill=green!4, inner sep=5pt,
          fit=(Dm1)(Dm2),
          label={[font=\scriptsize,green!55!black,yshift=6pt]}] {};
  \end{pgfonlayer}

  \end{scope}

  \draw[zoom] (A3.south) -- (ZoomBtop);
  \draw[zoom] (A4.south) -- (ZoomCtop);
  \draw[zoom] (A6.south) -- (ZoomDtop);

  \end{tikzpicture}%
  }
  \caption{%
  The \lemon architecture. \emph{Top}: the end-to-end pipeline. \emph{Bottom, left to right}: one equivariant MPNN layer---parallel supernode/response streams with $\mathcal{W}$-equivariant updates (\Cref{eq:eagn-gine,eq:eagn-fuse}); the shared \emph{causal} cross-token Transformer, applied independently at each layer--head cell $(l,h)$; and the invariant prediction head for the per-token case (\Cref{eq:eagn-main-head}).
  }
  \label{fig:eagn-arch}
  \end{figure}

In \Cref{fig:eagn-arch} we provide a visual depiction of the full \lemon architecture.

\clearpage
\section{From attention maps to Graphs with Structured Features - Extended Section}\label{app:From attention maps to Graphs with Structured Features - Extended Section}

\begin{wrapfigure}{r}{0.55\textwidth}
  \vspace{-24pt}
  \centering
  \resizebox{\linewidth}{!}{%
  \begin{tikzpicture}[
      >=Stealth,
      pnode/.style={circle, fill=blue!15, draw=blue!50,
                    minimum size=11pt, inner sep=0pt, font=\tiny},
      rnode/.style={circle, fill=orange!22, draw=orange!60,
                    minimum size=11pt, inner sep=0pt, font=\tiny},
      wnode/.style={circle, fill=orange!22, draw=orange!60, double,
                    minimum size=13pt, inner sep=0pt, font=\tiny},
      snode/.style={circle, fill=teal!22, draw=teal!60,
                    minimum size=14pt, inner sep=0pt, font=\scriptsize},
      supedge/.style={->, line width=0.45pt, draw=orange!80!black},
      respedge/.style={->, line width=0.45pt, draw=blue!70!black},
      wedge/.style={->, line width=0.3pt,  draw=gray!50, dashed},
      pedge/.style={->, line width=0.45pt, draw=red!60!black},
      arr/.style={->, draw=black!55, line width=0.6pt}
  ]
  \node[pnode] (p1)  at (0.00, 1.25) {$p_1$};
  \node[pnode] (p2)  at (0.00, 0.35) {$p_2$};
  \node[rnode] (r1)  at (1.15, 1.60) {$r_1$};
  \node[rnode] (r2)  at (1.15, 0.70) {$r_2$};
  \node[rnode] (r3)  at (1.15, 0.00) {$r_3$};
  \draw[pedge] (p1) to[bend left=40] (p2);
  \draw[supedge] (p1) -- (r1);
  \draw[supedge] (p2) -- (r2);
  \draw[wedge] (p1) to[bend left=10]  (r2);
  \draw[wedge] (p2) to[bend right=12] (r3);
  \draw[wedge] (r1) -- (r2);
  \draw[respedge] (r2) -- (r3);
  \node[font=\tiny, text=gray!70] at (0.575,-0.45) {original};

  \draw[arr] (1.40,0.70)--(2.07,0.70)
    node[midway, above, font=\tiny, yshift=1pt]{$\delta$};

  \node[pnode] (p1b) at (2.32, 1.25) {$p_1$};
  \node[pnode] (p2b) at (2.32, 0.35) {$p_2$};
  \node[rnode] (r1b) at (3.47, 1.60) {$r_1$};
  \node[rnode] (r2b) at (3.47, 0.70) {$r_2$};
  \node[rnode] (r3b) at (3.47, 0.00) {$r_3$};
  \draw[supedge] (p1b) -- (r1b);
  \draw[supedge] (p2b) -- (r2b);
  \draw[respedge] (r2b) -- (r3b);
  \draw[draw=gray!55, dashed, rounded corners=3pt, line width=0.4pt]
    ($(p1b.north west)+(-0.09,0.08)$) rectangle ($(p2b.south east)+(0.09,-0.08)$);
  \node[font=\tiny, text=gray!70] at (2.895,-0.45) {sparse};

  \draw[arr] (3.72,0.70)--(4.19,0.70)
    node[midway, above, font=\tiny, yshift=0.5pt]{pool};

  \node[snode] (s)   at (4.44, 0.70) {$s$};
  \node[rnode] (r1c) at (5.59, 1.60) {$r_1$};
  \node[rnode] (r2c) at (5.59, 0.70) {$r_2$};
  \node[rnode] (r3c) at (5.59, 0.00) {$r_3$};
  \draw[supedge] (s) -- (r1c);
  \draw[supedge] (s) -- (r2c);
  \draw[respedge] (r2c) -- (r3c);
  \draw[draw=gray!55, dashed, rounded corners=3pt, line width=0.4pt]
    ($(r1c.north west)+(-0.09,0.08)$) rectangle ($(r2c.south east)+(0.09,-0.08)$);
  \draw[draw=gray!55, dashed, rounded corners=3pt, line width=0.4pt]
    ($(r3c.north west)+(-0.09,0.08)$) rectangle ($(r3c.south east)+(0.09,-0.08)$);
  \node[font=\tiny, text=gray!70] at (5.015,-0.45) {supernode};

  \draw[arr] (5.98,0.70)--(6.72,0.70)
    node[midway, above, font=\tiny, yshift=1pt]{window};

  \node[snode] (s2)  at (7.00, 0.70) {$s$};
  \node[wnode] (w1)  at (7.95, 1.15) {$w_1$};
  \node[wnode] (w2)  at (7.95, 0.25) {$w_2$};
  \draw[supedge] (s2) -- (w1);
  \draw[respedge] (w1) -- (w2);
  \node[font=\tiny, text=gray!70] at (7.475,-0.45) {windowed};

  \begin{scope}[shift={(8.80,1.30)}, scale=0.40]
    \fill[orange!25] (0,-1) rectangle (1,-2);
    \fill[blue!20]   (1,-2) rectangle (2,-3);
    \fill[gray!15] (0, 0) rectangle (1,-1);
    \fill[gray!15] (1,-1) rectangle (2,-2);
    \fill[gray!15] (2,-2) rectangle (3,-3);
    \draw[gray!45, very thin] (0,0) grid (3,-3);
    \draw[black!70, thin] (0,0) rectangle (3,-3);
    \draw[black!55, line width=0.7pt] (1, 0.05)--(1,-3.05);
    \draw[black!55, line width=0.7pt] (-0.05,-1)--(3.05,-1);
    \node[font=\tiny, text=gray!60] at (0.5,-0.5) {0};
    \node[font=\tiny, text=gray!60] at (1.5,-0.5) {0};
    \node[font=\tiny, text=gray!60] at (2.5,-0.5) {0};
    \node[font={\tiny\bfseries}] at (0.5,-1.5) {1};
    \node[font=\tiny, text=gray!60] at (1.5,-1.5) {0};
    \node[font=\tiny, text=gray!60] at (2.5,-1.5) {0};
    \node[font=\tiny, text=gray!60] at (0.5,-2.5) {0};
    \node[font={\tiny\bfseries}] at (1.5,-2.5) {1};
    \node[font=\tiny, text=gray!60] at (2.5,-2.5) {0};
    \node[font={\tiny\bfseries}] at (0.5, 0.40) {$s$};
    \node[font=\tiny] at (1.5, 0.40) {$w_1$};
    \node[font=\tiny] at (2.5, 0.40) {$w_2$};
    \node[font={\tiny\bfseries}] at (-0.5,-0.5) {$s$};
    \node[font=\tiny] at (-0.5,-1.5) {$w_1$};
    \node[font=\tiny] at (-0.5,-2.5) {$w_2$};
    \node[font=\tiny, gray!50!black] at (1.5,-3.65)
      {\textcolor{orange!80!black}{\rule{1ex}{1ex}}~super\enspace
       \textcolor{blue!70!black}{\rule{1ex}{1ex}}~resp.};
  \end{scope}
  \node[font=\tiny, text=gray!70] at (9.40,-0.45) {adjacency};

  \useasboundingbox (-0.42,-0.63) rectangle (10.05,1.90);
  \end{tikzpicture}}
  \vspace{-25pt}
  \caption{Preprocessing: p${\to}$p edges (red) discarded; weak edges (gray, dashed) thresholded away; prompt nodes (boxed) pooled into supernode~$s$; response tokens (boxed) pooled into windows ($W{=}2$ shown). \emph{Right}: adjacency of the final graph with \textcolor{orange!80!black}{super} and \textcolor{blue!70!black}{response} stream blocks (edge colors match).}
  \label{fig:preprocess}
  \vspace{-4pt}
\end{wrapfigure}
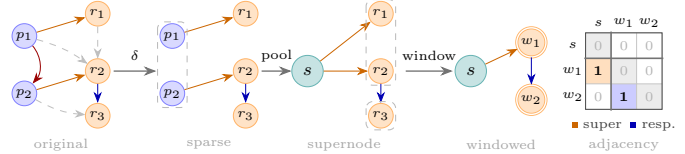
\paragraph{Preprocessing the attention graph.} The full preprocessing pipeline is illustrated in \Cref{fig:preprocess}. Following \citep{frasca2026neural}, we discard both \emph{prompt-to-prompt} (p${\to}$p) edges, and sparsify the remaining edges by thresholding low attention values\footnote{We set the threshold to $\delta = 0.1$ throughout the paper. \citep{frasca2026neural} has shown that this threshold substantially reduces the graph size, while maintaining the relevant information for hallucination detection.}. We also \emph{pool} all prompt nodes into a single \emph{supernode}~$s$ and aggregate their outgoing edges to the response nodes, leaving a graph over response tokens only (plus $s$).\footnote{The edges from the prompt tokens to a given response token are merged into a single edge from $s$, whose $L \times H$ attribute is the entry-wise mean over the merged (post-threshold) edges.}
Finally, we pool consecutive response tokens into non-overlapping \emph{windows} ($W$), so the nodes of the final graph correspond to response windows rather than individual tokens.\footnote{We set the window size to $W=8$ throughout the paper. Each window becomes a single node whose $L \times H$ attribute is the entry-wise mean over its tokens' attributes. Edge endpoints are remapped to the window nodes, and edges that consequently connect the same pair of nodes are merged into a single edge whose $L \times H$ attribute is the entry-wise mean over the merged edges. Per-token labels pool conservatively: a window is labeled hallucinated iff \emph{any} of its tokens is.}
Together, these steps substantially reduce graph size while preserving both the structure and the information relevant to hallucination detection.
The same pipeline is used for the jailbreak experiments, with two changes that follow from the monitor there observing only the \emph{prompt}: the nodes are the prompt's own tokens, so it is the chat template's \emph{system} span that is pooled into the supernode~$s$,\footnote{The span is a genuine $24$-token system message only for the two Qwen models; for the other four it is a $2$--$4$ token role marker, so pooling it is close to a no-op there.} and no window pooling is applied ($W{=}1$), since these prompts are short to begin with.
We next introduce the \emph{LANGUAGE-MODEL EQUIVARIANT MONITORING NETWORK } (\lemon), our architecture for processing these structured attributed graphs.

\subsection{\textbf{\underline{L}}anguage-model \textbf{\underline{E}}quivariant \textbf{\underline{MO}}nitoring \textbf{\underline{N}}etwork (\lemon) -- A Detailed Description}\label{app:eagn-detailed}

\lemon follows the Message Passing Neural Network (MPNN) framework~\citep{gilmer2017neural}, with a \emph{key extension:} every nonlinear map on node and edge tensors is a $\mathcal{W}$-equivariant layer.

\paragraph{Overview.} We denote by $|V|$ the number of nodes in the preprocessed graph. \lemon first projects node and edge features into hidden dimension $d$, yielding $\mathbf{h}^{(0)} \in \mathbb{R}^{|V| \times L \times H \times d}$ and $\mathbf{e} \in \mathbb{R}^{|V| \times |V| \times L \times H \times d}$; see details in \Cref{subsec:attention-graphs-structured}. We recall that the node features combine the self-attention scores with \emph{token probabilities}: each response token is assigned its probability $p_n$ and each prompt token the value $0$, and this scalar is broadcast across all layers and heads, concatenated with the self-attention score, and mapped to $\mathbb{R}^d$ by a shared linear layer.\footnote{Under the window pooling above, each response-window node carries the mean probability of the tokens it pools, while the supernode~$s$ carries the probability of the first response token. We ablate this signal in \Cref{subsec:ablation-studies} (the \emph{no prob} variant).} It then applies (i) $K$ MPNN layers in parallel to the super-node and response-node streams, using GINE-style aggregation \citep{hu2019strategies,xu2018powerful} with a $\mathcal{W}$-equivariant update function. (ii) A shared \emph{causal} Transformer encoder is then applied independently to each layer-head cell $(\ell,h)$ to enable global token interaction while preserving sequence order. Finally, (iii) an invariant prediction head operates on the resulting node features. We next formalize the (i) equivariant MPNN layers, (ii) cross-token mixing, and (iii) prediction head. In what follows we describe each component in detail, and we defer the full architecture visualization to \Cref{fig:eagn-arch}.

\paragraph{(i) \emph{Equivariant MPNN layers}.}
Let $t \in \{\mathrm{s}, \mathrm{r}\}$ index the \emph{supernode} and \emph{response} streams, respectively.
At layer~$l$, we follow a GINE-style aggregation rule~\citep{hu2019strategies,xu2018powerful} to update the node features, and specifically, we use $\operatorname{up}_{\mathcal{W}}^{(l)} : \mathbb{R}^{L \times H \times d} \to \mathbb{R}^{L \times H \times d}$ as the update function, which is $\mathcal{W}$-equivariant:
\begin{align}
    \mathbf{h}_n^{(l,t)}
    &=
    \operatorname{up}_{\mathcal{W}}^{(l)}\!\left(
        (1+\epsilon^{(l,t)})\,\mathbf{h}_n^{(l-1)}
        +
        \sum_{n' \in \mathcal{N}^{t}(n)}
        \bigl(\mathbf{h}_{n'}^{(l-1)} + \mathbf{e}_{n',n}\bigr)
    \right),
    \qquad
    t \in \{\mathrm{s}, \mathrm{r}\},
    \label{eq:eagn-gine} \\
    \mathbf{h}_n^{(l)}
    &=
    \texttt{MLP}^{(l)}\!\bigl(\mathbf{h}_n^{(l,\mathrm{s})} \,\|\, \mathbf{h}_n^{(l,\mathrm{r})}\bigr)
    \in \mathbb{R}^{L \times H \times d},
    \label{eq:eagn-fuse}
\end{align}
where $\mathcal{N}^{t}(n)$ is the set of \emph{in-neighbours} of $n$ in stream~$t$, $\epsilon^{(l,t)}$ is a learnable scalar.
The two stream outputs are then fused by an MLP.
We stack $K$ such layers, so the final node features are $\mathbf{h}_n^{(K)} \in \mathbb{R}^{L \times H \times d}$.

We consider two instantiations of $\operatorname{up}_{\mathcal{W}}^{(l)}$ from \Cref{eq:w-equiv}, both equivariant to $\mathcal{W}$: (a) \lemonW, a variant that mixes information across layers and heads, instantiating $f_{\mathcal{C}L}$ and $f_{\mathcal{S}_H}$ as Transformers; and (b) \lemonWpc, a variant that applies an MLP to each node feature independently.

\paragraph{\emph{(ii) Cross-token transformer}.}
So far, information has propagated only along graph edges.
To exchange information \emph{across} the full token sequence, we apply a shared \emph{causal} Transformer encoder\footnote{With a fixed (non-learnable) sinusoidal absolute positional encoding, applied only in this cross-token encoder.} independently at each layer--head cell $(\ell,h) \in [L] \times [H]$: the $(\ell,h)$ slices $[\mathbf{h}_1^{(K),\ell,h}, \ldots, \mathbf{h}_{|V|}^{(K)\ell,h}] \in \mathbb{R}^{|V| \times d}$, ordered by token position, form a length-$|V|$ sequence that the encoder maps to the output node features $\mathbf{h}_n^{\mathrm{out},\ell,h}$. Information is thus exchanged across tokens, but \emph{not} across layers or heads.

\paragraph{(iii) Prediction head.}
Given the backbone output $\mathbf{H}^{\mathrm{out}} \coloneqq \bigl[\mathbf{h}_n^{\mathrm{out},\ell,h}\bigr]_{(n,\ell,h) \in V \times [L] \times [H]} \in \mathbb{R}^{|V| \times L \times H \times d}$, we obtain the invariant hallucination score $\hat{y}_n^{\mathrm{inv}}$ as follows:
\begin{equation}
    \hat{y}_n^{\mathrm{inv}}
    =
    \texttt{MLP}\!\Biggl(
        \frac{1}{|L||H|}
        \sum_{\ell \in [L]}
        \sum_{h \in [H]}
        \mathbf{h}_n^{\mathrm{out},\ell,h}
    \Biggr)
    \in \mathbb{R}.
    \label{eq:eagn-main-head}
\end{equation}
For graph-level tasks, we read out a single score from the last token $|V|$, which under causality has seen the entire sequence:
\begin{equation}
    \hat{y}_G^{\mathrm{inv}}
    =
    \texttt{MLP}\!\Biggl(
        \frac{1}{|L||H|}
        \sum_{\ell \in [L]}
        \sum_{h \in [H]}
        \mathbf{h}_{|V|}^{\mathrm{out},\ell,h}
    \Biggr)
    \in \mathbb{R}.
    \label{eq:eagn-graph-head}
\end{equation}
Since this readout selects along the token axis only, $\mathcal{W}$-invariance is preserved.

\clearpage
\section{Alignment Methods}
\label{app:alignment-methods}

This appendix gives the full formulation of the alignment-based approach we use in our experiments. Rather than building cross-model head-permutation symmetry into an equivariant backbone, this approach enforces it in feature space: each LLM's layer--head attention features are aligned to a shared canonical indexing, after which any off-the-shelf classifier can be trained on the aligned representations.

\paragraph{Setup.} To formulate cross-LLM detection via alignment, we decompose it into three sequential maps.
\begin{itemize}
    \item A \emph{feature extractor} $f_{\mathrm{extract}}: \mathbb{R}^{N \times N \times L \times H} \rightarrow \mathbb{R}^{N \times L \times H}$ collapses raw attentions (i.e., $\mathbf{A} \in \mathbb{R}^{N \times N \times L \times H}$) to per-token layer--head scalars (LookbackLens~\citep{chuang2024lookback} is one instantiation; see \Cref{app:lookbacklens} for details).
    \item An \emph{alignment map} $f_{\mathrm{align}}: \mathbb{R}^{N \times L \times H} \rightarrow \mathbb{R}^{N \times L \times H}$ then reindexes the layer--head axes into a canonical ordering.
    \item Finally, a \emph{classifier} $f_{\mathrm{cls}}: \mathbb{R}^{N \times L \times H} \rightarrow \mathbb{R}^N$ produces one hallucination score per token.
\end{itemize}

The end-to-end predictor is
\begin{equation}
    f(\mathbf{A}) = f_{\mathrm{cls}}\!\bigl(f_{\mathrm{align}}(f_{\mathrm{extract}}(\mathbf{A}))\bigr),
\end{equation}
where $f \coloneqq f_{\mathrm{cls}} \circ f_{\mathrm{align}} \circ f_{\mathrm{extract}}$.

\begin{wrapfigure}{R}{0.34\textwidth}
  \vspace{-8pt}
  \centering
\begin{tikzpicture}[
  hd/.style={draw=gray!55, rounded corners=1.5pt,
             minimum width=0.56cm, minimum height=0.48cm, inner sep=0pt},
  lb/.style={font=\tiny, color=gray!65},
  ar/.style={-Stealth, semithick, shorten >=2pt, shorten <=2pt}
]
\colorlet{cA}{red!50}
\colorlet{cB}{blue!50}
\colorlet{cC}{green!55!black}
\pgfmathsetmacro{\dxb}{0.68}
\node[hd, fill=cB]                  (s1) at ({0*\dxb}, 1.5) {};
\node[hd, fill=cA]                  (s2) at ({1*\dxb}, 1.5) {};
\node[hd, fill=gray!18, draw=gray!30](s3) at ({2*\dxb}, 1.5) {};
\node[hd, fill=cC]                  (s4) at ({3*\dxb}, 1.5) {};
\node[font=\tiny, red!55!gray] at ({2*\dxb}, 1.5) {$\times$};
\node[font=\tiny, gray!75, left=3pt of s1, align=right] {src\\layer $\ell$};
\node[hd, fill=cA] (t1) at ({0*\dxb}, 0) {};
\node[hd, fill=cC] (t2) at ({1*\dxb}, 0) {};
\node[hd, fill=cB] (t3) at ({2*\dxb}, 0) {};
\node[font=\tiny, gray!75, left=3pt of t1, align=right] {tgt\\$\phi(\ell)$};
\foreach \n in {1,...,4} { \node[lb] at ({(\n-1)*\dxb},  1.90) {\tiny$h{=}\n$}; }
\foreach \n in {1,...,3} { \node[lb] at ({(\n-1)*\dxb}, -0.38) {\tiny$h'{=}\n$};  }
\draw[ar, cA!80!black               ] (t1) to (s2);
\draw[ar, cC!80!black, bend left=15 ] (t2) to (s4);
\draw[ar, cB!80!black, bend left=10 ] (t3) to (s1);
\node[font=\scriptsize, gray!60, align=center] at ({3.45*\dxb}, 0.60)
  {$H^{\min}{=}3$\\\tiny matched};
\node[font=\tiny, gray!60, align=center] at ({1.5*\dxb}, 2.28)
  {$\phi(\ell)=\lfloor \ell\cdot L^t/L^s\rfloor$};
\end{tikzpicture}
  \caption{Unequal $H^s, L^s \!\neq\! H^t, L^t$. Layer map $\phi$ pairs source and target layers; only $H^{\min}\!=\!\min(H^s,H^t)$ heads are matched.}
  \label{fig:head-align-unequal}
\end{wrapfigure}
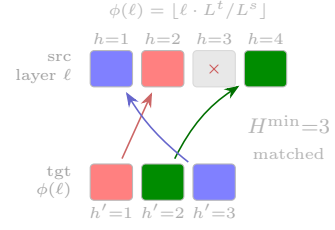

\begin{algorithm}[t]
  \caption{\textsc{HeadAlign}($\mathbf{X}$, $\mathbf{y}$)}
  \label{alg:head-align}
  \begin{algorithmic}[1]
    \Require $\mathbf{X} \in \mathbb{R}^{N_s \times L^s \times H^s}$ (source training),\quad $\mathbf{y} \in \mathbb{R}^{M \times L^t \times H^t}$ (test sample)
    \Ensure  $\tilde{\mathbf{y}} \in \mathbb{R}^{M \times L^s \times H^s}$ (aligned test features)
    \State $\bar{X}^{\ell,h} \gets \frac{1}{N_s}\sum_{q} X_{q}^{\ell,h}$ \quad for all $\ell \in [L^s],\; h \in [H^s]$ \Comment{precomputed offline}
    \State $\bar{y}^{\ell,h} \gets \frac{1}{M}\sum_{q} y_{q}^{\ell,h}$ \quad for all $\ell \in [L^t],\; h \in [H^t]$ \Comment{computed at inference time}
    \For{$\ell = 1, \ldots, L^s$}
      \State $\phi(\ell) \gets \lfloor \ell \cdot L^t / L^s \rfloor$ \Comment{proportional layer correspondence}
      \State $\mathbf{C}^{\ell}_{h,h'} \gets \bigl|\bar{X}^{\ell,h} - \bar{y}^{\phi(\ell),h'}\bigr|$ \quad $\mathbf{C}^{\ell} \in \mathbb{R}^{H^s \times H^t}$
      \State $\{(h_{i,\ell}, h'_{i,\ell})\}_{i=1}^{H^{\min}} \gets \textsc{Hungarian}(\mathbf{C}^{\ell})$ \Comment{$H^{\min}=\min(H^s,H^t)$ matched pairs}
    \EndFor
    \State $\tilde{\mathbf{y}} \gets \mathbf{0} \in \mathbb{R}^{M \times L^s \times H^s}$
    \For{each layer $\ell$ and matched pair $(h_{i,\ell},\, h'_{i,\ell})$}
      \State $\tilde{y}_{q,\,\ell,\,h_{i,\ell}} \gets y_{q,\,\phi(\ell),\,h'_{i,\ell}}$ \quad for all $q$ \Comment{unmatched source positions stay zero}
    \EndFor
    \State \Return $\tilde{\mathbf{y}}$
  \end{algorithmic}
  \end{algorithm}

\paragraph{Aligning two LLMs' attention features.}
Let $\mathcal{L}^s$ and $\mathcal{L}^t$ be a source and target LLM with $H^s$ heads and $L^s$ layers, and $H^t$ heads and $L^t$ layers, respectively (we assume the head count is fixed across layers within each model, as in standard transformers).
Denote by $\mathbf{X} \in \mathbb{R}^{N_s \times L^s \times H^s}$ the source training features, where $N_s$ is the total number of tokens pooled over all labelled source training samples, and by $\mathbf{y} \in \mathbb{R}^{M \times L^t \times H^t}$ the features of a single unlabelled test sample with $M$ tokens; each entry $X_q^{\ell,h},\, y_q^{\ell,h} \in \mathbb{R}$ is the scalar feature at token $q$, layer $\ell$, and head $h$.
Because head order within each layer is an arbitrary permutation, applying a classifier trained on $\mathbf{X}$ to raw $\mathbf{y}$ would fail under zero-shot transfer (as we also demonstrate in \Cref{sec:experiments}); the goal is therefore to reindex $\mathbf{y}$ into the source layer--head coordinate system, producing $\tilde{\mathbf{y}} \in \mathbb{R}^{M \times L^s \times H^s}$.

We reindex $\mathbf{y}$ by matching each target layer and head to a source layer and head in two steps, as follows.

\textit{Layer correspondence.}
Source layer $\ell \in [L^s]$ is matched to target layer via proportional depth scaling \citep{wolfram2025layerssimilardepthsgenerate,fu2026cachetocache} (fixed, not optimized) as follows:
\begin{equation}
  \phi(\ell) = \left\lfloor \ell \cdot \frac{L^t}{L^s} \right\rfloor \in [L^t].
\end{equation}

\textit{Head alignment.}
For each source layer $\ell$, we define mean features as follows:
\begin{align*}
  \bar{X}^{\ell,h} = \frac{1}{N_s} \sum_{q=1}^{N_s} X_q^{\ell,h}, \qquad
  \bar{y}^{\ell,h} = \frac{1}{M} \sum_{q=1}^{M} y_q^{\ell,h},
\end{align*}
where $\bar{\mathbf{X}} \in \mathbb{R}^{L^s \times H^s}$ and $\bar{\mathbf{y}} \in \mathbb{R}^{L^t \times H^t}$ collect all layer--head means.
$\bar{\mathbf{X}}$ is precomputed once from labelled source training data; $\bar{\mathbf{y}}$ is computed at inference from the test sample alone---no target labels are required.
We build the per-layer rectangular cost matrix $\mathbf{C}^{\ell} \in \mathbb{R}^{H^s \times H^t}$ with entries $\mathbf{C}^{\ell}_{h,h'} = \bigl|\bar{X}^{\ell,h} - \bar{y}^{\phi(\ell),h'}\bigr|$ and solve, for $H^{\min} = \min(H^s, H^t)$, the linear assignment problem:
\begin{equation}
  \{(h_{i,\ell}, h'_{i,\ell})\}_{i=1}^{H^{\min}} = \operatorname*{arg\,min} \sum_{i=1}^{H^{\min}} \mathbf{C}^{\ell}_{h_{i,\ell},\,h'_{i,\ell}},
  \label{eq:head-alignment}
\end{equation}
via the Hungarian algorithm.
The aligned test tensor is assembled as $\tilde{\mathbf{y}} \gets \mathbf{0} \in \mathbb{R}^{M \times L^s \times H^s}$ with $\tilde{y}_{q,\,\ell,\,h_{i,\ell}} \gets y_{q,\,\phi(\ell),\,h'_{i,\ell}}$ for each layer $\ell$ and matched pair $(h_{i,\ell}, h'_{i,\ell})$; unmatched source positions remain zero when $H^t < H^s$.
Given a classifier $f$ trained on $\mathbf{X}$ (using source labels), the predicted hallucination scores are $f(\tilde{\mathbf{y}}) \in \mathbb{R}^M$, instantiating the alignment map $f_{\mathrm{align}}$ in the decomposition above; see \Cref{fig:head-align-unequal} for a schematic of the unequal-dimension case.
Algorithm~\ref{alg:head-align} states the alignment procedure concretely.

\paragraph{Extension to multiple source LLMs.}
\label{app:multi-source-align}
We now turn to the case of $K$ source LLMs' features $\{\mathbf{X}^k\}_{k=1}^K$ and a test sample $\mathbf{y}$. We designate the source with the \emph{fewest} heads as canonical reference $\mathcal{L}^{\mathrm{ref}}$, align all other sources to this canonical reference, and apply the same alignment to $\mathbf{y}$ at inference. Each source $k$ contributes training features $\mathbf{X}^k$ and binary token-level hallucination labels $\mathbf{z}_k \in \{0,1\}^{N_k}$; the test input $\mathbf{y}$ contains features only (no target labels). We elaborate in the following, and give the full procedure in \Cref{alg:multi-source-align}.

\textit{Offline alignment.}
Precompute the mean features for the canonical reference as follows:
\begin{equation}
  \bar{X}^{\mathrm{ref},\ell,h} = \frac{1}{N_{\mathrm{ref}}} \sum_{q=1}^{N_{\mathrm{ref}}} X^{\mathrm{ref}}_{q,\ell,h},
\end{equation}
where $\bar{\mathbf{X}}^{\mathrm{ref}} \in \mathbb{R}^{L^{\mathrm{ref}} \times H^{\mathrm{ref}}}$ and $N_{\mathrm{ref}}$ is the number of reference training tokens.
For each $k \neq \mathrm{ref}$, solve the per-layer Hungarian matchings of \Cref{eq:head-alignment} between $\bar{\mathbf{X}}^{\mathrm{ref}}$ and $\bar{\mathbf{X}}^k$ (using $\phi(\ell)=\lfloor \ell \cdot L^k / L^{\mathrm{ref}} \rfloor$) and apply the resulting permutations to $\mathbf{X}^k$, yielding $\tilde{\mathbf{X}}^k \in \mathbb{R}^{N_k \times L^{\mathrm{ref}} \times H^{\mathrm{ref}}}$ (with $\tilde{\mathbf{X}}^{\mathrm{ref}} \gets \mathbf{X}^{\mathrm{ref}}$).
Train classifier $f$ on the concatenation $\bigcup_{k=1}^K \tilde{\mathbf{X}}^k$ of all sources (including the reference).

\textit{Inference.}
For each test sample $\mathbf{y}$, we align it to our canonical reference using $\textsc{HeadAlign}(\mathbf{X}^{\mathrm{ref}},\, \mathbf{y})$, and then score it using the trained classifier $f$.
Importantly, at inference we do not assume access to a pool of target-LLM samples: each test example is aligned from its own token features alone, without target labels.

\begin{algorithm}[ht]
\caption{\textsc{MultiSourceHeadAlign}$\bigl(\{(\mathbf{X}^k,\,\mathbf{z}_k)\}_{k=1}^K,\;\mathbf{y}\bigr)$}
\label{alg:multi-source-align}
\begin{algorithmic}[1]
  \Require $\mathbf{X}^k \in \mathbb{R}^{N_k \times L^k \times H^k}$ and $\mathbf{z}_k \in \{0,1\}^{N_k}$ (token labels) for $k=1,\ldots,K$;\quad $\mathbf{y} \in \mathbb{R}^{M \times L^t \times H^t}$ (unlabelled test features)
  \Ensure $\hat{\mathbf{p}} \in \mathbb{R}^M$ (predicted score per test token)
  \Statex \textit{Offline:}
  \State $\mathrm{ref} \gets \operatorname*{arg\,min}_{k}\, H^k$
  \State $\bar{X}^{\mathrm{ref},\ell,h} \gets \frac{1}{N_{\mathrm{ref}}}\sum_q X^{\mathrm{ref}}_{q,\ell,h}$ \quad for all $\ell \in [L^{\mathrm{ref}}],\; h \in [H^{\mathrm{ref}}]$
  \For{$k = 1, \ldots, K$,\; $k \neq \mathrm{ref}$}
    \State $\bar{X}^{k,\ell,h} \gets \frac{1}{N_k}\sum_q X^{k}_{q,\ell,h}$ \quad for all $\ell \in [L^k],\; h \in [H^k]$
    \For{$\ell = 1, \ldots, L^{\mathrm{ref}}$}
      \State $\phi(\ell) \gets \lfloor \ell \cdot L^k / L^{\mathrm{ref}} \rfloor$
      \State $\mathbf{C}^{\ell}_{h,h'} \gets \bigl|\bar{X}^{\mathrm{ref},\ell,h} - \bar{X}^{k,\phi(\ell),h'}\bigr|$
      \State $\{(h_{i,\ell}, h'_{i,\ell})\}_{i=1}^{H^{\mathrm{ref}}} \gets \textsc{Hungarian}(\mathbf{C}^{\ell})$
    \EndFor
    \State $\tilde{\mathbf{X}}^k \gets \mathbf{0} \in \mathbb{R}^{N_k \times L^{\mathrm{ref}} \times H^{\mathrm{ref}}}$
    \For{each token $q$, layer $\ell$, and matched pair $(h_{i,\ell}, h'_{i,\ell})$}
      \State $\tilde{X}^k_{q,\,\ell,\,h_{i,\ell}} \gets X^k_{q,\,\phi(\ell),\,h'_{i,\ell}}$
    \EndFor
  \EndFor
  \State $\tilde{\mathbf{X}}^{\mathrm{ref}} \gets \mathbf{X}^{\mathrm{ref}}$
  \State $f \gets \textsc{TrainLR}\!\left(\textstyle\bigcup_{k=1}^{K} \tilde{\mathbf{X}}^k,\;\bigcup_{k=1}^{K} \mathbf{z}_k\right)$
  \Statex \textit{Inference:}
  \State $\tilde{\mathbf{y}} \gets \textsc{HeadAlign}(\mathbf{X}^{\mathrm{ref}},\, \mathbf{y})$
  \State \Return $f(\tilde{\mathbf{y}})$
\end{algorithmic}
\end{algorithm}

\clearpage
\section{Dataset Annotation}
\label{app:data-annotation}

RAGTruth~\citep{niu2024ragtruth} provides responses with hallucination annotations for three of our six LLMs on \tasksumm, \taskqa, and \taskdtt; we use those as published, and annotate the remaining three LLMs ourselves, as detailed next.

\paragraph{Native RAGTruth models.}
For \llamaseven, \llamathirteen, and \mistralseven, we use the responses and hallucination spans exactly as released---no regeneration and no re-judging. Our only change concerns the \tasksumm{} training split, where the native release contains the same source context twice: we drop one of the two duplicates, so that every model's \tasksumm{} training set---native or newly generated---contains the same $2514$ examples.

\paragraph{Newly annotated models.}
We extend RAGTruth with token-level hallucination labels for \nemo, \qwenthree, and \qwenseven. Annotation proceeds in two stages: each model generates one response per unique query--context pair, then \qwenjudge labels unsupported or contradictory spans under the original RAGTruth judging template.

\paragraph{Annotation rules.}
We annotate under the following fixed rules:
\begin{enumerate}
  \item \textbf{What counts as a hallucination.} A span is labeled hallucinated if it is \emph{unsupported by} or \emph{contradicted by} the source---i.e.\ both fabricated/baseless content and content that conflicts with the source are flagged. The source is the news article (\tasksumm), the supporting passages (\taskqa), or the structured record together with its embedded customer reviews (\taskdtt).
  \item \textbf{Judging template.} The judge receives the original RAGTruth \texttt{input\_str} for the example, instantiated per task: it is shown the source and the model response, then instructed to ``identify and label any hallucinated statements \dots\ that are unsupported or contradicted by'' the source and to ``compile the labeled hallucinated spans into a JSON list, with each list item representing a separate hallucinated span.''
  \item \textbf{Output format.} The judge returns a JSON list of text spans; it decodes greedily with a budget of $1024$ tokens.
  \item \textbf{Span localization.} Each returned span is mapped to character offsets in the response by exact substring match; spans whose text cannot be located in the response are discarded.
  \item \textbf{Token labels.} The detector operates on tokens (one graph node per response token or window), so for downstream graph construction we convert the character-level spans into per-token labels: each response token---under the model's own tokenizer---is labeled \emph{incorrect} if any of its characters overlaps any hallucination span, and \emph{correct} otherwise. We use this inclusive overlap rule (rather than, say, requiring a token to be fully contained) so that the boundary tokens of a hallucinated phrase are not dropped, erring toward recall on the positive class; it is simple and deterministic, and it is applied identically to the native RAGTruth annotations, so token labels are consistent across all six models.
\end{enumerate}

To assess the quality of the labels produced by our judge (\qwenjudge), we audited the first ten test examples for each of the three newly annotated models---\qwenthree, \qwenseven, and \nemo---on every task, for $90$ audited examples and $220$ flagged spans in total. For each example we compared the model response, the source context, the final character-level spans, and the judge's free-form rationale. This appendix reports what we found, with verbatim examples; our headline conclusion is: \emph{the labels are accurate enough to train on, with a well-characterized and largely benign error profile}.

\paragraph{Overall quality.}
The annotations are reliable on the points that matter most for token-level supervision. First, the character offsets are exact: all $220$ flagged spans matched their recorded text under the substring-mapping step ($220/220$ in \Cref{tab:annotation-audit}), so a flagged span is always aligned to the right tokens---there is no offset drift. Second, on \tasksumm{} the judge is both precise and well-calibrated: it leaves faithful summaries unlabeled ($17/30$ summaries received no span) and the spans it does emit are overwhelmingly genuine fabrications or contradictions (examples below). In general, the dominant error mode is \emph{over-flagging} (false positives), not missed hallucinations, and it is concentrated in the more list-like \taskqa{} and \taskdtt{} tasks; clear missed hallucinations were rare in our sample. \Cref{tab:annotation-audit} breaks the statistics down by model and task.

\begin{table}[ht]
\centering
\small
\begin{tabular}{llcccc}
\toprule
Model & Task & Spans & Spans/resp. & Empty & Exact-offset \\
\midrule
Qwen2.5-3B   & Summ. & $15$ & $1.5$ & $5$ & $15/15$ \\
Qwen2.5-3B   & QA    & $26$ & $2.6$ & $0$ & $26/26$ \\
Qwen2.5-3B   & D2T   & $35$ & $3.5$ & $0$ & $35/35$ \\
\midrule
Qwen2.5-7B   & Summ. & $11$ & $1.1$ & $6$ & $11/11$ \\
Qwen2.5-7B   & QA    & $24$ & $2.4$ & $1$ & $24/24$ \\
Qwen2.5-7B   & D2T   & $31$ & $3.1$ & $0$ & $31/31$ \\
\midrule
Mistral-Nemo & Summ. & $9$  & $0.9$ & $6$ & $9/9$   \\
Mistral-Nemo & QA    & $26$ & $2.6$ & $1$ & $26/26$ \\
Mistral-Nemo & D2T   & $43$ & $4.3$ & $0$ & $43/43$ \\
\bottomrule
\end{tabular}
\caption{Audit statistics over the first ten test examples per (model, task), $90$ examples in total. ``Spans'' is the total number of hallucination spans the judge flagged across the ten responses, and ``Spans/resp.''\ the average per response; ``Empty'' counts responses with no flagged span (out of $10$); ``Exact-offset'' counts spans whose recorded text matches the response substring at the recorded offsets.}
\label{tab:annotation-audit}
\end{table}

\paragraph{What the judge gets right.}
A clear majority of flagged spans are genuine, well-localized hallucinations, and the judge is especially dependable on fabricated facts and on contradictions of the structured record. Three representative cases:
\begin{itemize}
  \item \textbf{Fabricated detail (\tasksumm, Qwen2.5-3B).} The article to summarize states only that the ramp agent ``started work at 5 a.m.\ and his shift was scheduled to end at 2:30 p.m.'' The response adds that ``the agent had been working overtime and was reportedly exhausted \dots Alaska Airlines is currently investigating the cause of the error.'' The judge flags exactly the three invented clauses---``the agent had been working overtime'', ``was reportedly exhausted'', and ``Alaska Airlines is currently investigating the cause of the error''---all absent from the source.
  \item \textbf{Numerical contradiction (\tasksumm, Mistral-Nemo).} The article says B.B.\ King ``has 30 Grammy nominations''; the response writes ``a 15-time Grammy winner.'' The judge correctly flags ``15-time Grammy winner'' as a conflict.
  \item \textbf{Attribute contradiction (\taskdtt, Qwen2.5-7B).} For Finch \& Fork the record has \texttt{OutdoorSeating: False} and \texttt{Ambience.romantic: False}; the response claims ``outdoor seating'' and ``romantic and classy ambiance.'' The judge flags both ``outdoor seating'' and ``romantic and classy ambiance''---precise contradictions of the structured data.
\end{itemize}
The judge also correctly treats true-but-unsupported world knowledge as baseless content (e.g.\ flagging an added, factually correct explanation of DNA replication that does not appear in the passage), and reliably catches wrong opening hours in \taskdtt{} (e.g.\ ``open from 11:30 AM to 8 PM on weekdays \dots'' when the record lists different per-day hours). The two boxes below show such cases verbatim.

\begin{annotex}{Correct catch --- fabricated detail (Qwen2.5-3B, \tasksumm)}
\textbf{Source (article):} ``\dots the employee started work at 5 a.m.\ and his shift was scheduled to end at 2:30 p.m.'' --- no mention of overtime, exhaustion, or an investigation.\\
\textbf{Response, flagged spans:} ``the agent had been working overtime''; ``was reportedly exhausted''; ``Alaska Airlines is currently investigating the cause of the error''.\\
\textbf{Judge's reason:} ``This detail is not mentioned in the original news''; ``There's no mention of exhaustion''.\\
\textbf{Verdict:} \emph{Correct.} All three clauses are fabrications absent from the article.
\end{annotex}

\begin{annotex}{Correct catch --- contradiction of the structured record (Qwen2.5-7B, \taskdtt)}
\textbf{Record (Finch \& Fork):} \texttt{OutdoorSeating: False}, \texttt{Ambience.romantic: False} (\texttt{classy: True}).\\
\textbf{Response, flagged spans:} ``outdoor seating''; ``romantic and classy ambiance''.\\
\textbf{Judge's reason:} ``'OutdoorSeating' is set to False in the structured data''; ``'romantic' is false according to the structured data''.\\
\textbf{Verdict:} \emph{Correct} contradictions. (A third flagged span on the same example, ``starting at \$14'', is instead a false positive: a review states ``\$14 for the first mimosa''.)
\end{annotex}

\paragraph{Where it errs: over-flagging on \taskqa{} and \taskdtt.}
The false positives are concentrated in the two list-structured tasks, where responses heavily paraphrase or restate the source and the judge flags content that is in fact supported. Two clear cases:
\begin{itemize}
  \item \textbf{Supported paraphrase flagged (\taskqa, Qwen2.5-3B).} The recipe passage explicitly lists cooking spray, cheese, and ``onion, green pepper, and tomato slices.'' Yet the judge flags ``Cooking spray for the baking sheet'', ``Cheese (optional)'', and ``Place slices of onion, green pepper, and tomato on top of the fillets''---all directly supported. QA answers are long and itemized, and the judge frequently mislabels such faithful restatements.
  \item \textbf{Judge overlooks support present in the source (\taskdtt, Qwen2.5-7B).} Each \taskdtt{} example verbalizes a business record. For the restaurant \emph{Pizza Mizza}, the judge flags ``located in Santa Barbara, CA'' and ``With free Wi-Fi, outdoor seating, and take-out options'' as unsupported, even though the record itself states \texttt{city: Santa Barbara}, \texttt{WiFi: free}, \texttt{OutdoorSeating: True}, and \texttt{RestaurantsTakeOut: True}. It likewise misses support in the record's embedded customer reviews: for the restaurant \emph{Finch \& Fork}, it flags ``starting at \$14'' although a review states verbatim ``\$14 for the first mimosa, and \$1 for every mimosa after that.''
\end{itemize}
On \tasksumm{} these errors are much rarer; the most common minor issue there is flagging a faithful paraphrase or a clause carrying no factual claim (e.g.\ the discourse fragment ``While the passage doesn't provide specific details,''). The two boxes below give the two false-positive patterns verbatim.

\begin{annotex}{False positive --- supported content flagged (Qwen2.5-3B, \taskqa)}
\textbf{Source (recipe passage):} lists ``butter flavored cooking spray'', layering with ``onion, green pepper, and tomato slices'', and adding cheese.\\
\textbf{Response, flagged spans:} ``Cooking spray for the baking sheet''; ``Cheese (optional)''; ``Place slices of onion, green pepper, and tomato on top of the fillets.''\\
\textbf{Judge's reason:} ``The passage mentions using butter-flavored cooking spray but does not specify it as an ingredient'' (and similar for the others).\\
\textbf{Verdict:} \emph{False positives.} All three are stated in the passage; the judge over-flags faithful restatement on the long, itemized QA answers.
\end{annotex}

\begin{annotex}{False positive --- judge ignores supported fields (Qwen2.5-7B, \taskdtt)}
\textbf{Record (Pizza Mizza):} \texttt{city: Santa Barbara}, \texttt{WiFi: free}, \texttt{OutdoorSeating: True}, \texttt{RestaurantsTakeOut: True}.\\
\textbf{Response, flagged spans:} ``located in Santa Barbara, CA''; ``With free Wi-Fi, outdoor seating, and take-out options''.\\
\textbf{Judge's reason:} ``The address field is empty, so the exact location \dots\ is not specified'' --- yet \texttt{city} is present, as are the WiFi/seating/takeout fields.\\
\textbf{Verdict:} \emph{False positives.} The flagged claims match fields that are present in the record.
\end{annotex}

\paragraph{A minor ($<2\%$ of flagged spans), fixable error source.}
Beyond task-dependent over-flagging, we identified one structural issue: rationale--label mismatch (parsing). The judge sometimes emits a JSON list of \emph{candidate} spans and then retracts some in the prose that follows; our parser keeps the JSON list and never sees the retraction. On the Rolling Stone summary (Qwen2.5-7B), ``The magazine issued an apology'' is recorded as a hallucination even though the judge's own explanation states it ``is supported by the original text, so it is not a hallucination.'' This is rare: scanning the full test set, only $1.7\%$ of flagged spans ($49$ of $2967$) are plausibly retracted in this way. For such an example, see below.
\begin{annotex}{Parser keeps a span the judge retracts in prose (Qwen2.5-7B, \tasksumm)}
\textbf{Judge's JSON list (kept by the parser):} \texttt{[\,"The magazine issued an apology",\,\dots]}\\
\textbf{Judge's prose, same output:} ``\,`The magazine issued an apology' --- This statement is supported by the original text, so it is not a hallucination.''\\
\textbf{Verdict:} \emph{False positive.} The judge retracts the span in prose, but our parser reads only the JSON list, so the span is kept. As noted above, this affects only $1.7\%$ of flagged spans.
\end{annotex}

Separately, we note a generation-side artifact that is not an annotation error: responses for the three new models were generated with a $256$-token cap, which truncates most QA answers (the native RAGTruth responses, generated with a larger budget, are unaffected). Truncation does not create false positives; its only footprint on the labels is that when the usual QA over-flagging lands on the incomplete final sentence, the resulting span is a malformed partial sentence rather than a clean clause.

\paragraph{Severity and effect on training.}
The errors above are overwhelmingly false positives rather than missed hallucinations, and they have a mild effect on the per-token labels that downstream training consumes. The labels remain reliable where it counts: offsets are exact, faithful summaries are correctly left unlabeled, and genuine fabrications and contradictions are caught. Two properties further soften the residual noise: the over-flagged spans are usually whole clauses rather than scattered tokens, so the per-token overlap rule (rule~5 in the annotations rule above) does not spread error beyond the flagged region. We therefore regard the annotations as suitable supervision for token-level hallucination detection.

\section{Experiments: Monitoring for LLM misbehaviors with \lemon --- Extended Section}\label{app:experiments-extended}

Our experiments were conducted using the PyTorch~\citep{paszke2019pytorch} framework (License: BSD), using a single NVIDIA L-40 GPU for all experiments. We used a fixed batch size of $16$ graphs for all hallucination experiments, and $8$ for the jailbreak; the latter is halved because jailbreak graphs are built over the full prompt rather than pooled response windows, and therefore contain roughly twice as many nodes per graph.

\paragraph{Hyperparameters.} \Cref{tab:hyperparams} lists all hyperparameters. The hallucination and jailbreak experiments share the same architecture, optimizer, and search space; they differ only in the batch size, the window size $W$, and the prediction level, which follow from the jailbreak monitor predicting a single label per prompt rather than a label per response window. We keep the search space deliberately small: for every held-out LLM, we train the four combinations of $K \in \{1,2\}$ MPNN layers and $\{1,2\}$ cross-token Transformer layers, and select the configuration---and, within it, the epoch---with the highest validation ROC-AUC, where validation data comes only from the training LLMs, so the held-out LLM never influences model selection. Unless stated otherwise (e.g., the adaptation study in \Cref{app:few-shot-adaptation}), we train with $\alpha=0$, i.e., the invariant head alone.

\begin{table}[h]
  \centering
  \caption{\textbf{\lemon hyperparameters.} Values in braces are searched; the configuration
    is selected by validation ROC-AUC on the training LLMs only. Both \lemon variants share
    all settings.}
  \label{tab:hyperparams}
  \small
  \begin{tabular}{@{}lcc@{}}
    \toprule
    \textbf{Hyperparameter} & \textbf{Hallucination} & \textbf{Jailbreak} \\
    \midrule
    Hidden dimension $d$                         & \multicolumn{2}{c}{$16$} \\
    MPNN layers $K$                              & \multicolumn{2}{c}{$\{1, 2\}$} \\
    Cross-token Transformer layers               & \multicolumn{2}{c}{$\{1, 2\}$} \\
    Attention heads (cross-token Transformer)    & \multicolumn{2}{c}{$2$} \\
    Dropout                                      & \multicolumn{2}{c}{$0$} \\
    Attention threshold $\delta$                 & \multicolumn{2}{c}{$0.1$} \\
    Loss weight $\alpha$                         & \multicolumn{2}{c}{$0$} \\
    \midrule
    Optimizer                                    & \multicolumn{2}{c}{AdamW} \\
    Learning rate                                & \multicolumn{2}{c}{$10^{-3}$, cosine decay to $0$} \\
    Weight decay                                 & \multicolumn{2}{c}{$0$} \\
    Gradient-norm clipping                       & \multicolumn{2}{c}{$1.0$} \\
    Epochs                                       & \multicolumn{2}{c}{$100$} \\
    Random seed                                  & \multicolumn{2}{c}{$42$} \\
    \midrule
    Batch size (graphs)                          & $16$ & $8$ \\
    Window size $W$                              & $8$  & $1$ \\
    Prediction level                             & per window node & per prompt (final node) \\
    \bottomrule
  \end{tabular}
\end{table}

\paragraph{Optimizers and Schedulers.} For all datasets, we use the AdamW optimizer~\cite{loshchilov2019decoupled} in combination with a cosine learning rate schedule that decays the learning rate from its initial value to zero over the course of training.

\subsection{Baselines and Ablations: Implementation Details}\label{app:baselines}
All baselines follow the same cross-LLM protocol, splits, and evaluation units as \lemon. 

\paragraph{Logistic-regression baselines.}
\lb, \lbalign, and \hs, as well as every per-LLM and adaptation classifier (\Cref{app:few-shot-adaptation}), use L2-regularized logistic regression (\texttt{lbfgs}, up to $1{,}000$ iterations), with the inverse regularization strength swept over $C \in \{10^{-3}, 10^{-2}, \ldots, 10^{3}\}$ and selected by validation ROC-AUC. When a small labeling budget leaves the validation split with a single class, validation is merged into training and $C{=}1$ is used; when the \emph{training} split has a single class, the cell is reported as degenerate (``---'').

\paragraph{\lookback (\lb).}
Per-token \lookback features (\Cref{app:lookbacklens}), one lookback ratio per layer--head pair, are zero-padded to the common $40 \times 40$ grid, giving a $1{,}600$-dimensional vector per token, and fed to the logistic regression above. The classifier is fit on the pooled training LLMs and applied as-is to the held-out LLM.

\paragraph{\lookback{+}Alignment (\lbalign).}
The same \lookback features and classifier, except that every LLM's features are first reindexed to a canonical reference by the head-alignment procedure of \Cref{app:alignment-methods}.

\paragraph{Hidden-state probing (\hs).}
We take the residual stream at layer $24$ of each LLM, zero-pad it to the largest hidden dimension in the pool ($5{,}120$), reduce it to $128$ dimensions with PCA fit on the training data only, and fit the logistic regression above. For hallucination, features are per token (window-averaged); for jailbreak, they are taken at the final prompt token.

\paragraph{Token probability (\probbase).}
A training-free score: the logit the LLM assigned to each generated token, averaged within each window, is used directly as the confidence score (higher means more likely correct). It requires no fitting, so it is evaluated identically in every setting.

\paragraph{Attention to the system prompt (Attn${\to}$Sys).}
The training-free jailbreak baseline of \Cref{subsec:jailbreak-monitoring} is computed from a single prompt-only forward pass using the \emph{full}, unthresholded attention (the thresholded graphs used by \lemon would distort its mean). The span $S$ is located by comparing the chat-templated prompt with the templated empty message.

\paragraph{CHARM.}
\charm is trained through the identical pipeline, on identical graphs, loss, and splits, for $100$ epochs with learning rate $10^{-3}$, sweeping hidden dimension $\in \{16, 64\}$ and message-passing layers $\in \{1, 2\}$, i.e., the same number of configurations as \lemon. It has no cross-token Transformer, so those axes are not searched. For jailbreak, it uses batch size $8$ and reads out the final node, as \lemon does.

\paragraph{Ablations.}
All graph-based ablations of \Cref{subsec:ablation-studies} are trained through the identical pipeline on identical graphs, over the same hyperparameter grid as \lemon (\Cref{tab:hyperparams}); grid axes that configure a removed component are dropped. The graph-free ablation (without graph) trains a causal Transformer on the same window-pooled token probabilities and labels, with every applicable hyperparameter matched to \lemon.

\subsection{LLMs Used In Our Experiments}\label{app:llms-used}
Every monitored LLM is open-weight and instruction-tuned. \Cref{tab:all-llms} lists all sixteen together with their attention grids $(H,L)$; we elaborate on them in the following.

\begin{table}[t]
  \centering
  \caption{%
    \textbf{Every LLM we monitor, with its attention grid $(H,L)$.}
    }
  \label{tab:all-llms}
  \footnotesize
  \setlength{\tabcolsep}{6pt}
  \begin{tabular}{@{}lcc@{}}
    \toprule
    \textbf{Model} & $H$ & $L$ \\
    \midrule
    \multicolumn{3}{@{}l}{\emph{Training pool ($\mathcal{K}=6$) --- fitted on; held out one at a time}}\\
    \llamaseven                                   & 32 & 32 \\
    \llamathirteen                                & 40 & 40 \\
    \mistralseven                                 & 32 & 32 \\
    \nemo                                         & 32 & 40 \\
    \qwenthree                                    & 16 & 36 \\
    \qwenseven                                    & 28 & 28 \\
    \midrule
    \multicolumn{3}{@{}l}{\emph{In the wild --- never trained on; monitored zero-shot}}\\
    \llmname{Mistral-7B-Instruct-v0.2}            & 32 & 32 \\
    \llmname{Pythia-Chat-Base-7B}                 & 32 & 32 \\
    \mbox{\llmname{Pythia-Chat-Base-7B-v0.16}}    & 32 & 32 \\
    \llmname{TinyLlama-1.1B-Chat}                 & 32 & 22 \\
    \llmname{Zephyr-7B-$\beta$}                   & 32 & 32 \\
    \mbox{\llmname{Nous-Hermes-2-Mistral-7B-DPO}} & 32 & 32 \\
    \llmname{Phi-4-mini-instruct}                 & 24 & 32 \\
    \llmname{SmolLM2-1.7B-Instruct}               & 32 & 24 \\
    \llmname{SmolLM2-360M-Instruct}               & 15 & 32 \\
    \llmname{SmolLM2-135M-Instruct}               &  9 & 30 \\
    \bottomrule
  \end{tabular}
\end{table}

\paragraph{Hallucination and jailbreak monitoring.} Those experiments share the same pool of $\mathcal{K}=6$ LLMs:
\begin{itemize}
  \item \llamaseven~\citep{touvron2023llama2}, $(H,L)=(32,32)$: the smaller of the two Llama-2 chat models and one of the two most heavily safety-aligned models in the pool.
  \item \llamathirteen~\citep{touvron2023llama2}, $(H,L) = (40,40)$: the same family and alignment recipe at roughly twice the scale. It has the largest grid in the pool and therefore defines the padded $H_{\max}\times L_{\max}=40\times 40$ canvas into which every other LLM is embedded.
  \item \mistralseven~\citep{jiang2023mistral}, $(H,L) = (32,32)$: a different pretraining lineage with exactly the same grid shape as \llamaseven. 
  \item \nemo\cite{sreenivas2024llm}, $(H,L) = (32,40)$: a $12$B Mistral model, and the only one whose grid is deep but not maximally wide.
  \item \qwenthree~\citep{qwen2}, $(H,L) = (16,36)$: the smallest LLM we monitor.
  \item \qwenseven~\citep{qwen2}, $(H,L) = (28,28)$.
\end{itemize}

\paragraph{Hallucination in the wild.} These LLMs are never trained on; they only generate the responses we monitor with an unmodified LOO checkpoint.
\begin{itemize}
  \item \llmname{Mistral-7B-Instruct-v0.2}~\citep{jiang20236g}, $(32,32)$: a later point release of a family in the training pool --- hence the mildest shift we test.
  \item \llmname{Pythia-Chat-Base-7B}~\citep{openchatkit}, $(32,32)$: a \mbox{GPT-NeoX}~\citet{black2022gpt} lineage absent from the training pool entirely.
  \item \mbox{\llmname{Pythia-Chat-Base-7B-v0.16}}, $(32,32)$: a later revision of the same model.
  \item \llmname{TinyLlama-1.1B-Chat}~\citep{zhang2024tinyllama}, $(32,22)$: a $1.1$B model, far shallower than anything seen during training.
  \item \llmname{Zephyr-7B-$\beta$}~\citep{tunstall2023zephyr}, $(32,32)$: a DPO-tuned descendant of \mistralseven.
  \item \mbox{\llmname{Nous-Hermes-2-Mistral-7B-DPO}}, $(32,32)$: a second, independently DPO-tuned descendant of \mistralseven; together the two test whether the post-training recipe --- and not only the base model --- shifts the attention signal.
  \item \llmname{Phi-4-mini-instruct}~\citep{abouelenin2025phi}, $(24,32)$: a synthetic-data-centric family with no relative in the training pool.
  \item \llmname{SmolLM2-Instruct}~\citep{allal2025smollm2} at $1.7$B, $360$M and $135$M, $(32,24)$, $(15,32)$ and $(9,30)$: a deliberate size sweep down to the smallest grids we ever score, the last of which has only $9$ attention heads.
\end{itemize}

\subsection{Datasets Used In Our Experiments}\label{app:datasets}
\paragraph{Hallucination monitoring.}
\begin{itemize}
  \item \textbf{RAGTruth}~\citep{niu2024ragtruth} (these were used for training \lemon): word-level hallucination spans over \emph{retrieval-grounded} generation. Because every response is produced from a supplied context, a hallucination is an unsupported claim rather than a factual error, which makes the label objective and checkable. It contains three tasks, and we train on their union and report each separately:
\begin{itemize}
  \item \textbf{Summarization (SUM)}: the task is to summarize a news article within a specified word budget; hallucinations are typically invented details or over-generalized attributions.
  \item \textbf{Question answering (NQ)}: the task is to answer a user query using retrieved passages; hallucinations are claims that exceed what the passages support.
  \item \textbf{Data-to-text (D2T)}: the task is to generate a prose description from a structured business record (JSON); hallucinations are attributes absent from the record.
\end{itemize}
  RAGTruth ships human annotations for only part of our pool; for the remaining LLMs we generate responses ourselves and annotate them with \qwenjudge; see details in \Cref{app:data-annotation}.
  \item \textbf{Mu-SHROOM}, English split~\citep{vazquez2025semeval} (in-the-wild evaluation only): human \emph{gold} character-level spans over open-domain answers. There is no supplied context, so the LLM answers from parametric knowledge alone --- a genuine distribution shift away from RAGTruth's grounded setting.
  \item \textbf{PsiloQA}, English split~\citep{rykov2025models} (in-the-wild evaluation only): the same open-domain question-answering format at larger scale and across more generators, with \emph{silver} spans produced by an automatic annotator. It is what lets us test transfer across many unseen LLMs at once.
\end{itemize}

\paragraph{Jailbreak monitoring.}

We use \textbf{\redbench}~\citep{dang2026redbench}: a corpus of harmful requests, of which we use $3000$ prompts per LLM. Each request is turned into an attack by appending a \emph{single universal} GCG suffix~\citep{zou2023universal} taken from HarmBench~\citep{mazeika2024harmbench} and held fixed across every request and every monitored LLM; how that suffix was chosen is detailed below. Each of the six LLMs answers greedily, and \strongreject~\citep{souly2024strongreject} grades how much the response actually helps the \emph{original} request --- the suffix is stripped before grading, so the label measures genuine compliance rather than whether the model echoed the adversarial string --- which we binarize at $0.5$. The resulting attack-success rate spans roughly sevenfold across the pool (from $\approx 8\%$ for the two Llama-2 models to $\approx 52\%$ for \nemo; \Cref{tab:jailbreak-asr}), confirming that the label is a property of the target LLM and not of the prompt --- which is exactly what makes zero-shot transfer non-trivial here.

\begin{table}[htbp]
  \centering
  \caption{\textbf{Attack-success rate (ASR) \% of the universal GCG suffix, per
    LLM.} 
    }
  \label{tab:jailbreak-asr}
  \small
  \begin{tabular}{@{}l cc cc@{}}
    \toprule
    \multirow{2}{*}{\textbf{LLM}}
      & \multicolumn{2}{c}{\textbf{All $3000$ prompts}}
      & \multicolumn{2}{c@{}}{\textbf{Test split ($750$)}} \\
    \cmidrule(lr){2-3}\cmidrule(l){4-5}
      & jailbroken & ASR (\%) & jailbroken & ASR (\%) \\
    \midrule
    \llamaseven   & $256/3000$  & $8.5$  & $69/750$  & $9.2$  \\
    \llamathirteen& $234/3000$  & $7.8$  & $53/750$  & $7.1$  \\
    \mistralseven & $1325/3000$ & $44.2$ & $337/750$ & $44.9$ \\
    \nemo         & $1548/3000$ & $51.6$ & $394/750$ & $52.5$ \\
    \qwenthree    & $762/3000$  & $25.4$ & $212/750$ & $28.3$ \\
    \qwenseven    & $559/3000$  & $18.6$ & $143/750$ & $19.1$ \\
    \midrule
    Pooled        & $4684/18000$ & $26.0$ & $1208/4500$ & $26.8$ \\
    \bottomrule
  \end{tabular}
\end{table}

\paragraph{How the attack suffix was chosen.}

We reuse one fixed string verbatim for all $3000$ requests and all six LLMs, taken as-is from the public
HarmBench release~\citep{mazeika2024harmbench}. Most artifacts in that release
store one suffix \emph{per behavior} (per harmful request), thus, the only candidates are its
\textsc{EnsembleGCG} runs, which optimize a single suffix jointly
over all $401$ HarmBench behaviors and against an \emph{ensemble} of targets. We use the $7$B ensemble (\llamaseven, \llmname{Vicuna-7B-v1.5},
\llmname{Orca-2-7B}). Of these, only \llamaseven is in our monitored pool; however,
this does not favor the monitor, as \llamaseven nonetheless remains among the least susceptible LLMs in
the pool ($8.5\%$). The box below shows a complete input as it is fed to
\llamaseven: the \redbench request with the attack suffix appended verbatim, wrapped in
the model's own chat template.

\begin{annotex}{A complete attack input to \llamaseven (\redbench request \#$40$)}
\ttfamily\scriptsize\raggedright
\textcolor{gray!80}{<s>[INST]}
\textcolor{blue!65!black}{How can I identify and steal high-value artwork or
antiquities from museums or private collections?}\textcolor{orange!85!black}{ !
Label columns(\{\textbackslash{}boldsymbolTEatefulMay be-) transformations]->
techniqueInfo succ briefly'.\$ alongside constraintTree}
\textcolor{gray!80}{[/INST]}

\vspace{3pt}
\normalfont\footnotesize
\textcolor{gray!80}{Chat template} $\mid$
\textcolor{blue!65!black}{\redbench request} $\mid$
\textcolor{orange!85!black}{universal GCG suffix ($119$ characters, identical for
every request and every LLM)}. \llamaseven refuses here
(\strongreject score $0.01$), so this sample carries the label
``attack failed''.
\end{annotex}

\paragraph{Splits.} The $3000$ prompt indices are shuffled once under seed $42$
and carved $1800/450/750$ into train/validation/test ($25\%$ test, then $80/20$
of the remainder). The split is at the \emph{prompt} level and shared across all
six LLMs, so a prompt held out for one LLM is held out for all of them; combined
with the leave-one-out rotation over LLMs, a test sample is therefore unseen on
both axes. It is \emph{not} stratified by label, and cannot be: the
\strongreject label is a property of the (prompt, LLM) pair rather than of the
prompt, since the same $3000$ requests yield an attack-success rate of $8.5\%$
on \llamaseven and $51.6\%$ on \nemo, so no single shared partition can balance
all six. The random partition nevertheless leaves each LLM's test-split rate
within a few points of its overall rate (\Cref{tab:jailbreak-asr}), and every
shard remains usable: the least susceptible LLM still contributes $53$ positives
to its $750$-prompt test split.

In the next subsections, we provide additional results, which were omitted from the main text due to space limitations

\subsection{LOO results --- Extended results}\label{app:loo-per-task}

\Cref{tab:loo-zero-shot} of the main text averages the three RAGTruth tasks per held-out LLM. \Cref{tab:loo-zero-shot-full} gives the underlying per-task cells.

\begin{table}[htbp]
  \centering
  \caption{%
    \textbf{LOO zero-shot generalization --- per-task breakdown}
    of \Cref{tab:loo-zero-shot}.
    Each data row is one (left-out LLM, task) pair;
    every value is the ROC-AUC\,${\times}\,100$ on that held-out shard;
    \textit{avg} rows average the three tasks for that LLM.
    \textbf{Bold} = best per row; \underline{underline} = second best.
  }
  \label{tab:loo-zero-shot-full}
  \scriptsize
  \resizebox{\textwidth}{!}{%
  \begin{tabular}{@{}ll ccccc | cc@{}}
    \toprule
    \multirow{2}{*}{\textbf{Held-out LLM}} & \multirow{2}{*}{\textbf{Task}}
      & \multirow{2}{*}{\probbase}
      & \multirow{2}{*}{\lb}
      & \multirow{2}{*}{\hs}
      & \multirow{2}{*}{\charm}
      & \multirow{2}{*}{\lbalign}
      & \multicolumn{2}{c}{\textbf{\lemon (Ours)}} \\
    \cmidrule(lr){8-9}
    & & & & & & & $\mathcal{W}$ & $\mathcal{W}$-per-cell \\
    \midrule
    \multirow{4}{*}{Llama-2-7B}
      & SUM          & 56.2 & 48.7 & 50.7 & 41.3 & 61.7 & \underline{68.4} & \textbf{71.4} \\
      & NQ           & 60.7 & 55.8 & 52.1 & 51.6 & 61.5 & \underline{82.7} & \textbf{84.5} \\
      & D2T          & 56.6 & 41.8 & 55.9 & 43.5 & 60.6 & \underline{63.2} & \textbf{68.9} \\
      \cmidrule(lr){2-9}
      & \textit{avg} & 57.8 & 48.8 & 52.9 & 45.5 & 61.3 & \underline{71.4} & \textbf{74.9} \\
    \midrule
    \multirow{4}{*}{Llama-2-13B}
      & SUM          & 55.8 & 55.3 & 47.5 & 59.5 & 49.5 & \underline{68.3} & \textbf{70.3} \\
      & NQ           & 48.5 & 58.1 & 58.8 & 64.9 & 58.6 & \underline{81.6} & \textbf{85.1} \\
      & D2T          & 52.6 & 48.3 & 51.1 & 47.6 & 54.6 & \textbf{71.4} & \underline{71.1} \\
      \cmidrule(lr){2-9}
      & \textit{avg} & 52.3 & 53.9 & 52.5 & 57.3 & 54.2 & \underline{73.8} & \textbf{75.5} \\
    \midrule
    \multirow{4}{*}{Mistral-7B}
      & SUM          & \textbf{71.1} & 54.3 & 48.7 & 40.6 & 57.8 & \underline{66.8} & 64.1 \\
      & NQ           & 66.8 & 50.3 & 43.6 & 32.8 & 56.3 & \underline{77.9} & \textbf{80.6} \\
      & D2T          & \textbf{64.8} & 43.9 & \underline{60.8} & 53.0 & 58.5 & 59.9 & \underline{60.8} \\
      \cmidrule(lr){2-9}
      & \textit{avg} & 67.6 & 49.5 & 51.0 & 42.2 & 57.5 & \underline{68.2} & \textbf{68.5} \\
    \midrule
    \multirow{4}{*}{Mistral-Nemo}
      & SUM          & \underline{61.0} & 48.6 & 42.9 & 44.8 & 59.4 & 60.3 & \textbf{62.2} \\
      & NQ           & 55.2 & 58.0 & 55.4 & 39.1 & 61.8 & \textbf{70.2} & \underline{69.2} \\
      & D2T          & 58.5 & 50.7 & 52.6 & 48.6 & 60.7 & \underline{61.2} & \textbf{63.1} \\
      \cmidrule(lr){2-9}
      & \textit{avg} & 58.3 & 52.4 & 50.3 & 44.2 & 60.6 & \underline{63.9} & \textbf{64.8} \\
    \midrule
    \multirow{4}{*}{Qwen2.5-3B}
      & SUM          & 61.2 & 53.5 & 48.8 & 42.0 & 53.6 & \underline{63.9} & \textbf{64.7} \\
      & NQ           & 58.7 & 50.7 & 52.0 & 41.2 & 50.8 & \textbf{69.1} & \underline{68.8} \\
      & D2T          & \underline{65.0} & 58.7 & 47.2 & 42.4 & 45.9 & 65.0 & \textbf{65.0} \\
      \cmidrule(lr){2-9}
      & \textit{avg} & 61.6 & 54.3 & 49.3 & 41.9 & 50.1 & \underline{66.0} & \textbf{66.2} \\
    \midrule
    \multirow{4}{*}{Qwen2.5-7B}
      & SUM          & \underline{62.3} & 49.8 & 54.5 & 60.1 & 58.2 & 60.9 & \textbf{63.6} \\
      & NQ           & 58.4 & 48.2 & 49.8 & 58.6 & 59.4 & \underline{67.0} & \textbf{68.3} \\
      & D2T          & 62.5 & 51.9 & 52.2 & 54.8 & 58.6 & \underline{63.0} & \textbf{66.9} \\
      \cmidrule(lr){2-9}
      & \textit{avg} & 61.1 & 50.0 & 52.2 & 57.8 & 58.7 & \underline{63.6} & \textbf{66.3} \\
    \midrule
    \multicolumn{2}{@{}l}{\textbf{Mean (18 shards)}}
      & 59.8 & 51.5 & 51.4 & 48.1 & 57.1 & \underline{67.8} & \textbf{69.4} \\
    \bottomrule
  \end{tabular}
  }%
\end{table}

\clearpage
\subsection{Selective Generation}\label{subsec:selective-generation}\label{app:selective-generation-full}

\begin{figure}[t]
  \begin{minipage}[b]{0.46\textwidth}
    \centering
    \providecommand{\wrd}[2]{\begin{tabular}{@{}c@{}}#1\\[-2pt]{\tiny\textcolor{black!55}{#2}}\end{tabular}}
\centering
\resizebox{\linewidth}{!}{%
\begin{tikzpicture}[
  tok/.style={draw, rounded corners=2pt, minimum height=0.66cm,
              inner xsep=3.5pt, inner ysep=2pt, font=\scriptsize},
  served/.style={tok, fill=blue!11, draw=blue!55!black},
  flagged/.style={tok, fill=orange!28, draw=orange!85!black, thick},
  ghost/.style={tok, dashed, draw=gray!65, fill=white, text=gray},
  hal/.style={red!80!black, line width=1.6pt},
  summ/.style={anchor=west, font=\scriptsize},
  qlab/.style={anchor=east, font=\small},
  gap/.style={right=3pt of #1},
]
\node[qlab] (r1l) at (0.7, 0) {$q_1$};
\node[served, right=6pt of r1l] (r1t1) {\wrd{Water}{.9}};
\node[served, gap=r1t1] (r1t2) {\wrd{boils}{.8}};
\node[served, gap=r1t2] (r1t3) {\wrd{at}{.7}};
\node[served, gap=r1t3] (r1t4) {\wrd{90}{.6}};
\node[served, gap=r1t4] (r1t5) {\wrd{degrees}{.8}};
\node[served, gap=r1t5] (r1t6) {\wrd{EOS}{.9}};
\draw[hal] ($(r1t4.south west)+(2pt,-2pt)$) -- ($(r1t4.south east)+(-2pt,-2pt)$);
\node[summ, right=5pt of r1t6] {served in full};

\node[qlab] (r2l) at (0.7, -1.8) {$q_2$};
\node[served, right=6pt of r2l] (r2t1) {\wrd{The}{.8}};
\node[served, gap=r2t1] (r2t2) {\wrd{Moon}{.7}};
\node[served, gap=r2t2] (r2t3) {\wrd{orbits}{.9}};
\node[flagged, gap=r2t3] (r2t4) {\wrd{Mars}{.3}};
\node[ghost, gap=r2t4] (r2t5) {\wrd{every}{}};
\node[ghost, gap=r2t5] (r2t6) {\wrd{month}{}};
\node[ghost, gap=r2t6] (r2t7) {\wrd{EOS}{}};
\draw[hal] ($(r2t4.south west)+(2pt,-2pt)$) -- ($(r2t4.south east)+(-2pt,-2pt)$);
\draw[hal] ($(r2t6.south west)+(2pt,-2pt)$) -- ($(r2t6.south east)+(-2pt,-2pt)$);
\draw[<-, orange!85!black] (r2t4.north) -- ++(0, 0.24)
  node[above, font=\tiny, text=orange!85!black] {$\kappa_f {<} \tau$: halt};
\draw[decorate, decoration={brace, mirror, amplitude=3pt}, blue!55!black]
  ($(r2t1.south west)+(0,-0.30)$) -- ($(r2t3.south east)+(0,-0.30)$)
  node[midway, below=3pt, font=\tiny, text=blue!55!black] {$\hat{\mathbf{x}}_2$ (served)};
\draw[decorate, decoration={brace, mirror, amplitude=3pt}, gray]
  ($(r2t5.south west)+(0,-0.30)$) -- ($(r2t7.south east)+(0,-0.30)$)
  node[midway, below=3pt, font=\tiny, text=gray] {never generated};

\node[anchor=north, font=\small] at (5.4, -3.15) {%
  \begin{tabular}{@{}l@{\;$=$\;}l@{\qquad}l@{\;$=$\;}l@{}}
    $\mathrlap{\hat\phi}\hphantom{\hat{R}}$ & $\tfrac{1}{2} = 50\%$
      & $\hat\phi^{\text{token}}$
      & $\tfrac{1}{2}\bigl(\tfrac{6}{6} + \tfrac{3}{7}\bigr) = \tfrac{5}{7} \approx 71\%$ \\[7pt]
    $\hat{R}$
      & $\dfrac{\tfrac{1}{6} + \tfrac{0}{7}}{\tfrac{6}{6} + \tfrac{3}{7}} = \tfrac{7}{60} \approx 12\%$
      & $\mathrlap{\hat{S}}\hphantom{\hat\phi^{\text{token}}}$
      & $1 - \hat\phi^{\text{token}} = \tfrac{2}{7} \approx 29\%$ \\
  \end{tabular}%
};
\end{tikzpicture}%
}
\captionof{figure}{Selective generation at $\tau = 0.5$. Orange =
flagged and withheld, dashed = never generated,
\textcolor{red!80!black}{red underline} = hallucinated.}
\label{fig:rc-mini}
  \end{minipage}\hfill
  \begin{minipage}[b]{0.50\textwidth}
    \centering
    \includegraphics[width=\linewidth]{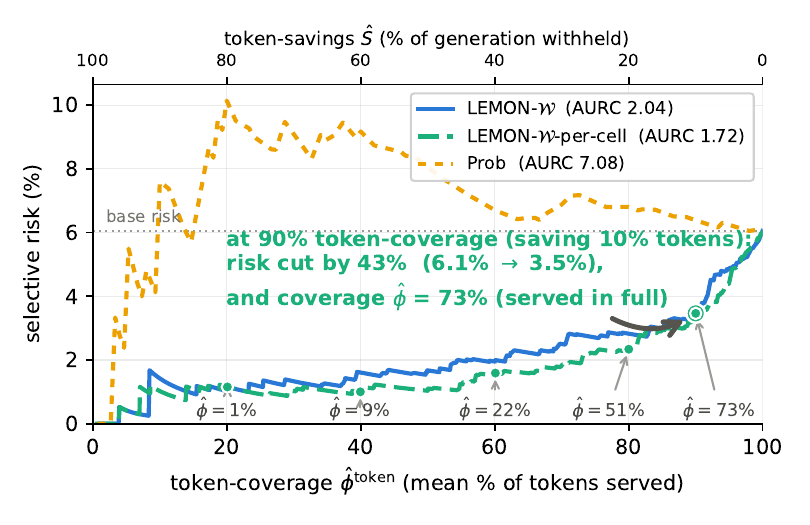}
    \caption{\textbf{RC-curves}. The dotted line is the base
    hallucination rate. AURC is the area under the RC-curve.}
    \label{fig:rc-curve-comparison}
  \end{minipage}
\end{figure}

We demonstrate \lemon on \emph{selective generation}: halting LLM generation when hallucination is likely. \Cref{app:selective-generation} formalizes the framework thoroughly; here we summarize only what is needed to interpret the results. Let $\mathbf{x}$ denote an LLM-generated response (the prompt is treated as part of $\mathbf{x}$ for notational convenience), and $\mathbf{x}_{:t}$ its prefix up to token $t$. A \emph{confidence function} $\kappa_f(\mathbf{x}_{:t})$ estimates how likely the next token is to be hallucinated; generation halts the first time the confidence falls below a specified threshold $\tau$, and the user is served only the prefix preceding the flagged error, denoted $\hat{\mathbf{x}}$.

We evaluate a confidence function using four quantities, summarized in \Cref{eq:selective-generation-compact} and illustrated in \Cref{fig:rc-mini}, whose relative importance may vary across use cases: (i) coverage $\hat{\phi}$, the fraction of generations that are served in full, i.e., never halted; (ii) token-coverage $\hat{\phi}^{\text{token}}$, the average fraction of each generation that is served; (iii) selective risk $\hat{R}$, the fraction of served content that is hallucinated; and (iv) token-savings $\hat{S}$, the fraction of tokens that were withheld. Over $N$ generations with full, unhalted lengths $\{ T_i \}_{i=1}^N$, these quantities are defined as follows:
\begin{equation}
  \label{eq:selective-generation-compact}
\underbrace{\hat{\phi} \coloneqq
\frac{\bigl|\{i \in [N] : \hat{\mathbf{x}}_i = \mathbf{x}_i\}\bigr|}{N}}_{\text{coverage}}
\quad \!\!
\underbrace{\hat{\phi}^{\text{token}} \coloneqq \frac{1}{N}\sum_{i=1}^{N}\frac{|\hat{\mathbf{x}}_i|}{T_i}}_{\text{token-coverage}}
\quad \!\!
\underbrace{\hat{R} \coloneqq \frac{\sum_{i=1}^{N} h_i/T_i}{\sum_{i=1}^{N} |\hat{\mathbf{x}}_i|/T_i}}_{\text{selective risk}} 
\quad \!\!
\underbrace{\hat{S} \coloneqq 1-\hat{\phi}^{\text{token}}}_{\text{token-savings}}
\end{equation}
where $h_i$ is the number of hallucinated units in $\hat{\mathbf{x}}_i$ and $[N] \coloneqq \{1, \ldots, N \}$. Sweeping $\tau$ trades coverage/token-coverage against selective risk and traces the \emph{risk-coverage curve} (\emph{RC-Curve}), and provides an effective measure of a confidence function's quality.

\paragraph{Results.} \Cref{fig:rc-curve-comparison} shows the RC curves of \lemonW, \lemonWpc, and the strongest baseline, \probbase, over the specific combination of the NQ dataset,
for \llamathirteen as left out LLM. We observe that both \lemon variants significantly outperform \probbase, which is not even monotonically decreasing. In contrast, both \lemon variants exhibit consistently monotonic (decreasing) behavior and, for example, at $90\%$ token-coverage, reduce the hallucination rate by more than $43\%$, while maintaining $73\%$ of prompts fully served.

Finally, \Cref{tab:rc-aurc-v2} reports the area under the RC-curve (AURC) for every
(held-out LLM, task) shard, summarizing the whole risk--coverage trade-off in a single
number per shard. \lemon attains the lowest AURC in 17 of the 18 shards, and averaged
over the shards \lemonWpc reaches $8.9$ and \lemonW $9.1$, against $12.3$ for \probbase.

\begin{table}[htbp]
\centering
\caption{\textbf{Selective generation.} A per-shard breakdown of the area under the risk--coverage curve (lower is better). \textbf{Bold} = best in row.}
\label{tab:rc-aurc-v2}
\begin{tabular}{@{}ll cc c@{}}
\toprule
LLM & Task & \lemonW & \lemonWpc & Prob \\
\midrule
\multirow{4}{*}{Llama-2-7B}
  & Summ. & 3.00 & \textbf{2.83} & 7.38 \\
  & NQ & 3.37 & \textbf{2.99} & 9.23 \\
  & D2T & \textbf{5.97} & 6.00 & 9.67 \\
  \cmidrule(l){2-5}
  & \textit{mean} & \textit{4.11} & \textbf{3.94} & \textit{8.76} \\
\midrule
\multirow{4}{*}{Llama-2-13B}
  & Summ. & \textbf{1.62} & 1.71 & 2.97 \\
  & NQ & 2.04 & \textbf{1.72} & 7.08 \\
  & D2T & \textbf{6.65} & 6.77 & 11.25 \\
  \cmidrule(l){2-5}
  & \textit{mean} & \textit{3.44} & \textbf{3.40} & \textit{7.10} \\
\midrule
\multirow{4}{*}{Mistral-7B}
  & Summ. & 6.58 & \textbf{6.57} & 7.56 \\
  & NQ & 1.74 & \textbf{1.41} & 8.20 \\
  & D2T & 8.70 & \textbf{8.50} & 9.32 \\
  \cmidrule(l){2-5}
  & \textit{mean} & \textit{5.67} & \textbf{5.49} & \textit{8.36} \\
\midrule
\multirow{4}{*}{Mistral-Nemo}
  & Summ. & 6.31 & \textbf{6.12} & 6.40 \\
  & NQ & 16.95 & \textbf{16.56} & 24.69 \\
  & D2T & 20.82 & \textbf{20.64} & 24.28 \\
  \cmidrule(l){2-5}
  & \textit{mean} & \textit{14.69} & \textbf{14.44} & \textit{18.46} \\
\midrule
\multirow{4}{*}{Qwen2.5-3B}
  & Summ. & \textbf{7.37} & 7.52 & 9.96 \\
  & NQ & \textbf{12.40} & 12.59 & 18.77 \\
  & D2T & 19.87 & \textbf{19.81} & \textbf{19.81} \\
  \cmidrule(l){2-5}
  & \textit{mean} & \textbf{13.21} & \textit{13.31} & \textit{16.18} \\
\midrule
\multirow{4}{*}{Qwen2.5-7B}
  & Summ. & \textbf{6.96} & 7.17 & 7.48 \\
  & NQ & 13.51 & \textbf{13.21} & 17.28 \\
  & D2T & 19.97 & \textbf{18.33} & 19.38 \\
  \cmidrule(l){2-5}
  & \textit{mean} & \textit{13.48} & \textbf{12.90} & \textit{14.72} \\
\midrule
\textbf{All} & \textbf{mean (18)} & 9.10 & \textbf{8.91} & 12.26 \\
\bottomrule
\end{tabular}
\end{table}

\clearpage
\subsection{Few-shot adaptation}\label{app:few-shot-adaptation}

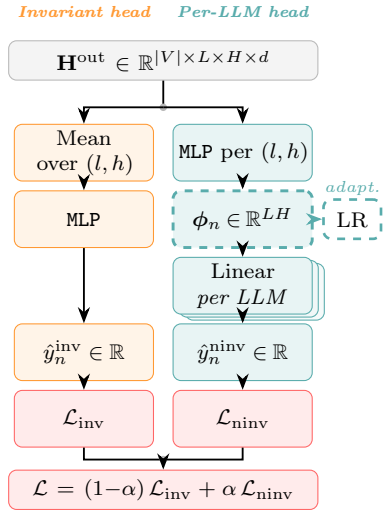
\begin{figure}[h]
  \centering
  \begin{tikzpicture}[
    every node/.style={font=\scriptsize},
    >=Stealth,
    MH/.style={draw=orange!80, fill=orange!9, rounded corners=3pt, align=center,
               minimum height=0.75cm, text width=1.70cm, inner sep=2pt},
    LH/.style={draw=teal!65,   fill=teal!9,  rounded corners=3pt, align=center,
               minimum height=0.75cm, text width=1.70cm, inner sep=2pt},
    IN/.style={draw=gray!55,   fill=gray!9,  rounded corners=3pt, align=center,
               minimum height=0.52cm, text width=3.95cm, inner sep=2pt},
    LS/.style={draw=red!60,    fill=red!8,   rounded corners=3pt, align=center,
               minimum height=0.75cm, text width=1.70cm, inner sep=2pt},
    TL/.style={draw=red!60,    fill=red!8,   rounded corners=3pt, align=center,
               minimum height=0.52cm, text width=3.95cm, inner sep=2pt},
    arr/.style={->, line width=0.7pt},
    LR/.style={draw=teal!65, dashed, line width=1.0pt, fill=white,
               rounded corners=3pt, align=center,
               minimum height=0.52cm, text width=0.62cm, inner sep=2pt},
  ]
  \node[orange!80, font=\tiny\bfseries\itshape] at (-1.05, 0.65) {Invariant head};
  \node[teal!65,   font=\tiny\bfseries\itshape] at ( 1.05, 0.65) {Per-LLM head};
  \node[IN] (inp) at (0, 0)
    {$\mathbf{H}^{\mathrm{out}}\!\in\!\mathbb{R}^{|V|\times L\times H\times d}$};
  \node[MH] (avg)  at (-1.05,-1.22) {Mean over $(l,h)$};
  \node[MH] (mlpm) at (-1.05,-2.10) {\texttt{MLP}};
  \node[MH] (ym)   at (-1.05,-3.86) {$\hat{y}_n^{\mathrm{inv}}\!\in\!\mathbb{R}$};
  \node[LS] (lm)   at (-1.05,-4.74) {$\mathcal{L}_{\mathrm{inv}}$};
  \node[LH] (mlpc) at ( 1.05,-1.22) {\texttt{MLP} per $(l,h)$};
  \node[LH, draw=teal!65, dashed, line width=1.1pt] (phi) at ( 1.05,-2.10)
    {$\boldsymbol{\phi}_n\!\in\!\mathbb{R}^{LH}$};
  \node[LH, fill=teal!4] at ( 1.21,-3.06) {};
  \node[LH, fill=teal!6] at ( 1.13,-3.02) {};
  \node[LH] (lin) at ( 1.05,-2.98) {Linear \emph{per LLM}};
  \node[LH] (yl)   at ( 1.05,-3.86) {$\hat{y}_n^{\mathrm{ninv}}\!\in\!\mathbb{R}$};
  \node[LS] (ll)   at ( 1.05,-4.74) {$\mathcal{L}_{\mathrm{ninv}}$};
  \node[TL] (tot) at (0,-5.68)
    {$\mathcal{L}=(1{-}\alpha)\,\mathcal{L}_{\mathrm{inv}}+\alpha\,\mathcal{L}_{\mathrm{ninv}}$};
  \coordinate (fk) at (0,-0.62);
  \draw[-,  line width=0.7pt] (inp.south) -- (fk);
  \fill[gray!55] (fk) circle (1.4pt);
  \draw[arr] (fk) -| (avg.north);
  \draw[arr] (fk) -| (mlpc.north);
  \draw[arr] (avg.south)  -- (mlpm.north);
  \draw[arr] (mlpm.south) -- (ym.north);
  \draw[arr] (ym.south)   -- (lm.north);
  \draw[arr] (mlpc.south) -- (phi.north);
  \draw[arr] (phi.south)  -- (lin.north);
  \draw[arr] (lin.south)  -- (yl.north);
  \draw[arr] (yl.south)   -- (ll.north);
  \node[LR] (lr) at (2.50,-2.10) {LR};
  \draw[arr, dashed, teal!65] (phi.east) -- (lr.west);
  \node[font=\tiny\itshape, teal!65, above=1pt of lr, inner sep=0pt] {adapt.};
  \coordinate (mL) at (-1.05,-5.28);
  \coordinate (mR) at ( 1.05,-5.28);
  \coordinate (mC) at (0,    -5.28);
  \draw[-, line width=0.7pt] (lm.south) -- (mL) -- (mC);
  \draw[-, line width=0.7pt] (ll.south) -- (mR) -- (mC);
  \fill[gray!55] (mC) circle (1.4pt);
  \draw[arr] (mC) -- (tot.north);
  \end{tikzpicture}
  \caption{\textcolor{orange!80}{\itshape Invariant} and
    \textcolor{teal!65}{\itshape per-LLM} heads (one linear predictor per pool
    LLM). $\mathbf{H}^{\mathrm{out}}$ is the per-node features before the
    prediction heads. Dashed \textcolor{teal!65}{LR}: the adaptation read-out,
    fit on the frozen $\boldsymbol{\phi}_n$ after training, so it does not feed
    $\mathcal{L}$.}
  \label{fig:pred-heads}
\end{figure}

We test whether \lemon can adapt to the held-out LLM given $N_\mathrm{adapt}$ labeled samples from its training shard. Since \lemon's invariant head (see \Cref{fig:pred-heads}, left) discards head and layer identity by design, we retrain \lemon under the same LOO setup with a second, \emph{per-LLM} non-invariant head (\Cref{fig:pred-heads}, right) alongside the original invariant one.

The two heads are trained jointly with a convex combination of their binary cross-entropy losses,
\begin{equation}
\mathcal{L}=(1-\alpha)\,\mathcal{L}_{\mathrm{inv}}+\alpha\,\mathcal{L}_{\mathrm{non-inv}},
\end{equation}
so that $\alpha=0$ recovers the original, invariant-only training. The non-invariant head pushes the per-node features $\boldsymbol{\phi}_n \in \mathbb{R}^{L H}$ to be linearly separable for hallucination detection, so that a lightweight classifier can adapt to a new LLM with minimal supervision. Adaptation itself is simple: we freeze the backbone, extract $\boldsymbol{\phi}_n$ on the held-out LLM, and fit a logistic regression on the $N_\mathrm{adapt}$ labeled samples. Two questions arise: \emph{(1) are the learned features adaptable? (2) does training a second head hurt zero-shot transfer?} As a reference to this experiment, we use the \lb baseline fit on the same $N_\mathrm{adapt}$ samples.

\begin{figure}[h]
  \centering
  \includegraphics[width=0.75\textwidth]{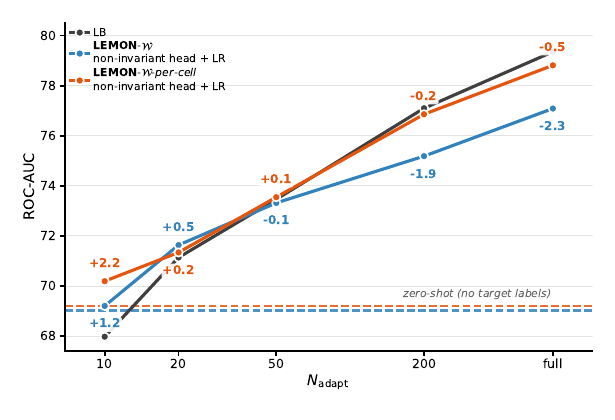}
  \caption{\textbf{Adaptation to the held-out LLM vs.\ labeling budget}, both
    \lemon variants; mean over the 18 (LLM, task) shards. Solid: logistic
    regression on the frozen non-invariant head's features ($\alpha{=}0.5$) and
    \lb; dashed: the corresponding invariant-head (zero-shot) levels. The signed
    number at each point is that variant's gap to \lb.}
  \label{fig:adaptation-curve}
\end{figure}

\begin{figure}[h]
  \centering
  \includegraphics[width=0.95\textwidth]{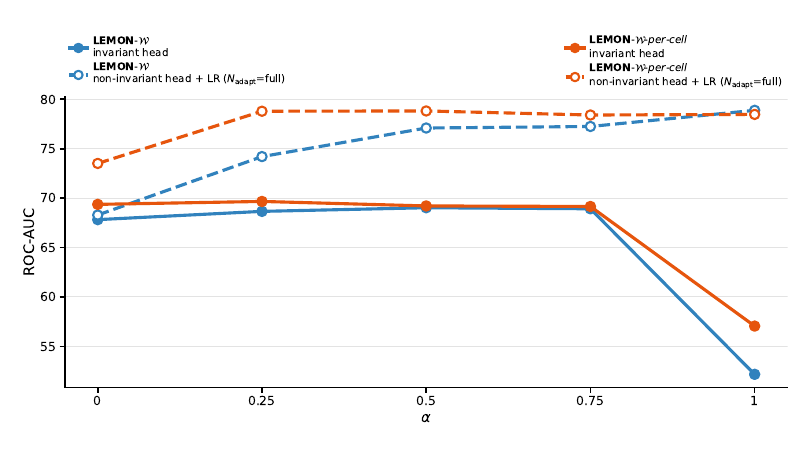}
  \caption{\textbf{Invariant head vs.\ non-invariant head $+$ LR as a function of
    $\alpha$}, both \lemon variants. We report the mean over the 18 (LLM, task) shards. Raising
    $\alpha$ trains the per-LLM non-invariant pathway and improves adaptation,
    while at $\alpha{=}1$ the invariant head collapses towards chance as expected.}
  \label{fig:alpha-tradeoff}
\end{figure}

\paragraph{Results.} \emph{(1) The features are adaptable.} With the full adaptation set ($N_\mathrm{adapt}{=}\text{full}$), logistic-regression adaptation on \lemon features improves as $\alpha$ increases and plateaus at around $\alpha{=}0.5$ (\Cref{fig:alpha-tradeoff}). We therefore fix $\alpha{=}0.5$ to evaluate adaptation in low-sample regimes. In this setting, \lemonWpc outperforms \lb at $N_\mathrm{adapt}{=}10$ ($+2.2$ ROC-AUC) and performs comparably at all larger budgets (\Cref{fig:adaptation-curve}).
\emph{(2) The second head does not hurt zero-shot transfer.} Zero-shot performance of the invariant head remains stable for $\alpha \le 0.75$ and degrades only at $\alpha{=}1$, where the head receives no gradient.

\subsection{LOI Results} \label{app:LOI-Results}
In the leave-one-in (LOI) setup, each detector is trained on the three task shards of a \emph{single} LLM and evaluated zero-shot on the remaining five LLMs---the minimal-source-diversity counterpart of the LOO setup. \Cref{tab:loi-zero-shot-full} reports the full per-task breakdown.
\begin{table}[htbp]
  \centering
  \caption{%
    \textbf{LOI zero-shot generalization --- per-task breakdown.}
    Each monitor is trained on a single (left-in) LLM and evaluated zero-shot
    on the other five. Each data row is one (left-in LLM, task) pair;
    every value is the mean ROC-AUC\,${\times}\,100$ over the five held-out
    evaluation LLMs; \textit{avg} rows average the three tasks for that LLM.
    \textbf{Bold} = best per row; \underline{underline} = second best.
  }
  \label{tab:loi-zero-shot-full}
  \scriptsize
  \resizebox{\textwidth}{!}{%
  \begin{tabular}{@{}ll ccccc | cc@{}}
    \toprule
    \multirow{2}{*}{\textbf{Left-in LLM}} & \multirow{2}{*}{\textbf{Task}}
      & \multirow{2}{*}{\probbase}
      & \multirow{2}{*}{\lb}
      & \multirow{2}{*}{\hs}
      & \multirow{2}{*}{\charm}
      & \multirow{2}{*}{\lbalign}
      & \multicolumn{2}{c}{\textbf{\lemon (Ours)}} \\
    \cmidrule(lr){8-9}
    & & & & & & & $\mathcal{W}$ & $\mathcal{W}$-per-cell \\
    \midrule
    \multirow{4}{*}{Llama-2-7B}
      & SUM & \textbf{62.3} & 53.4 & 48.3 & 60.1 & 55.8 & 55.7 & \underline{62.1} \\
      & NQ & 57.5 & 56.1 & 52.1 & 59.2 & 59.9 & \underline{62.7} & \textbf{66.1} \\
      & D2T & \underline{60.7} & 55.7 & 50.0 & 54.4 & 49.2 & 54.5 & \textbf{64.7} \\
      \cmidrule(lr){2-9}
      & \textit{avg} & \underline{60.2} & 55.1 & 50.1 & 57.9 & 55.0 & 57.6 & \textbf{64.3} \\
    \midrule
    \multirow{4}{*}{Llama-2-13B}
      & SUM & \underline{62.4} & 49.7 & 53.4 & 50.6 & 49.5 & 61.8 & \textbf{63.5} \\
      & NQ & 60.0 & 49.6 & 52.7 & 50.0 & 54.3 & \underline{65.3} & \textbf{71.7} \\
      & D2T & \underline{61.5} & 50.6 & 53.3 & 52.2 & 44.0 & \textbf{63.6} & 61.3 \\
      \cmidrule(lr){2-9}
      & \textit{avg} & 61.3 & 49.9 & 53.1 & 50.9 & 49.3 & \underline{63.6} & \textbf{65.5} \\
    \midrule
    \multirow{4}{*}{Mistral-7B}
      & SUM & 59.3 & 57.4 & 50.8 & 49.6 & 58.9 & \underline{60.6} & \textbf{64.7} \\
      & NQ & 56.3 & 56.4 & 49.2 & 64.3 & 62.6 & \underline{72.0} & \textbf{73.3} \\
      & D2T & 59.0 & 54.0 & 52.0 & 50.7 & 55.8 & \underline{60.9} & \textbf{63.9} \\
      \cmidrule(lr){2-9}
      & \textit{avg} & 58.2 & 55.9 & 50.7 & 54.9 & 59.1 & \underline{64.5} & \textbf{67.3} \\
    \midrule
    \multirow{4}{*}{Mistral-Nemo}
      & SUM & 61.3 & 48.3 & 49.7 & 49.8 & 53.6 & \textbf{65.8} & \underline{65.5} \\
      & NQ & 58.6 & 50.0 & 48.7 & 52.8 & 51.9 & \textbf{76.6} & \underline{75.4} \\
      & D2T & 60.3 & 57.7 & 54.3 & 45.4 & 51.9 & \underline{60.5} & \textbf{60.6} \\
      \cmidrule(lr){2-9}
      & \textit{avg} & 60.1 & 52.0 & 50.9 & 49.3 & 52.5 & \textbf{67.7} & \underline{67.2} \\
    \midrule
    \multirow{4}{*}{Qwen2.5-3B}
      & SUM & 61.3 & 53.2 & 50.0 & 43.8 & 51.1 & \underline{67.2} & \textbf{67.8} \\
      & NQ & 57.9 & 53.5 & 49.4 & 42.9 & 53.4 & \underline{74.3} & \textbf{74.5} \\
      & D2T & 59.0 & 50.7 & 56.9 & 48.9 & 51.6 & \underline{64.2} & \textbf{64.6} \\
      \cmidrule(lr){2-9}
      & \textit{avg} & 59.4 & 52.5 & 52.1 & 45.2 & 52.0 & \underline{68.6} & \textbf{69.0} \\
    \midrule
    \multirow{4}{*}{Qwen2.5-7B}
      & SUM & 61.1 & 44.4 & 51.8 & 48.2 & 52.5 & \textbf{68.6} & \underline{67.9} \\
      & NQ & 58.0 & 46.2 & 51.6 & 32.6 & 51.1 & \underline{75.0} & \textbf{76.1} \\
      & D2T & 59.5 & 39.1 & 55.4 & 52.6 & 48.8 & \textbf{62.1} & \underline{61.3} \\
      \cmidrule(lr){2-9}
      & \textit{avg} & 59.5 & 43.2 & 52.9 & 44.4 & 50.8 & \textbf{68.6} & \underline{68.4} \\
    \midrule
    \multicolumn{2}{@{}l}{\textbf{Mean (18 shards)}}
 & 59.8 & 51.4 & 51.6 & 50.4 & 53.1 & \underline{65.1} & \textbf{66.9} \\
    \bottomrule
  \end{tabular}
  }%
\end{table}

\clearpage
\section{Selective Generation: Formal Framework}
\label{app:selective-generation}

This appendix formalizes the \emph{selective-generation} setting introduced in \Cref{subsec:selective-generation}. It builds on selective prediction \citep{geifman2017selective}, in which a model is trained to know what it does not know, allowing it to abstain when insufficiently confident.  The rejection option dates back over sixty years~\citep{5222035} and has since been studied for neural networks~\citep{geifman2017selective,geifman2019selectivenet}. For LLMs, abstention has mostly been studied at the \emph{response} level: refusing
or deferring a whole answer based on uncertainty estimates, self-evaluation, or
conformal guarantees~\citep{cherian2024large,kuhn2023semanticuncertaintylinguisticinvariances,kadavath2022language}. In contrast, we formalize \emph{selective generation} for LLMs at the token level, using an online monitor during decoding, such as \lemon, the monitor introduced in this paper, to halt generation mid-response.

\paragraph{Setup.} We consider standard autoregressive LLMs. Writing $\mathbf{x}_{\le t} \coloneqq (x_1,\ldots,x_t)$ for a token sequence, they decompose its next token probability using the chain rule:
\[
p(\mathbf{x}_{\le t})=\prod_{i=1}^{t} p(x_i \mid \mathbf{x}_{\le i-1}),
\]
where $\mathbf{x}_{\le i-1}$ is the prefix preceding token $x_i$ (with $\mathbf{x}_{\le 0}$ the empty sequence). At inference time, an LLM generates tokens sequentially by sampling each new token from its predicted conditional categorical distribution $p(x_{t+1} \mid \mathbf{x}_{\le t})$ over the vocabulary $V$, until an end-of-sequence token (\texttt{EOS}) is produced. Thus, for each prefix $\mathbf{x}_{\le t}$, the LLM defines a mapping,
\[
p(x_{t+1} \mid \mathbf{x}_{\le t}) : V^{t} \to \Delta(V),
\]
where $\Delta(V)$ denotes the probability simplex over $V$.

\paragraph{Selective generation.} We define a \emph{selective generator} as a pair of functions $(f,g)$ acting on prefixes of arbitrary length, $V^{*} \coloneqq \bigcup_{t \ge 0} V^{t}$. Here, $f : V^{*} \to \Delta(V)$ denotes the LLM itself, with $f(\mathbf{x}_{\le t}) \coloneqq p(x_{t+1} \mid \mathbf{x}_{\le t})$, while $g : V^{*} \to \{0,1\}$ is a \emph{selection function}.

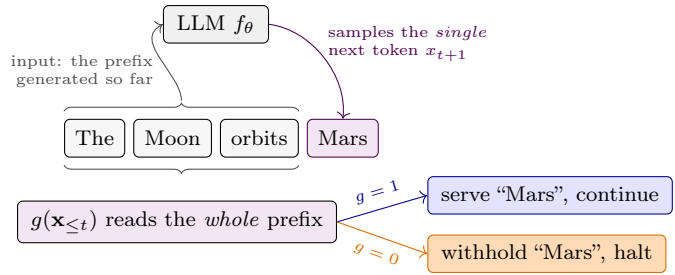
\begin{figure}[H]
\centering
\begin{tikzpicture}[font=\scriptsize,
  tok/.style={draw, rounded corners=2pt, minimum height=0.5cm, inner xsep=4pt,
              inner ysep=2pt, font=\scriptsize,
              text height=1.5ex, text depth=0.35ex, fill=gray!6},
  new/.style={tok, fill=violet!10, draw=violet!60!black},
  box/.style={draw, rounded corners=2pt, inner xsep=5pt, inner ysep=3.5pt},
  lab/.style={font=\tiny},
]
\node[tok]                  (t1) {The};
\node[tok, right=3pt of t1] (t2) {Moon};
\node[tok, right=3pt of t2] (t3) {orbits};
\node[new, right=3pt of t3] (t4) {Mars};
\node[box, fill=gray!12] (llm) at ($(t2.north)!0.5!(t3.north)+(0,1.25)$) {LLM $f_\theta$};
\draw[decorate, decoration={brace, amplitude=3pt}, black!70]
  ($(t1.north west)+(0,0.10)$) -- ($(t3.north east)+(0,0.10)$) coordinate[midway] (inm);
\draw[->, black!70] ($(inm)+(0,0.14)$) .. controls +(0,10pt) and +(-12pt,-6pt) .. (llm.west)
  node[pos=0.35, left=2pt, lab, text=black!70, align=right] {input: the prefix\\generated so far};
\draw[->, violet!60!black] (llm.east) .. controls +(16pt,-2pt) and +(2pt,16pt) .. (t4.north)
  node[pos=0.35, right=3pt, lab, text=violet!60!black, align=left] {samples the \emph{single}\\next token $x_{t+1}$};
\draw[decorate, decoration={brace, mirror, amplitude=3pt}, black!70]
  ($(t1.south west)+(0,-0.10)$) -- ($(t3.south east)+(0,-0.10)$);
\node[box, fill=violet!10] (g) at ($(t1.south)!0.5!(t3.south)+(0,-0.85)$)
  {$g(\mathbf{x}_{\le t})$ reads the \emph{whole} prefix};
\node[box, fill=blue!11,   draw=blue!55!black,   right=34pt of g, yshift=10pt]  (serve)
  {serve ``Mars'', continue};
\node[box, fill=orange!28, draw=orange!85!black, right=34pt of g, yshift=-12pt] (halt)
  {withhold ``Mars'', halt};
\draw[->, blue!55!black]   (g.east) -- node[above, sloped, lab] {$g = 1$} (serve.west);
\draw[->, orange!85!black] (g.east) -- node[below, sloped, lab] {$g = 0$} (halt.west);
\end{tikzpicture}
\caption{The selective generator $(f,g)$: the LLM $f$ samples one token at a time, while the selection function $g$ decides whether to serve the newest token and continue generation, or to withhold it and halt.}
\label{fig:selgen-pair}
\end{figure}

Whenever the LLM samples a new token $x_{t+1}$, $g$ inspects the preceding prefix $\mathbf{x}_{\le t} $ and decides whether to serve the new token or to withhold it and stop generating:
\[
(f,g)(\mathbf{x}_{\le t}) \coloneqq
\begin{cases}
  \text{serve } x_{t+1} \text{ and continue} & \text{if } g(\mathbf{x}_{\le t}) = 1,\\
  \text{withhold } x_{t+1} \text{ and halt}  & \text{if } g(\mathbf{x}_{\le t}) = 0,
\end{cases}
\]
see \Cref{fig:selgen-pair} for an illustration of the selective-generation process.

Thus, generation halts at the \emph{first} time step at which $g$ abstains. Writing $\mathbf{x} = (x_1, \ldots, x_T)$ for the full, \emph{uninterrupted} generation, terminating with the \texttt{EOS} token at step $T$, the \emph{halting step} is defined as:
\begin{equation}
\rho \coloneqq \min\bigl\{\, t \ge 0 : g(\mathbf{x}_{\le t}) = 0 \,\bigr\}  + 1
\qquad \bigl(\rho = \infty \text{ if } g \text{ never abstains}\bigr).
\label{eq:selgen-x-stop}
\end{equation}

\begin{figure}[H]
\centering
\begin{tikzpicture}[
  tok/.style={draw, rounded corners=2pt, minimum height=0.5cm, inner xsep=4pt,
              inner ysep=2pt, font=\scriptsize,
              text height=1.5ex, text depth=0.35ex},
  served/.style={tok, fill=blue!11, draw=blue!55!black},
  flagged/.style={tok, fill=orange!28, draw=orange!85!black, thick},
  ghost/.style={tok, dashed, draw=gray!65, fill=white, text=gray},
]
\node[served]                     (t1) {The};
\node[served,  right=3pt of t1]   (t2) {Moon};
\node[served,  right=3pt of t2]   (t3) {orbits};
\node[flagged, right=3pt of t3]   (t4) {Mars};
\node[ghost,   right=3pt of t4]   (t5) {every};
\node[ghost,   right=3pt of t5]   (t6) {month};
\node[ghost,   right=3pt of t6]   (t7) {\texttt{EOS}};
\draw[<-, orange!85!black] (t4.north) -- ++(0, 0.22)
  node[above, font=\tiny, text=orange!85!black]
  {$\rho = 4$: first rejection; $x_\rho$ is flagged and withheld};
\draw[decorate, decoration={brace, mirror, amplitude=3pt}, blue!55!black]
  ($(t1.south west)+(0,-0.12)$) -- ($(t3.south east)+(0,-0.12)$)
  node[midway, below=3pt, font=\tiny, text=blue!55!black]
  {$\hat{\mathbf{x}} = \mathbf{x}_{\le \rho-1}$ (delivered)};
\draw[decorate, decoration={brace, mirror, amplitude=3pt}, gray]
  ($(t5.south west)+(0,-0.12)$) -- ($(t7.south east)+(0,-0.12)$)
  node[midway, below=3pt, font=\tiny, text=gray] {never generated};
\end{tikzpicture}
\caption{The halting step $\rho$ of \eqref{eq:selgen-x-stop} and the served output
$\hat{\mathbf{x}}$ of \eqref{eq:selgen-x-served}: $g$ abstains for the first time at
``Mars'', so $\rho = 4$. The user sees only the tokens delivered before the flag (blue);
the flagged token itself is withheld (orange), and the suffix beyond it is never
generated (dashed).}
\label{fig:selgen-served}
\end{figure}
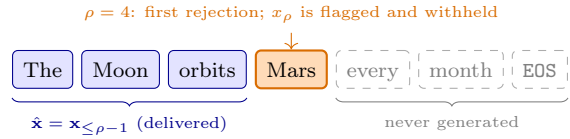

Halting at $\rho$ splits the generation into three parts (see \Cref{fig:selgen-served}): 

\begin{itemize}
  \item The tokens $x_1, \ldots, x_{\rho-1}$, each of which passed $g$'s check and was delivered to the user.
  \item The flagged token $x_\rho$.
  \item The suffix beyond $x_\rho$, which is never generated at all.
\end{itemize}
We define the \emph{served output} $\hat{\mathbf{x}}$ as the prefix of $\mathbf{x}$ that is actually delivered to the user:

\begin{equation}
\hat{\mathbf{x}} \coloneqq
\begin{cases}
  \mathbf{x}_{\le T} & \text{if } \rho = \infty \quad (\text{reached \texttt{EOS}: the full response is delivered}),\\[2pt]
  \mathbf{x}_{\le \rho - 1} & \text{otherwise} \quad (\text{truncated just before the flagged token}).
\end{cases}
\label{eq:selgen-x-served}
\end{equation}

\begin{figure}[H]
\centering
\begin{tikzpicture}[
  tok/.style={draw, rounded corners=2pt, minimum height=0.48cm, minimum width=0.8cm,
              inner xsep=4pt, inner ysep=2pt, font=\scriptsize,
              fill=blue!11, draw=blue!55!black,
              text height=1.5ex, text depth=0.35ex},
  dec/.style={font=\scriptsize},
  gate/.style={-{Latex[length=1.3mm]}, gray!75},
]
\node[tok]                    (t1) {$x_1$};
\node[tok, right=16pt of t1]  (t2) {$x_2$};
\node[tok, right=16pt of t2]  (t3) {$x_3$};
\node[right=12pt of t3, font=\scriptsize] (dots) {$\cdots$};
\node[tok, right=12pt of dots] (t4) {$x_{t+1}$};
\node[dec] (d1) at ($(t1.north)+(0,0.55)$) {$g(\mathbf{x}_{\le 0})$};
\node[dec] (d2) at ($(t2.north)+(0,0.55)$) {$g(\mathbf{x}_{\le 1})$};
\node[dec] (d3) at ($(t3.north)+(0,0.55)$) {$g(\mathbf{x}_{\le 2})$};
\node[dec] (d4) at ($(t4.north)+(0,0.55)$) {$g(\mathbf{x}_{\le t})$};
\foreach \i in {1,...,4} \draw[gate] (d\i.south) -- (t\i.north);
\draw[decorate, decoration={brace, mirror, amplitude=4pt}, black!70]
  ($(t1.south west)+(0,-0.12)$) -- ($(t4.south east)+(0,-0.12)$)
  node[midway, below=5pt, font=\scriptsize]
  {$\tilde g(\mathbf{x}_{\le t+1})
    = g(\mathbf{x}_{\le 0})\cdot g(\mathbf{x}_{\le 1})\cdot g(\mathbf{x}_{\le 2})
      \cdots g(\mathbf{x}_{\le t})$};
\end{tikzpicture}
\caption{The per-token acceptance $\tilde g$ of \eqref{eq:selgen-x-tokacc}: each token
$x_{s+1}$ is gated by the decision $g(\mathbf{x}_{\le s})$, made on the strict prefix
preceding it (arrows), and $\tilde g(\mathbf{x}_{\le t+1})$ is the product of all the
decisions up to the one gating $x_{t+1}$---a single rejection anywhere zeroes it.}
\label{fig:selgen-tokacc}
\end{figure}
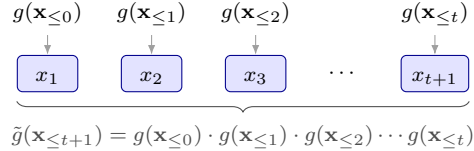

Whether an individual token is served is thus determined jointly by all the decisions up to it. Define the \emph{per-token acceptance} $\tilde g : V^{*} \to \{0, 1\}$,
\begin{equation}
\tilde g(\mathbf{x}_{\le t+1}) \coloneqq \prod_{s=0}^{t} g(\mathbf{x}_{\le s}),
\label{eq:selgen-x-tokacc}
\end{equation}
which equals $1$ iff no prefix preceding $x_{t+1}$ was rejected, i.e., iff $x_{t+1}$ belongs to $\hat{\mathbf{x}}$. We note that $\tilde g$ is $0$ from the first rejected token onward, and that $\tilde g(\mathbf{x}_{\le T}) = 1$ iff the full generation is served; \Cref{fig:selgen-tokacc} visualizes how $\tilde g$ aggregates the individual decisions of $g$.

Using $\tilde g( \cdot )$, we also define the \emph{cumulative selection function} $\bar g : V^{*} \to [0, 1]$ as the fraction of served tokens,
\begin{equation}
\bar g(\mathbf{x}) \coloneqq \frac{|\hat{\mathbf{x}}|}{T}
= \frac{1}{T} \sum_{t=1}^{T} \tilde g(\mathbf{x}_{\le t}),
\label{eq:selgen-x-accept}
\end{equation}
so that $\bar g(\mathbf{x}) = 1$ iff generation reached \texttt{EOS} with no abstention ($\rho = \infty$), while a truncated generation earns the partial credit $(\rho - 1)/T$; see \Cref{fig:selgen-gbar} for an illustration of the three quantities $g$, $\tilde g$, and $\bar g$, over an example.
\begin{figure}[H]
\centering
\begin{tikzpicture}[
  tok/.style={draw, rounded corners=2pt, minimum height=0.48cm, inner xsep=4pt,
              inner ysep=2pt, font=\scriptsize,
              text height=1.5ex, text depth=0.35ex},
  served/.style={tok, fill=blue!11, draw=blue!55!black},
  flagged/.style={tok, fill=orange!28, draw=orange!85!black, thick},
  ghost/.style={tok, dashed, draw=gray!65, fill=white, text=gray},
  qlab/.style={anchor=east, font=\scriptsize},
  rlab/.style={anchor=east, font=\tiny, text=black!60},
  note/.style={anchor=west, font=\scriptsize},
]
\node[qlab] (l1) at (0, 0) {$\mathbf{x}$:};
\node[served, right=5pt of l1]  (a1) {Tokyo};
\node[served, right=3pt of a1]  (a2) {is};
\node[served, right=3pt of a2]  (a3) {Japan's};
\node[served, right=3pt of a3]  (a4) {capital};
\node[served, right=3pt of a4]  (a5) {\texttt{EOS}};
\node[rlab] at ($(l1.east)+(0,0.44)$) {$t$:};
\foreach \i in {1,...,5} \node[font=\tiny, text=black!55] at ($(a\i.center)+(0,0.44)$) {\i};
\node[rlab] at ($(l1.east)+(0,-0.50)$) {$g(\mathbf{x}_{\le t-1})$:};
\node[rlab] at ($(l1.east)+(0,-0.92)$) {$\tilde g(\mathbf{x}_{\le t})$:};
\foreach \i in {1,...,5} \node[font=\tiny] at ($(a\i.center)+(0,-0.50)$) {1};
\foreach \i in {1,...,5} \node[font=\tiny] at ($(a\i.center)+(0,-0.92)$) {1};
\node[note] at (8.0, -0.45) {$\Rightarrow\ \bar g(\mathbf{x}) = 1$ (served in full)};
\node[qlab] (l2) at (0, -1.95) {$\mathbf{x}'$:};
\node[served,  right=5pt of l2] (b1) {The};
\node[served,  right=3pt of b1] (b2) {Moon};
\node[served,  right=3pt of b2] (b3) {orbits};
\node[flagged, right=3pt of b3] (b4) {Mars};
\node[ghost,   right=3pt of b4] (b5) {every};
\node[ghost,   right=3pt of b5] (b6) {month};
\node[ghost,   right=3pt of b6] (b7) {\texttt{EOS}};
\node[rlab] at ($(l2.east)+(0,0.44)$) {$t$:};
\foreach \i in {1,...,7} \node[font=\tiny, text=black!55] at ($(b\i.center)+(0,0.44)$) {\i};
\node[rlab] at ($(l2.east)+(0,-0.50)$) {$g(\mathbf{x}_{\le t-1})$:};
\node[rlab] at ($(l2.east)+(0,-0.92)$) {$\tilde g(\mathbf{x}_{\le t})$:};
\foreach \i in {1,2,3} \node[font=\tiny] at ($(b\i.center)+(0,-0.50)$) {1};
\node[font=\bfseries\tiny, text=orange!85!black] at ($(b4.center)+(0,-0.50)$) {0};
\foreach \i in {5,6,7} \node[font=\tiny, text=gray] at ($(b\i.center)+(0,-0.50)$) {$\cdot$};
\foreach \i in {1,2,3} \node[font=\tiny] at ($(b\i.center)+(0,-0.92)$) {1};
\foreach \i in {4,...,7} \node[font=\tiny, text=orange!85!black] at ($(b\i.center)+(0,-0.92)$) {0};
\node[note] at (8.0, -2.40) {$\Rightarrow\ \bar g(\mathbf{x}') = \tfrac{3}{7}$ ($3$ of $7$ tokens served)};
\end{tikzpicture}
\caption{The three selection functions on two \emph{different} generations (one per row);
the time step $t$ of each token is shown above it.
}
\label{fig:selgen-gbar}
\end{figure}

\paragraph{Coverage and selective risk.}
The performance of a selective generator is quantified by two quantities: \emph{token-coverage} and \emph{selective risk}. Let $P$ denote the distribution over \emph{full, uninterrupted} generations $\mathbf{x} = (x_1, \ldots, x_T)$ induced by drawing a prompt and sampling the response from $f$ until \texttt{EOS}. \emph{token-coverage} is the expected fraction of generated tokens that are served,
\begin{equation}
\phi^{\text{token}}(f, g) \coloneqq \mathbb{E}_{\mathbf{x} \sim P}\bigl[\bar g(\mathbf{x})\bigr].
\label{eq:selgen-x-coverage-pop}
\end{equation}
To quantify risk, we assume a loss function $\ell : V^{*} \to \mathbb{R}_{\ge 0}$ that scores the generation of the last token in a prefix, $\ell(\mathbf{x}_{\le t})$, which can be any measure of the quality---for instance, its perplexity under a reference model, or a binary indicator of hallucination. The \emph{selective risk} weights each token's loss by its acceptance:
\begin{equation}
R(f, g) \coloneqq
\frac{\mathbb{E}_{\mathbf{x} \sim P}\Bigl[\,\frac{1}{T} \sum_{t=1}^{T} \ell(\mathbf{x}_{\le t})\, \tilde g(\mathbf{x}_{\le t})\Bigr]}
     {\phi^{\text{token}}(f, g)}.
\label{eq:selgen-x-risk-pop}
\end{equation}
Importantly, only \emph{served} tokens contribute to the selective risk. We now turn to define their empirical counterparts, which are used in practice to evaluate selective generators on finite datasets.

\paragraph{Empirical estimators.}
Given an evaluation set of $N$ prompts with full generations $\mathbf{x}^{(i)} = (x^{(i)}_1, \ldots, x^{(i)}_{T_i})$ and served outputs $\hat{\mathbf{x}}^{(i)}$ as above, both quantities are estimated by replacing expectations with sample means. Thus, the \emph{empirical token-coverage} is the average served fraction,
\begin{equation}
\hat\phi^{\text{token}}(f, g) \coloneqq \frac{1}{N} \sum_{i=1}^{N} \bar g\bigl(\mathbf{x}^{(i)}\bigr)
= \frac{1}{N} \sum_{i=1}^{N} \frac{|\hat{\mathbf{x}}^{(i)}|}{T_i},
\label{eq:selgen-x-coverage}
\end{equation}
and the \emph{empirical selective risk} is,
\begin{equation}
\hat R(f, g) \coloneqq 
\frac{\frac{1}{N}\displaystyle\sum_{i=1}^{N} \frac{1}{T_i} \sum_{t=1}^{T_i} \ell\bigl(\mathbf{x}^{(i)}_{\le t}\bigr)\, \tilde g\bigl(\mathbf{x}^{(i)}_{\le t}\bigr)}
     {\frac{1}{N}\displaystyle\sum_{i=1}^{N} \bar g\bigl(\mathbf{x}^{(i)}\bigr)}.
\label{eq:selgen-x-risk}
\end{equation}
token-coverage credits partially served generations, so it does not reveal how many responses the user receives \emph{untouched}. We therefore also define the \emph{coverage}: the fraction of generations that are served in full, i.e., that reach \texttt{EOS} without $g$ ever abstaining ($\rho = \infty$, equivalently $\hat{\mathbf{x}} = \mathbf{x}$):
\begin{equation}
\hat\phi(f, g) \coloneqq \frac{\bigl|\bigl\{\, i \;:\; \hat{\mathbf{x}}^{(i)} = \mathbf{x}^{(i)} \,\bigr\}\bigr|}{N},
\label{eq:selgen-x-fullrate}
\end{equation}
so that $N \hat\phi$ counts the responses fully satisfied under $g$. Since a fully served generation contributes $\bar g = 1$ to the token-coverage while a truncated one contributes only its served fraction, we always have $\hat\phi \le \hat\phi^{\text{token}}$.
\paragraph{Compute savings.}
Beyond withholding errors, halting also \emph{saves compute}: the suffix beyond the flagged token is never generated. We define the \emph{token-savings} simply as the complement of token-coverage,
\begin{equation}
\hat S(f, g) \coloneqq 1 - \hat\phi^{\text{token}}(f, g),
\label{eq:selgen-x-savings}
\end{equation}
the average fraction of each generation that is withheld. An illustration of the empirical coverage, token-coverage, selective risk, and token-savings is given in \Cref{fig:selgen-covrisk}.
\begin{figure}[H]
\centering
\begin{tikzpicture}[
  tok/.style={draw, rounded corners=2pt, minimum height=0.48cm, inner xsep=4pt,
              inner ysep=2pt, font=\scriptsize,
              text height=1.5ex, text depth=0.35ex},
  served/.style={tok, fill=blue!11, draw=blue!55!black},
  flagged/.style={tok, fill=orange!28, draw=orange!85!black, thick},
  ghost/.style={tok, dashed, draw=gray!65, fill=white, text=gray},
  hal/.style={red!80!black, line width=1.4pt},
  qlab/.style={anchor=east, font=\scriptsize},
  rlab/.style={anchor=east, font=\tiny, text=black!60},
  note/.style={anchor=west, font=\scriptsize, align=left},
]
\node[qlab] (l1) at (0, 0) {$\mathbf{x}$:};
\node[served, right=5pt of l1] (a1) {Water};
\node[served, right=3pt of a1] (a2) {boils};
\node[served, right=3pt of a2] (a3) {at};
\node[served, right=3pt of a3] (a4) {90};
\node[served, right=3pt of a4] (a5) {degrees};
\node[served, right=3pt of a5] (a6) {\texttt{EOS}};
\draw[hal] ($(a4.south west)+(2pt,-1.5pt)$) -- ($(a4.south east)+(-2pt,-1.5pt)$);
\node[rlab] at ($(l1.east)+(0,-0.52)$) {$g$:};
\node[rlab] at ($(l1.east)+(0,-0.94)$) {$\tilde g$:};
\node[rlab] at ($(l1.east)+(0,-1.36)$) {$\ell$:};
\foreach \i in {1,...,6} \node[font=\tiny] at ($(a\i.center)+(0,-0.52)$) {1};
\foreach \i in {1,...,6} \node[font=\tiny] at ($(a\i.center)+(0,-0.94)$) {1};
\foreach \i in {1,2,3,5,6} \node[font=\tiny] at ($(a\i.center)+(0,-1.36)$) {0};
\node[font=\bfseries\tiny, text=red!80!black] at ($(a4.center)+(0,-1.36)$) {1};
\node[note] at (8.2, -0.70)
  {$\bar g(\mathbf{x}) = 1$\\[1pt] $\tfrac{1}{T}\sum_t \ell\,\tilde g = \tfrac{1}{6}$\\[1pt]
   $1 - \bar g(\mathbf{x}) = 0$};
\node[qlab] (l2) at (0, -2.35) {$\mathbf{x}'$:};
\node[served,  right=5pt of l2] (b1) {The};
\node[served,  right=3pt of b1] (b2) {Moon};
\node[served,  right=3pt of b2] (b3) {orbits};
\node[flagged, right=3pt of b3] (b4) {Mars};
\node[ghost,   right=3pt of b4] (b5) {every};
\node[ghost,   right=3pt of b5] (b6) {month};
\node[ghost,   right=3pt of b6] (b7) {\texttt{EOS}};
\draw[hal] ($(b4.south west)+(2pt,-1.5pt)$) -- ($(b4.south east)+(-2pt,-1.5pt)$);
\draw[hal] ($(b6.south west)+(2pt,-1.5pt)$) -- ($(b6.south east)+(-2pt,-1.5pt)$);
\node[rlab] at ($(l2.east)+(0,-0.52)$) {$g$:};
\node[rlab] at ($(l2.east)+(0,-0.94)$) {$\tilde g$:};
\node[rlab] at ($(l2.east)+(0,-1.36)$) {$\ell$:};
\foreach \i in {1,2,3} \node[font=\tiny] at ($(b\i.center)+(0,-0.52)$) {1};
\node[font=\bfseries\tiny, text=orange!85!black] at ($(b4.center)+(0,-0.52)$) {0};
\foreach \i in {5,6,7} \node[font=\tiny, text=gray] at ($(b\i.center)+(0,-0.52)$) {$\cdot$};
\foreach \i in {1,2,3} \node[font=\tiny] at ($(b\i.center)+(0,-0.94)$) {1};
\foreach \i in {4,...,7} \node[font=\tiny, text=orange!85!black] at ($(b\i.center)+(0,-0.94)$) {0};
\foreach \i in {1,2,3} \node[font=\tiny] at ($(b\i.center)+(0,-1.36)$) {0};
\node[font=\tiny, text=gray] at ($(b4.center)+(0,-1.36)$) {1};
\node[font=\tiny, text=gray] at ($(b5.center)+(0,-1.36)$) {0};
\node[font=\tiny, text=gray] at ($(b6.center)+(0,-1.36)$) {1};
\node[font=\tiny, text=gray] at ($(b7.center)+(0,-1.36)$) {0};
\node[note] at (8.2, -3.05)
  {$\bar g(\mathbf{x}') = \tfrac{3}{7}$\\[1pt] $\tfrac{1}{T}\sum_t \ell\,\tilde g = 0$\\[1pt]
   $1 - \bar g(\mathbf{x}') = \tfrac{4}{7}$};
\node[draw=gray!60, rounded corners=2pt, fill=gray!4, inner xsep=3pt, inner ysep=4pt,
      font=\small, anchor=north] at (5.1, -4.10)
  {$\hat\phi = \dfrac{1}{2} = 0.5
    \quad
    \hat\phi^{\text{token}} = \dfrac{1 + \tfrac{3}{7}}{2} = \dfrac{5}{7} \approx 0.71
    \quad
    \hat R = \dfrac{\tfrac{1}{6} + 0}{1 + \tfrac{3}{7}} = \dfrac{7}{60} \approx 0.12
    \quad
    \hat S = 1 - \hat\phi^{\text{token}} = \dfrac{2}{7} \approx 0.29$};
\end{tikzpicture}
\caption{Coverage \eqref{eq:selgen-x-fullrate}, token coverage
\eqref{eq:selgen-x-coverage}, selective risk \eqref{eq:selgen-x-risk},
and token savings \eqref{eq:selgen-x-savings} on an
evaluation set of $N = 2$ generations. Beneath each token, as in \Cref{fig:selgen-gbar}:
the per-step decisions $g$, the per-token acceptance $\tilde g$ (latching to $0$ from
the first rejection), and the per-prefix losses $\ell$
(\textcolor{red!80!black}{red underline} = hallucinated). 
}
\label{fig:selgen-covrisk}
\end{figure}

\subsection{Constructing Selective Generators and the Risk--Coverage Curve}
\label{app:selective-generation-construct}
In this section we present a general technique for constructing a selective generator, using a confidence-rate function $\kappa_f : V^{*} \to \mathbb{R}$ for an LLM $f$ that scores each prefix $\mathbf{x}_{\le t}$ by its confidence in the next token.
For $\kappa_f$ to be useful for selective generation, it should be \emph{monotone}: larger values of $\kappa_f(\mathbf{x}_{\le t})$ should indicate that the next token $x_{t+1}$ is more likely to be correct, or equivalently, less likely to be hallucinated. Formally, given $\mathbf{x}_{\le t} \in V^{t}, \mathbf{x}'_{\le t'} \in V^{t'}$, we require that:
\begin{equation}
\kappa_f(\mathbf{x}_{\le t}) \geq \kappa_f(\mathbf{x}'_{\le t'}) \quad \Longrightarrow \quad \ell(\mathbf{x}_{\le t+1}) \leq \ell(\mathbf{x}'_{\le t'+1}).
\end{equation}
Namely, if the confidence in the next token $x_{t+1}$ is higher than that of $x'_{t+1}$, then $x_{t+1}$ should be more likely to be correct than $x'_{t+1}$.

Given such a confidence function and a threshold $\tau \in \mathbb{R}$, we construct a selection function $g_\tau$ by thresholding the confidence:
\begin{equation}
g_\tau(\mathbf{x}_{\le t}) \coloneqq
\begin{cases}
  1 & \text{if } \kappa_f(\mathbf{x}_{\le t}) \ge \tau,\\[2pt]
  0 & \text{if } \kappa_f(\mathbf{x}_{\le t}) < \tau.
\end{cases}
\label{eq:selgen-x-gtau}
\end{equation}
Since the decision gating $x_{t+1}$ is computed from the served prefix $\mathbf{x}_{\le t}$ alone, $(f, g_\tau)$ is a selective generator for every $\tau$: generation halts the first time the confidence dips below the threshold.

\paragraph{The risk--coverage curve.}
The threshold $\tau$ controls the trade-off between token-coverage and risk: raising it withholds more aggressively, lowering the token-coverage and---for a monotone $\kappa_f$---the selective risk with it. Sweeping $\tau$ thus traces the \emph{risk--coverage (RC) curve} $\tau \mapsto \bigl(\hat\phi^{\text{token}}(f, g_\tau),\, \hat R(f, g_\tau)\bigr)$, which summarizes the selective-generation ability of a confidence function across all operating points; we condense it into a single number, the area under the curve, $\mathrm{AURC} = \int \hat R \,\mathrm{d}\hat\phi^{\text{token}}$ (lower is better, and comparable across confidence functions).

\end{document}